\documentclass[runningheads]{llncs}

\usepackage[preprint]{eccv} 

\usepackage{eccvabbrv}

\usepackage{graphicx}
\usepackage{booktabs}
\usepackage{multirow}
\usepackage{array}
\usepackage{enumitem}

\usepackage[numbers,sort&compress]{natbib}
\usepackage{pifont}

\usepackage[table]{xcolor}
\usepackage[accsupp]{axessibility}

\usepackage[pagebackref,breaklinks,colorlinks,citecolor=eccvblue]{hyperref}

\usepackage{orcidlink}
\usepackage{graphicx}
\usepackage{caption} 
\usepackage{subcaption}
\newcommand{\name}{SpaMEM}

\newcolumntype{L}[1]{>{\raggedright\arraybackslash}p{#1}}
\newcolumntype{C}[1]{>{\centering\arraybackslash}p{#1}}
\newcolumntype{R}[1]{>{\raggedleft\arraybackslash}p{#1}}

\definecolor{spaCyanLight}{RGB}{230,243,248}
\definecolor{spaCyanMid}{RGB}{218,238,246}
\definecolor{spaCyanRow}{RGB}{208,232,242}
\definecolor{spaSec1}{RGB}{253,237,235}
\definecolor{spaSec2}{RGB}{249,231,239}
\definecolor{spaSec3}{RGB}{240,232,247}
\definecolor{spaSec4}{RGB}{233,241,233}
\definecolor{spaGoodGreen}{RGB}{34,139,34}
\definecolor{spaPartialAmber}{RGB}{217,153,12}
\definecolor{spaBadRed}{RGB}{192,57,43}

\newcommand{\fullsup}{{\color{spaGoodGreen}\ding{51}\ding{51}\ding{51}}}
\newcommand{\partsup}{{\color{spaPartialAmber}\ding{51}}}
\newcommand{\nosup}{{\color{spaBadRed}\ding{55}}}
\newcommand{\partsupd}{{\color{spaPartialAmber}\ding{51}$^\dagger$}}

\begin{document}

\title{\name: Benchmarking Dynamic Spatial Reasoning via Perception--Memory Integration in Embodied Environments}

\titlerunning{\name: Dynamic Spatial Reasoning Benchmark}

\author{
Chih-Ting Liao$^{1,\ddagger}$\quad
Xi Xiao$^{2}$\quad
Chunlei Meng$^{3}$\quad
Zhangquan Chen$^{4}$\quad
Yitong Qiao$^{5}$\quad
Weilin Zhou$^{6}$\quad
Tianyang Wang$^{2}$\quad
Xu Zheng$^{7,*}$\quad
Xin Cao$^{1}$\\[3pt]
$^{1}$The University of New South Wales\quad
$^{2}$The University of Alabama at Birmingham\quad
$^{3}$Fudan University\\[1pt]
$^{4}$Tsinghua University\quad
$^{5}$Zhejiang University\quad
$^{6}$Xinjiang University\\[1pt]
$^{7}$The Hong Kong University of Science and Technology (Guangzhou)\\
{\tt\small zhengxu128@gmail.com} \\[3pt]
$^{\ddagger}$Project lead\quad $^{*}$Corresponding author
}
\institute{}
\authorrunning{Liao et al.}

\maketitle

\begin{abstract}
Multimodal large language models (MLLMs) have advanced static visual--spatial reasoning, yet they often fail to preserve long-horizon spatial coherence in embodied settings where beliefs must be continuously revised from egocentric observations under environmental change. We introduce \textbf{SpaMEM} (Spatial Memory from Action Sequences), a large-scale diagnostic benchmark that isolates the mechanics of spatial belief evolution via action-conditioned scene transformations (spawn, place, remove) over long interaction horizons.
SpaMEM is built on a physically grounded dataset with \textbf{10,601,392} high-fidelity images across \textbf{four modalities} (RGB, depth, instance, semantic segmentation), collected from \textbf{25,000+} interaction sequences in \textbf{1,000} procedurally generated houses. We formalize embodied spatial reasoning as a \textbf{three-level hierarchy} with \textbf{15} diagnostic tasks: \textit{Level 1} measures atomic spatial perception from single observations; \textit{Level 2} probes temporal reasoning with oracle textual state histories to factor out perceptual noise; and \textit{Level 3} requires end-to-end belief maintenance from raw visual streams under the same task dimensions. We further evaluate both \textit{short-term} (step-wise) updates and \textit{long-term} (episodic) reconstruction.
Benchmarking representative open-source VLM families reveals a consistent stacked bottleneck: coordinate-consistent grounding remains a hard ceiling, and the sharp collapse from Level 2 to Level 3 exposes a pronounced \emph{symbolic scaffolding dependency}, where models succeed with text-based bookkeeping but struggle to sustain robust visual memory. SpaMEM provides a granular diagnostic standard and motivates explicit mechanisms for state representation, belief revision, and long-horizon episodic integration.
\textbf{A subset of SpaMEM is publicly available at \url{https://huggingface.co/datasets/mill-ct-liao/SpaMEM}.}
\end{abstract}
\section{Introduction}
\label{sec:intro}

The emergence of MLLMs has significantly advanced static visual-spatial reasoning, yet their deployment in embodied environments reveals a profound gap in maintaining long-term spatial coherence. Unlike passive VQA tasks, true spatial intelligence requires an agent to integrate a continuous stream of egocentric observations into a stable allocentric belief. Recent evidence suggests that while MLLMs possess strong instantaneous perception, their internal spatial models suffer from rapid degradation during active exploration, often failing to reconcile new sensory evidence with prior knowledge \cite{zhang2026theory, huang2025vision}.

Current embodied benchmarks, however, frequently conflate three distinct failure modes. \textbf{First}, most evaluations rely on static layouts where models can exploit ``statistical co-occurrence biases'' (e.g., semantic priors of object placement) rather than performing genuine geometric reasoning. \textbf{Second}, there is a lack of decoupling between perception and memory; when an agent fails, it is unclear whether the cause is initial perceptual misalignment or a breakdown in memory retention. \textbf{Third}, the temporal stability of spatial beliefs over extended horizons remains under-explored, particularly in dynamic scenes where objects are moved or introduced mid-trajectory.

To address these challenges, we propose \textbf{SpaMEM (Spatial Memory from Action Sequences)}, a diagnostic benchmark designed to isolate and quantify the mechanics of spatial belief evolution. SpaMEM introduces dynamic scene transformations---including \textit{spawn}, \textit{place}, and \textit{remove} operations---over 25+ step sequences, effectively shattering static semantic priors. To provide a granular diagnosis, we formalize the spatial reasoning chain into a three-tier hierarchy: \textbf{(i) Atomic Perception (L1)}, which evaluates single-frame geometric grounding; \textbf{(ii) Symbolic Memory (L2)}, which isolates pure logic by bypassing the visual encoder and providing explicit text-based state descriptions; and \textbf{(iii) Dynamic Integration (L3)}, the end-to-end challenge of maintaining beliefs from raw visual streams under environmental flux.

\textbf{Our extensive evaluation across state-of-the-art MLLM families reveals a consistent stacked bottleneck in embodied reasoning.} We identify a severe \textit{Static-to-Dynamic Degradation}, where performance drops precipitously (e.g., InternVL3's $F_1$ score falling from 0.36 to 0.13) when transitioning from static frames to continuous episodic streams. Furthermore, our cross-modality analysis uncovers the \textbf{\textit{Logic-Perception Paradox}} (or \textit{Symbolic Scaffolding Dependency}). While models excel at accurate ``bookkeeping'' when provided with textual history, their internal visual world models collapse without such symbolic anchors. Finally, we observe a universal \textit{Space-Time Dissonance}: models maintain a rudimentary capacity to track the chronological order of events but fail to map them to coordinate-consistent spatial grounding. Crucially, providing explicit geometric priors (RGB-D) yields negligible benefits, suggesting that the central challenge lies in maintaining and updating persistent 3D state representations rather than sensory insufficiency.

The contributions of this work are threefold:
\textbf{(1)} We introduce SpaMEM, a benchmark that breaks 2D statistical priors through action-conditioned dynamic evolution;
\textbf{(2)} We provide a hierarchical evaluation protocol that successfully decouples perceptual errors from symbolic memory failures;
\textbf{(3)} We demonstrate that current state-of-the-art MLLMs exhibit significant temporal instability and belief inertia, providing a new diagnostic standard for future embodied AI.

\begin{figure}[t]
\centering
\includegraphics[width=\linewidth]{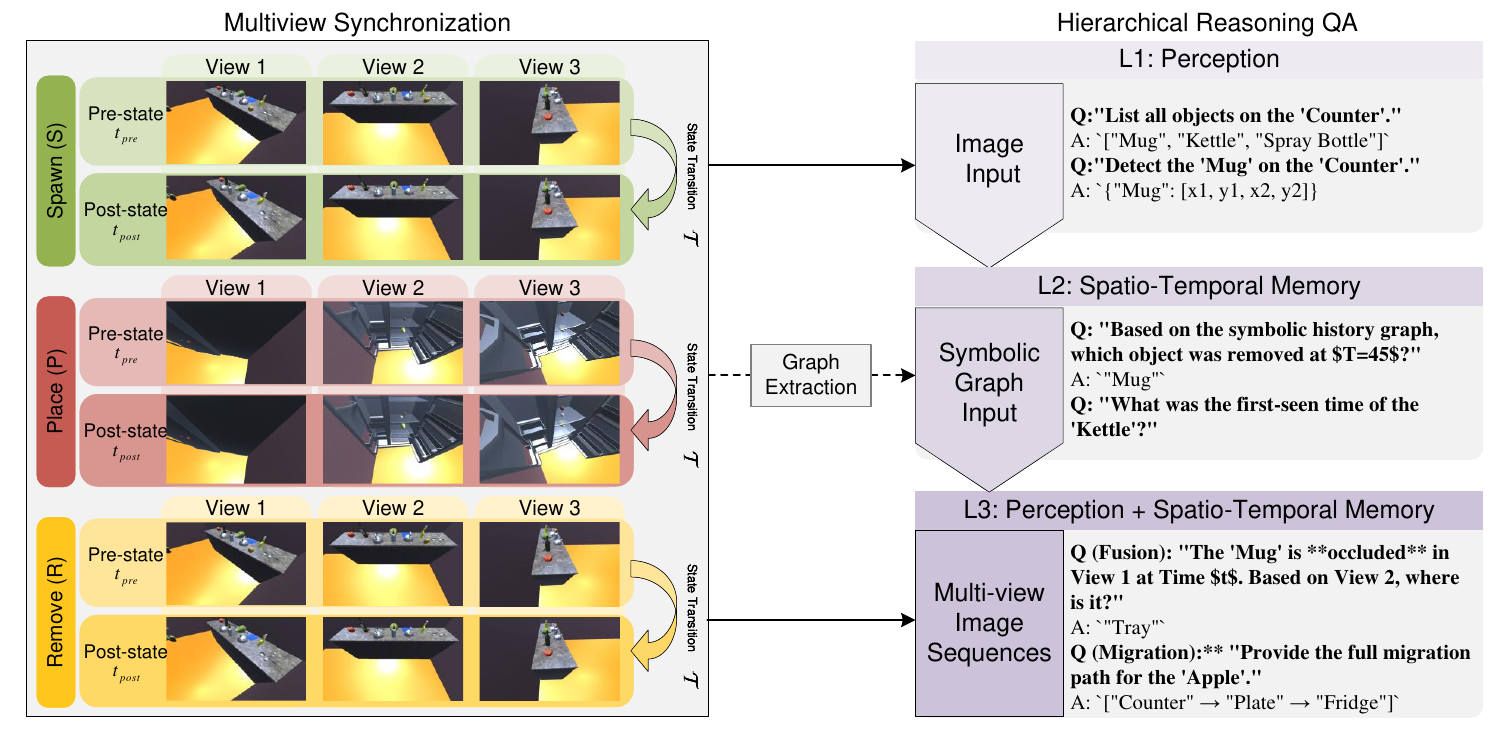}
\vspace{-12pt}
\caption{
\textbf{Overview of the SpaMEM benchmark.}
SpaMEM evaluates spatial reasoning under dynamic scene evolution.
Scenes evolve through action-conditioned transformations (\textit{spawn}, \textit{place}, \textit{remove}) over long temporal horizons.
The benchmark organizes evaluation into three hierarchical levels:
\textbf{L1} atomic spatial perception from single observations,
\textbf{L2} symbolic temporal reasoning with textual state descriptions,
and \textbf{L3} full visual-conditioned spatio-temporal integration.
}
\vspace{-24pt}
\label{fig:benchmark_overview}
\end{figure}
\section{Related Work}

\noindent \textbf{Static Environments vs. Dynamic Scene Evolution.}
A critical but often overlooked distinction in embodied AI is the state of the environment.
Most existing simulators, while supporting active movement, operate in \textit{static environments} where the spatial layout remains immutable throughout an episode
\cite{kolve2017ai2,deitke2022procthor,savva2019habitat,deitke2020robothor,shen2020igibson,li2021igibson2}.
This allows models to heavily rely on ``statistical co-occurrence biases''---predicting object locations based on 2D semantic priors rather than genuine 3D geometric understanding
\cite{survey2025spatial}.
SpaMEM diverges from these conventions by introducing \textit{dynamic scene evolution}.
By performing intentional object manipulations---such as \textit{spawn}, \textit{place}, and \textit{remove}---we create a fluid environment that forces models to constantly revise their internal spatial beliefs.
This setting is closely related to rearrangement-style embodied tasks that explicitly change object poses/states within an episode
\cite{weihs2021roomr,trabucco2022rearrangement,shukla2024maniskillhab},
and complements recent benchmarks that probe spatial belief construction under active exploration
\cite{zhang2026theory,rem2025}.
It further aligns with emerging MLLM spatial benchmarks that evaluate geometric reasoning beyond single-frame priors
\cite{open3dvqa2025,spatialscore2025,homesafebench2025}.

\noindent \textbf{Stability Decay and Long-term Spatial Memory.}
Maintaining a coherent world model in such dynamic settings requires robust long-term memory.
As interaction sequences scale beyond 15 steps, models frequently suffer from Stability Decay (or ``Soul Erosion'')
\cite{liu2024explore,rem2025}, where new perceptual inputs erroneously overwrite earlier, still-valid spatial beliefs.
This is often exacerbated by the context window limitations of Transformer-based attention mechanisms
\cite{liu2024lost}.
Related long-horizon embodied QA / navigation formulations also expose similar failure modes when agents must accumulate evidence over time
\cite{das2018eqa,das2018nmc,yu2019mteqa,chaplot2020emml}.
While recent architectural innovations propose episodic memory buffers or structured state representations
\cite{zhu20253dllm,onlineSI2026,huang2025vision,embodiedvsr2025},
SpaMEM provides a rigorous diagnostic benchmark by explicitly tracking object states through long sequences of dynamic changes.

\noindent \textbf{The Modality Gap and Belief Revision.}
A fundamental challenge in dynamic environments is \textit{Belief Revision}---the ability to update a cognitive map when new evidence contradicts prior memory.
As analyzed in the Theory of Space (ToS) benchmark \cite{zhang2026theory}, there is a catastrophic \textit{Modality Gap} where models excel at symbolic belief revision but fail at perceptual revision using RGB images
\cite{survey2025spatial,rem2025}.
A prevailing issue in current literature is the \textbf{conflation of perceptual errors and memory failures}
\cite{survey2025spatial,embodiedvsr2025}.
While an agent's inability to locate an object is often attributed to deficient long-term memory, recent evidence suggests that the primary bottleneck frequently resides in the initial perceptual grounding and geometric alignment
\cite{mosaic2026,spatialscore2025,open3dvqa2025,disjoint3dqa2025,viewspatial2025}.
This perspective is also consistent with egocentric 3D localization / interaction datasets that require stable geometry and object identity over time
\cite{mai2022egoloc,liu2022hoi4d,li2022egopat3d}.
Related long-horizon navigation settings that explicitly test map-like memory (e.g., multi-goal navigation) also report sharp degradation with increased episode complexity
\cite{wani2020multion,raychaudhuri2023reduce,gireesh2023sequence}.
\section{\name{}: Dataset and Evaluation Framework}
\label{sec:spamem}

We construct \name{}, a large-scale, physically-grounded resource that pairs an action-conditioned dynamic-scene generation pipeline with a hierarchical diagnostic evaluation protocol. This section is organized as follows.
\S\ref{subsec:dataset_comparison} positions \name{} against existing embodied and spatial-reasoning benchmarks.
\S\ref{subsec:eval_framework} introduces our evaluation framework---an \emph{Update}\,$\rightarrow$\,\emph{Retrieve} probing protocol applied at three hierarchical levels (L1/L2/L3) along four task families.
\S\ref{subsec:dataset_pipeline} describes the data generation pipeline, and \S\ref{subsec:dataset_stats} reports the resulting dataset composition.

\subsection{Comparison with Existing Benchmarks}
\label{subsec:dataset_comparison}

Existing benchmarks differ along several orthogonal axes: whether the scene evolves dynamically during an episode, whether perception and memory can be \emph{separately} probed, what output format is required, and which sensing channels are supplied. Table~\ref{tab:bench_comparison} summarizes \name{} against ten representative prior benchmarks across nine capability dimensions, ranging from passive video QA \cite{ye2024thinking,antol2015vqa,lei2018tvqa} to rearrangement-style dynamic environments \cite{weihs2021roomr,trabucco2022rearrangement,shukla2024maniskillhab} and recent dynamic-state and active-exploration benchmarks \cite{zhang2026theory,rem2025,embodiedvsr2025}.

Three properties distinguish \name{}:
\textbf{(1) Action-conditioned scene evolution} via an explicit $\{\textit{spawn}, \textit{place}, \textit{remove}\}$ action vocabulary that gradually changes scene composition and topology---strictly more expressive than pick-and-place or pose-perturbation regimes;
\textbf{(2) Symbolic$\rightarrow$visual paired evaluation} on the \emph{same} trajectory and the \emph{same} query set, where Level~2 supplies oracle textual history and Level~3 supplies only the visual stream, enabling controlled attribution of long-horizon failures to either perception or memory;
\textbf{(3) Step-wise scene-graph output}, requiring models to emit a JSON-formatted scene graph at each probe step (as opposed to free-form text, navigation actions, or VQA-style multiple-choice answers), so that the 15 diagnostic metrics decompose a single structured prediction rather than evaluating 15 independent VQA tasks.


\begin{table}[t]
\centering
\caption{\textbf{Capability comparison of \name{} with representative embodied and spatial-reasoning benchmarks across nine diagnostic dimensions.}
\textbf{Dyn.}: action-conditioned dynamic scene evolution beyond pose/state perturbation.
\textbf{Decoup.}: paired symbolic$\rightarrow$visual evaluation on the same trajectory (text-history $\leftrightarrow$ visual-only).
\textbf{Long-H.}: episodes $\ge 15$ steps with long-horizon memory probes.
\textbf{Grph-Out}: model required to emit step-wise structured scene graphs (vs.\ free-form text/action/index).
\textbf{Multi-V}: multi-viewpoint observations per state.
\textbf{RGB-D}: aligned depth supervision.
\textbf{Seg}: pixel-aligned instance/semantic segmentation.
\textbf{Belief-R}: explicit belief-revision events (remove/replace mid-trajectory).
\textbf{Diag.}: hierarchical L1/L2/L3-style diagnostic decomposition.
\fullsup{}: full support; \partsup{}: partial / restricted support; \nosup{}: not supported.
$^\dagger$ Partial: ToS provides paired text/visual evaluation but on \emph{separate} environments rather than a shared trajectory.}
\label{tab:bench_comparison}
\setlength{\tabcolsep}{3pt}
\renewcommand{\arraystretch}{1.05}
\resizebox{\textwidth}{!}{%
\begin{tabular}{@{}l|ccccccccc@{}}
\toprule
\rowcolor{spaCyanLight}
\textbf{Benchmark} & \textbf{Dyn.} & \textbf{Decoup.} & \textbf{Long-H.} & \textbf{Grph-Out} & \textbf{Multi-V} & \textbf{RGB-D} & \textbf{Seg} & \textbf{Belief-R} & \textbf{Diag.} \\
\midrule
\rowcolor{spaSec1}\multicolumn{10}{l}{\textit{(A) Passive video / VQA spatial benchmarks}} \\
VQA \cite{antol2015vqa}             & \nosup    & \nosup    & \nosup    & \nosup    & \nosup    & \nosup    & \nosup    & \nosup    & \nosup    \\
TVQA \cite{lei2018tvqa}             & \partsup  & \nosup    & \partsup  & \nosup    & \nosup    & \nosup    & \nosup    & \nosup    & \nosup    \\
VSI-Bench \cite{ye2024thinking}     & \nosup    & \nosup    & \partsup  & \nosup    & \fullsup  & \nosup    & \nosup    & \nosup    & \partsup  \\
\midrule
\rowcolor{spaSec2}\multicolumn{10}{l}{\textit{(B) Rearrangement / dynamic-scene benchmarks}} \\
RoomR \cite{weihs2021roomr}                 & \partsup  & \nosup    & \partsup  & \nosup    & \partsup  & \fullsup  & \fullsup  & \partsup  & \nosup    \\
Habitat Rearrange. \cite{trabucco2022rearrangement} & \partsup  & \nosup    & \partsup  & \nosup    & \partsup  & \fullsup  & \nosup    & \partsup  & \nosup    \\
ManiSkill-HAB \cite{shukla2024maniskillhab} & \partsup  & \nosup    & \partsup  & \nosup    & \partsup  & \fullsup  & \partsup  & \partsup  & \nosup    \\
\midrule
\rowcolor{spaSec3}\multicolumn{10}{l}{\textit{(C) Active-exploration / spatial-belief benchmarks}} \\
ToS \cite{zhang2026theory}                  & \fullsup  & \partsupd & \fullsup  & \nosup    & \fullsup  & \partsup  & \nosup    & \fullsup  & \partsup  \\
REM \cite{rem2025}                          & \fullsup  & \nosup    & \fullsup  & \nosup    & \fullsup  & \partsup  & \nosup    & \partsup  & \partsup  \\
EmbodiedVSR \cite{embodiedvsr2025}          & \partsup  & \nosup    & \fullsup  & \partsup  & \fullsup  & \partsup  & \nosup    & \partsup  & \partsup  \\
\midrule
\rowcolor{spaSec4}\multicolumn{10}{l}{\textit{(D) Long-horizon embodied QA / multi-goal benchmarks}} \\
EQA \cite{das2018eqa}                       & \nosup    & \nosup    & \fullsup  & \nosup    & \fullsup  & \partsup  & \nosup    & \nosup    & \nosup    \\
MultiON \cite{wani2020multion}              & \nosup    & \nosup    & \fullsup  & \nosup    & \fullsup  & \partsup  & \nosup    & \nosup    & \nosup    \\
\midrule
\rowcolor{spaCyanRow}
\textbf{\name{} (Ours)} & \fullsup & \fullsup & \fullsup & \fullsup & \fullsup & \fullsup & \fullsup & \fullsup & \fullsup \\
\bottomrule
\end{tabular}%
}
\end{table}

\subsection{Evaluation Framework}
\label{subsec:eval_framework}

\begin{figure}[t]
\centering
\includegraphics[width=\linewidth]{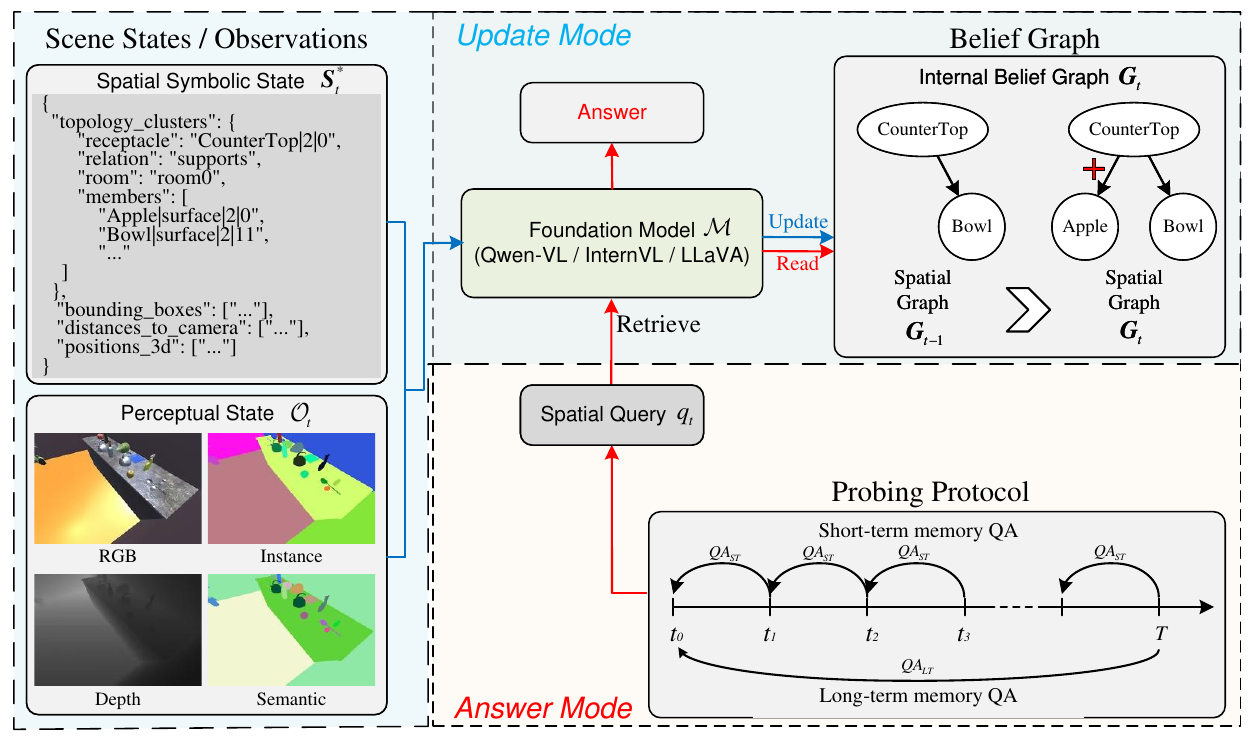}
\vspace{-18pt}
\caption{\textbf{\name{} evaluation framework.} At each step the model ingests the current observation $O_t$ (and, at L2, an oracle symbolic history $S_t^\ast$), maintains its internal state, and answers a templated query $q_t$ with a canonical JSON scene graph. The same interface is instantiated at three hierarchy levels (L1/L2/L3) and two temporal probing modes (short-term step-wise vs.\ long-term episodic).}
\label{fig:framework}
\end{figure}

\paragraph{Problem formulation.}
We evaluate a vision--language model $f_\theta$ as a \emph{structured state predictor} over an interaction episode.
At each timestep $t$, the model receives a perceptual observation $O_t$ (egocentric RGB and, optionally, aligned depth/instance/semantic channels) and, at Level~2 only, a symbolic history $S_t^{\ast}$ (a ground-truth textual summary of past states).
Given a query $q_t$ drawn from a fixed template set, the model produces a structured prediction
$\hat{y}_t = f_\theta(O_{\le t},\, S_t^{\ast},\, q_t)$
in a canonical JSON scene-graph format, which is scored against the simulator ground truth $S_t^{\ast}$ by the metric associated with $q_t$ (Fig.~\ref{fig:framework}).
Because every query is answered from the \emph{same} structured output rather than a task-specific head, the 15 diagnostic dimensions (\S\ref{subsec:tasks}) decompose a single prediction interface, and failures can be attributed to a specific competence rather than to output-format mismatch.

\paragraph{Three-level hierarchy.}
The three levels share question templates and ground-truth states but differ in the \emph{history channel} provided to the model, which lets us localize where reasoning breaks down.
\textbf{Level~1 (Atomic Perception)} removes the temporal axis entirely: the model answers from a single observation at $t{=}0$, restricted to a local \emph{topology cluster} (the set of objects supported by one receptacle). L1 establishes a perception ceiling before any memory is required.
\textbf{Level~2 (Symbolic Memory)} supplies the ground-truth textual history $S_t^{\ast}$ alongside the current observation. Since history is perfect by construction, L2 measures temporal reasoning and state maintenance in isolation from perceptual error, and thus serves as an \emph{oracle-perception upper bound} on L3.
\textbf{Level~3 (Visual Memory)} withholds $S_t^{\ast}$ and provides only the chronological visual stream $\{(I_0,t_0),\dots,(I_{T-1},t_{T-1})\}$. This is the realistic end-to-end setting: the model must construct and maintain its own history from raw pixels.
The L2$\rightarrow$L3 gap therefore isolates the cost of replacing oracle text with visual self-perception.

\paragraph{Temporal probing modes.}
At Levels~2 and~3 we issue queries under two regimes that share trajectories and templates but differ in \emph{when} they probe.
\textbf{Short-term} probing queries each step transition $t{-}1\rightarrow t$, targeting incremental updates and working-memory stability.
\textbf{Long-term} probing queries only at the episode end $t{=}T$, requiring retrieval over the full horizon $[0,T]$ and exposing long-horizon forgetting and belief drift.
Contrasting the two regimes separates models that fail to update from models that fail to retain.

\subsection{Task Families and Metrics}
\label{subsec:tasks}

\name{} measures 15 diagnostic dimensions grouped into four task families: \emph{Semantic \& Inventory} (object recognition and lifelong inventory), \emph{Spatio-Temporal Localization} (2D grounding, temporal lifespan, and depth), \emph{Relations \& Tracking} (neighborhood relations and migration-path reconstruction), and \emph{Counting \& Events} (category/instance counts, step-wise event detection, and cumulative state reconstruction). Table~\ref{tab:hierarchy_summary} maps each dimension to the metric used at every level; we define each metric once and reuse it across L1--L3.

For dimensions with established metrics we follow standard practice: set-level recognition and inventory use the $F_1$ score between predicted and ground-truth object sets; 2D grounding uses mean IoU (mIoU); depth uses absolute relative error (AbsRel) and Spearman's $\rho$ for depth-order; relative relations use Recall@$k$ and egocentric direction accuracy; and counting uses absolute error (AE), MAPE, and Acc@$k$. Two temporal dimensions require benchmark-specific metrics. \emph{Temporal lifespan} (TL) measures whether a model recovers the interval $[\text{first},\text{last}]$ during which an object exists, scored by a temporal IoU; \emph{migration-path reconstruction} (STT) measures whether the predicted sequence of containers an object visits matches the ground-truth path, scored by normalized edit similarity:
\begin{equation}
\label{eq:tiou}
\text{tIoU} =
\frac{\max\!\big(0,\,\min(p_{\text{last}},g_{\text{last}})-\max(p_{\text{first}},g_{\text{first}})+1\big)}
{\max(p_{\text{last}},g_{\text{last}})-\min(p_{\text{first}},g_{\text{first}})+1},
\qquad
\text{NES} = 1 - \frac{\text{ED}(P,G)}{\max(|P|,|G|,1)},
\end{equation}
where $\text{ED}(\cdot,\cdot)$ is the edit distance between predicted path $P$ and ground-truth path $G$. Event detection (SED) and cumulative state reconstruction (CSR) report the added and removed object sets separately, each scored by $F_1$, so that a model's ability to detect local changes can be distinguished from its ability to integrate them into a coherent global state.

\begin{table}[t]
\centering
\caption{\textbf{\name{} task hierarchy at a glance.} Each row is a diagnostic dimension; columns specify the metric used at each level. Metrics in parentheses are reported as secondary signals. tIoU and NES are defined in Eq.~\ref{eq:tiou}; all other metrics follow standard definitions (\S\ref{subsec:tasks}). \emph{Probing inputs}: L1 sees a single observation; L2 receives the observation plus oracle textual history; L3 receives only the chronological visual stream.}
\label{tab:hierarchy_summary}
\setlength{\tabcolsep}{4pt}
\renewcommand{\arraystretch}{1.10}
\resizebox{\textwidth}{!}{%
\begin{tabular}{@{}l l c c c@{}}
\toprule
\rowcolor{spaCyanLight}
\textbf{Task Family} & \textbf{Diagnostic Dimension} & \textbf{L1 (Atomic)} & \textbf{L2 (Symbolic Memory)} & \textbf{L3 (Visual Memory)} \\
\midrule
\rowcolor{spaSec1}\multicolumn{5}{l}{\textit{Semantic \& Inventory (SI)}} \\
SOR    & Object set per frame / step  & $F_1$       & SOR-M ($F_1$) & SOR-M ($F_1$) \\
LOR    & Lifelong object inventory    & ---         & LOR ($F_1$)   & LOR ($F_1$)   \\
\midrule
\rowcolor{spaSec2}\multicolumn{5}{l}{\textit{Spatio-Temporal Localization (STL)}} \\
VGL    & 2D bounding-box grounding    & mIoU        & VGL-M (mIoU)  & VGL-M (mIoU)  \\
TL     & Temporal lifespan of object  & ---         & TL (tIoU)     & TL (tIoU)     \\
DPE    & Depth / proximity            & AbsRel, $\rho$ & ---        & ---           \\
\midrule
\rowcolor{spaSec3}\multicolumn{5}{l}{\textit{Relations \& Tracking (RT)}} \\
RSR    & Neighborhood + egocentric dir.\ & R@3, Dir.\ Acc.\ & RSR-M (R@3) & RSR-M (R@3) \\
STT    & Migration-path reconstruction & ---        & STT (NES)     & STT (NES)     \\
\midrule
\rowcolor{spaSec4}\multicolumn{5}{l}{\textit{Counting (CT) \& Events (EV)}} \\
CC/IC  & Categorical / instance count  & AE, MAPE, Acc@\{0,1\} & --- & ---     \\
SED    & Step-wise event detection (add/remove) & --- & SED-Add, SED-Rem ($F_1$) & SED-Add, SED-Rem ($F_1$) \\
CSR    & Cumulative state reconstruction (add/remove) & --- & CSR-Add, CSR-Rem ($F_1$) & CSR-Add, CSR-Rem ($F_1$) \\
\bottomrule
\end{tabular}%
}
\end{table}

\subsection{Data Generation Pipeline}
\label{subsec:dataset_pipeline}

\begin{figure}[t]
\centering
\includegraphics[width=\linewidth]{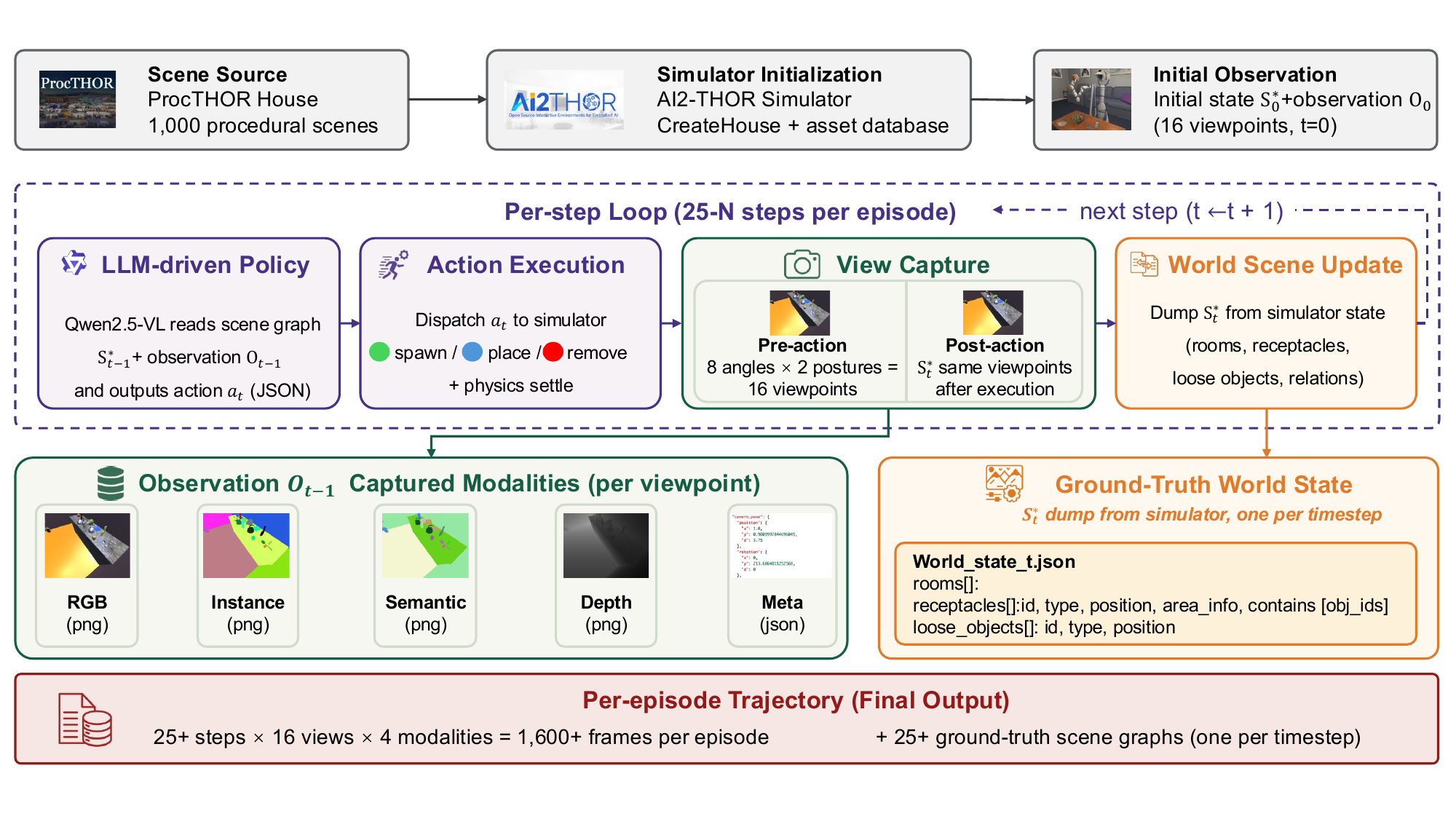}
\caption{\textbf{\name{} data generation pipeline.} At each step $t$, an LLM policy reads the world graph $\mathcal{G}_{t-1}$ and the previous observation $O_{t-1}$, samples an action $a_t \in \{\textit{spawn}, \textit{place}, \textit{remove}\}$, and the simulator executes the action and settles physics. Sixteen-view RGB/Depth/Instance/Semantic observations are rendered to form $O_t$, and the ground-truth scene state $S^\ast_t$ is dumped directly from simulator state (\emph{independent of the policy LLM}). QA templates then instantiate the 15 diagnostic probes from $S^\ast_t$.}
\label{fig:pipeline_overview}
\end{figure}

We construct \name{} through an automated pipeline that orchestrates an LLM policy inside the ProcTHOR-10K simulator \cite{deitke2022procthor}, balancing semantic reasoning over the scene with low-level physical constraints to ensure scene diversity and structural integrity. The full pipeline is illustrated in Fig.~\ref{fig:pipeline_overview}.

\paragraph{Structured scene representation: the World Graph.}
Rather than exposing raw point clouds or voxels to the decision-making agent, we abstract each 3D environment into a hierarchical \textit{World Graph} $\mathcal{G}$. For each room $\mathcal{R}_i$, the graph records:
\textbf{(i) Receptacles}---objects with containment properties (e.g., \textit{Cabinet}, \textit{Table}); for each, we compute the axis-aligned bounding box (AABB) and estimate the available surface area $A_{\text{free}}$ in cm$^2$.
\textbf{(ii) Object hierarchy}---a set of parent--child relations $\mathcal{E} = \{(o_j, r_k)\}$ where object $o_j$ is supported by receptacle $r_k$.
\textbf{(iii) Spatial attributes}---3D coordinates, rotation, and semantic category for every instance.
This structured representation allows the LLM policy to plan over the spatial layout without processing raw visual features during the decision stage.

\paragraph{LLM-guided interaction policy.}
We employ Qwen2.5-7B-Instruct \cite{qwen2025qwen25vl} as a centralized reasoning engine to generate sequences of environment-modifying actions. The agent operates on a discrete action space $\mathcal{A} = \{\textit{spawn}, \textit{place}, \textit{remove}\}$. To maintain physical realism, the agent is constrained to a curated set of small, pickable assets (e.g., \textit{Apple}, \textit{CellPhone}, \textit{Mug}) and is instructed to avoid unstable or illogical targets such as \textit{Toilet} or bare \textit{Floor}. The decision process follows a ``Think-and-Act'' cycle: at each step the LLM reads the current $\mathcal{G}$ and emits a JSON-formatted command specifying the target object and the destination receptacle. Crucially, the ground-truth scene state $S_t^\ast$ used to instantiate evaluation queries is dumped \emph{directly from the simulator}, not from the policy LLM, so \name{} ground truth is independent of policy bias.

\paragraph{Physically-grounded object placement.}
To bridge LLM textual outputs and the 3D simulator, we implement a \textit{Smart Placement} heuristic. When an action places object $o$ onto receptacle $r$, the system invokes
\begin{equation}
P_{\text{new}} = \mathrm{find\_smart\_position}(o, r, \mathcal{H}),
\end{equation}
where $\mathcal{H}$ is a set of category-specific heuristics (e.g., \texttt{sink\_min} and \texttt{hover\_max} parameters). For enclosed receptacles (\textit{Fridge}, \textit{Drawer}), the system automatically executes a pre-action \texttt{OpenObject} command, followed by a point-cloud-based search for valid coordinates above interior shelves. Failed placements trigger a recovery loop that prompts the LLM for an alternative location or object.

\paragraph{Multimodal acquisition and scalable execution.}
For every successful modification, the agent teleports to 8 azimuthal angles ($0^\circ$ to $315^\circ$) at two camera heights (standing and crouching), producing 16 distinct perspectives per state. Each viewpoint generates an RGB image, an aligned depth map, and pixel-aligned instance and semantic segmentation masks; we additionally log the 6-DoF camera pose, 2D/3D bounding boxes, and the updated $\mathcal{G}$. Generation is parallelized across workers, each managing an independent AI2-THOR \cite{kolve2017ai2} instance with a periodic ``soft reset'' every five houses to clear Unity engine artifacts, while a single GPU-accelerated server handles centralized LLM inference to minimize VRAM fragmentation.

\subsection{Dataset Statistics and Scene Dynamics}
\label{subsec:dataset_stats}

\begin{figure}[t]
  \centering
  \includegraphics[width=\linewidth]{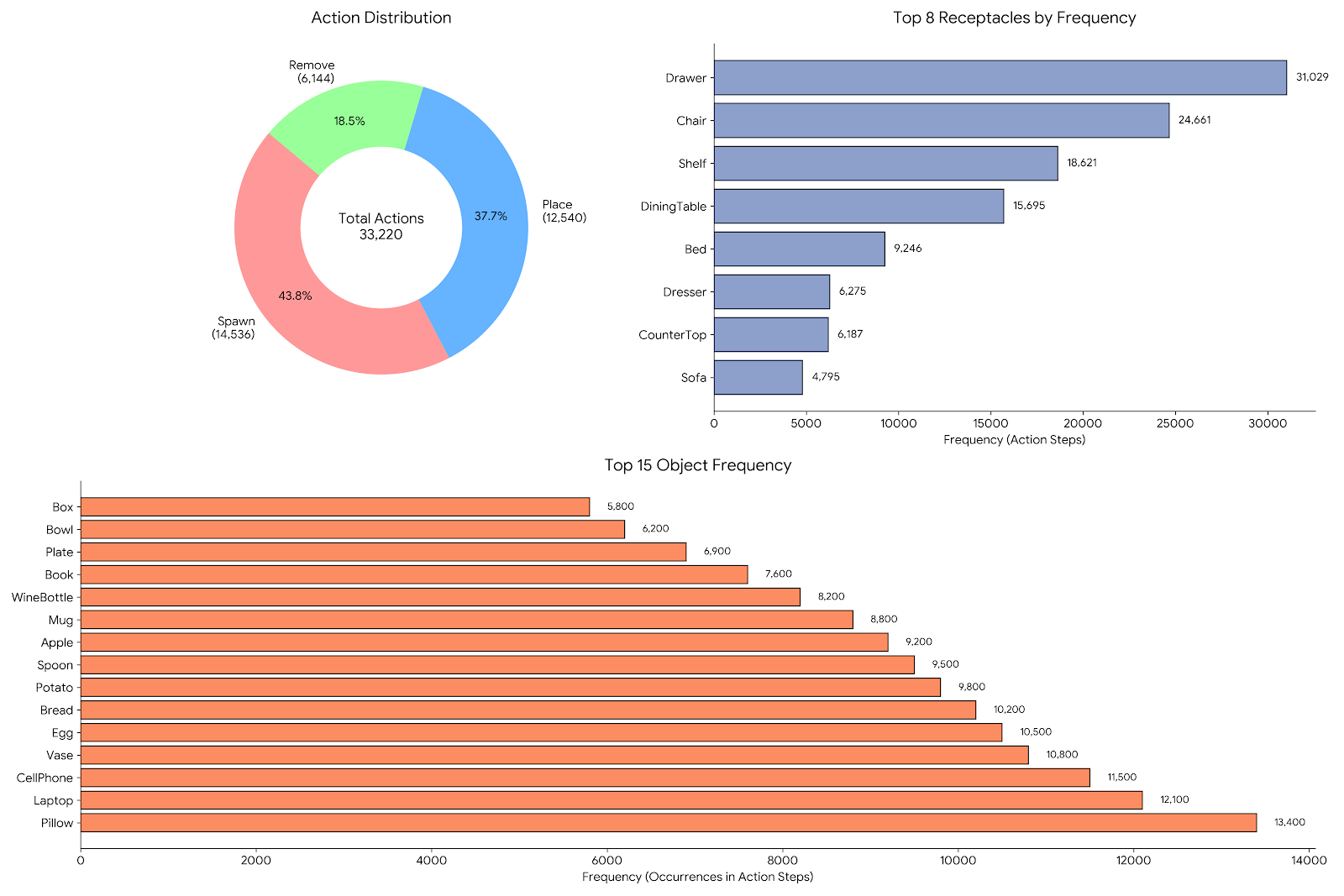}
  \caption{\textbf{\name{} dataset statistics.} (Left) Action distribution showing the balance between scene population (\textit{spawn}, \textit{place}) and belief-revision actions (\textit{remove}). (Right) Top-8 receptacle interaction frequency, highlighting the dominance of occluding containers such as \textit{Drawer}. (Bottom) Top-15 object frequency distribution across the 103 unique categories, illustrating semantic and scale diversity.}
  \label{fig:dataset_stats}
\end{figure}

Table~\ref{tab:dataset_stats} summarizes the overall scale of \name{}, and Fig.~\ref{fig:dataset_stats} visualizes the underlying action, object, and receptacle distributions.

\paragraph{Overall scale.}
\name{} spans \textbf{1{,}000 unique procedural houses} drawn from ProcTHOR-10K \cite{deitke2022procthor} and records multimodal observations (RGB, depth, instance, semantic). Aggregating 16 viewpoints per state change yields a total of \textbf{10{,}601{,}392 high-fidelity frames}, providing the visual and geometric redundancy needed to diagnose the Space-Time Dissonance reported in Sec.~\ref{sec:high_level_analysis}.

\paragraph{Action distribution and causal evolution.}
The environment evolves through \textbf{33{,}220 discrete action steps} (Fig.~\ref{fig:dataset_stats}, Left). The distribution is intentionally skewed to evaluate belief evolution: \textit{Spawn} (14{,}536) and \textit{Place} (12{,}540) operations dominate, progressively increasing scene complexity and clutter, while \textit{Remove} actions (6{,}144) explicitly stress-test belief revision and perceptual overwriting. This action-conditioned evolution shatters the static co-occurrence biases prevalent in conventional embodied benchmarks.

\paragraph{Semantic, geometric, and receptacle diversity.}
\name{} features \textbf{103 unique object types} interacting with \textbf{22 receptacle types}. As shown in Fig.~\ref{fig:dataset_stats} (bottom), high-volume objects are more easily grounded, whereas fine-grained or thin items such as \textit{Pencils} and \textit{Forks} constitute a \emph{perceptual blindspot} for current VLM tokenizers. Fig.~\ref{fig:dataset_stats} (right) further reveals that \textit{Drawer} is the most frequently interacted container (31{,}029 steps); the prevalence of constrained-volume receptacles---\textit{Shelves} (18{,}621) and \textit{Dressers} (6{,}275)---forces models to maintain robust episodic memory across line-of-sight breaks, in contrast to open-surface receptacles like \textit{Dining Tables} (15{,}695) and \textit{Beds} (9{,}246).

\begin{table}[t]
\centering
\caption{\textbf{\name{} dataset summary.} Scale, action composition, and environment diversity aggregated across 1{,}000 procedural houses generated with ProcTHOR-10K \cite{deitke2022procthor}.}
\label{tab:dataset_stats}
\setlength{\tabcolsep}{6pt}
\renewcommand{\arraystretch}{1.10}
\resizebox{\textwidth}{!}{%
\begin{tabular}{@{}l r l@{}}
\toprule
\rowcolor{spaCyanLight}
\textbf{Metric} & \textbf{Count} & \textbf{Description} \\
\midrule
\rowcolor{spaSec1}\multicolumn{3}{l}{\textit{Scale \& Volume}} \\
Total procedural houses        & 1{,}000          & Diverse indoor layouts via ProcTHOR-10K \\
Total interaction steps        & 33{,}220         & Discrete environment state changes \\
Total rendered frames          & 10{,}601{,}392   & Multimodal: RGB, Depth, Instance, Semantic \\
Observation viewpoints / state & 16               & 8 azimuthal angles $\times$ 2 camera heights \\
\midrule
\rowcolor{spaSec2}\multicolumn{3}{l}{\textit{Action Distribution}} \\
Spawn actions                  & 14{,}536         & Introduction of new objects \\
Place actions                  & 12{,}540         & Manipulation and spatial relocation \\
Remove actions                 & 6{,}144          & Deletion (probes belief revision) \\
\midrule
\rowcolor{spaSec3}\multicolumn{3}{l}{\textit{Scene Diversity}} \\
Unique object types            & 103              & e.g., AlarmClock, ButterKnife, CellPhone \\
Unique receptacle types        & 22               & e.g., Drawer, Chair, Shelf, Bed \\
\midrule
\rowcolor{spaSec4}\multicolumn{3}{l}{\textit{Evaluation Protocol}} \\
Hierarchy levels               & 3 (L1/L2/L3)     & Static / text-conditioned / visual-conditioned \\
Diagnostic metrics             & 15               & SOR, VGL, DPE, RSR, CC, IC, SOR-M, LOR, VGL-M, TL, RSR-M, STT, SED, CSR \\
Temporal probing modes         & 2                & Short-term (step-wise) and long-term (episodic) \\
\bottomrule
\end{tabular}%
}
\end{table}

\section{Experiments}
\label{sec:experiments}

\subsection{Setup}
\label{subsec:setup}

\paragraph{Models.}
We benchmark three representative open-source Vision--Language Model (VLM) families:
InternVL (InternVL2 \cite{chen2024internvl2}, InternVL2.5 \cite{chen2024internvl25}, InternVL3 \cite{chen2025internvl3}),
LLaVA (LLaVA-NeXT \cite{liu2024llavanext}, LLaVA-OneVision \cite{li2024llavaonevision}),
and Qwen-VL (Qwen2-VL \cite{wang2024qwen2vl}, Qwen2.5-VL \cite{qwen2025qwen25vl}, Qwen3-VL \cite{qwen2025qwen3vl}).
These models cover diverse training recipes and architectural designs and remain fully reproducible via public checkpoints.

\paragraph{Input modalities.}
We evaluate two input configurations to study the impact of explicit geometric cues.
RGB uses only egocentric RGB observations.
RGB-D pairs each RGB frame with an aligned depth map (treated as an additional visual channel synchronized at each timestep),
supporting depth and proximity estimation and facilitating spatial reasoning.

\paragraph{Temporal evaluation paradigms.}
Following Section~\ref{sec:spamem}, we evaluate two temporal modes.
\textit{Short-term} (immediate) mode queries the model step-by-step ($t \rightarrow t{+}1$) to assess incremental state updates.
\textit{Long-term} (episodic) mode queries only at the end of an episode ($t{=}T$) to assess global state reconstruction.
This design separates local event perception from global state integration.

\subsection{Level 1: Static Spatial Perception}
\label{subsec:l1_results_analysis}

Level~1 evaluates static spatial competence from single frames (Table~\ref{tab:l1_results} and Table~\ref{tab:l1_rgbd_results}).
Across model families and both RGB/RGB-D settings, we observe consistent failure patterns that already constrain performance before introducing temporal complexity.

\paragraph{\textbf{\emph{Grounding collapses.}}}
Visual grounding (VGL), measured by mean IoU (mIoU), is nearly non-functional: scores remain at $0.00$--$0.01$ for almost all models.
This indicates that while models may recognize object presence, they struggle to map semantics to coordinate-consistent localization in cluttered indoor scenes.

\paragraph{\textbf{\emph{Strong semantics does not imply spatial competence.}}}
Semantic object recognition (SOR) is comparatively stronger, with the InternVL family leading.
InternVL3 \cite{chen2025internvl3} achieves the highest $F_1$ (0.36 in RGB; 0.35 in RGB-D), whereas Qwen3-VL \cite{qwen2025qwen3vl} underperforms in these embodied indoor scenes ($F_1 \approx 0.10$--$0.14$).
This suggests that general-purpose knowledge does not automatically transfer to high-fidelity recognition under egocentric, cluttered conditions.

\paragraph{\textbf{\emph{Geometry helps, but fusion remains fragile.}}}
Depth and proximity estimation (DPE) improves when depth is provided: Qwen3-VL \cite{qwen2025qwen3vl} attains the lowest AbsRel (0.42) under RGB-D.
However, geometric gains do not reliably propagate to semantics: for example, InternVL2.5's \cite{chen2024internvl25} SOR $F_1$ slightly drops from 0.31 (RGB) to 0.29 (RGB-D),
indicating that multi-modal fusion may not consistently align low-level geometry with high-level semantic labeling.

\paragraph{\textbf{\emph{Egocentric direction is a blind spot.}}}
Relative spatial relationship (RSR) reasoning exposes a severe deficit in interpreting egocentric directions (left/right/behind):
direction accuracy stays near zero across models.
This points to missing egocentric inductive biases and limited ability to infer relative direction from a first-person viewpoint.

\paragraph{\textbf{\emph{Counting survives without localization.}}}
Counting tasks (CC/IC) are relatively more stable.
InternVL3 \cite{chen2025internvl3} reaches the best Acc@1 (0.53 in RGB; 0.52 in RGB-D), suggesting coarse instance awareness can be maintained even when precise grounding fails.
This provides limited support for inventory-style reasoning but remains insufficient for spatial manipulation.

\begin{table*}[t]
\centering
\caption{Evaluation Results on Level 1 (Static Spatial Perception) Tasks. Metrics are presented as Mean $\pm$ SD.}
\vspace{-8pt}
\label{tab:l1_results}
\resizebox{\textwidth}{!}{
\begin{tabular}{l | ccc | c | cc | cc | cc | cccc}
\toprule
\rowcolor{spaCyanLight}
\textbf{Model} & \multicolumn{3}{c|}{\textbf{SOR}} & \textbf{VGL} & \multicolumn{2}{c|}{\textbf{DPE}} & \multicolumn{2}{c|}{\textbf{RSR}} & \multicolumn{2}{c|}{\textbf{CC}} & \multicolumn{4}{c}{\textbf{IC}} \\
\rowcolor{spaCyanLight}
\multicolumn{1}{l|}{} & $F_1 \uparrow$ & Prec. $\uparrow$ & Rec. $\uparrow$ & mIoU $\uparrow$ & AbsRel $\downarrow$ & $\rho \uparrow$ & R@3 $\uparrow$ & Dir. $\uparrow$ & AE $\downarrow$ & MAPE $\downarrow$ & AE $\downarrow$ & Acc@0 $\uparrow$ & Acc@1 $\uparrow$ & MAPE $\downarrow$ \\
\midrule

\rowcolor{spaCyanMid}
\multicolumn{15}{l}{\textit{\textbf{InternVL series}}} \\

InternVL2 & 
0.30 {\scriptsize $\pm$ 0.27} & 0.36 {\scriptsize $\pm$ 0.30} & 0.28 {\scriptsize $\pm$ 0.29} & 
0.01 {\scriptsize $\pm$ 0.02} & 
0.58 {\scriptsize $\pm$ 0.16} & 0.23 {\scriptsize $\pm$ 0.71} & 
0.20 {\scriptsize $\pm$ 0.28} & 0.02 {\scriptsize $\pm$ 0.12} & 
3.45 {\scriptsize $\pm$ 2.97} & 0.59 {\scriptsize $\pm$ 0.29} & 
2.74 {\scriptsize $\pm$ 2.45} & 0.17 {\scriptsize $\pm$ 0.37} & 0.32 {\scriptsize $\pm$ 0.47} & 0.54 {\scriptsize $\pm$ 0.49} \\

InternVL2.5 & 
0.31 {\scriptsize $\pm$ 0.27} & 0.39 {\scriptsize $\pm$ 0.33} & 0.28 {\scriptsize $\pm$ 0.26} & 
0.00 {\scriptsize $\pm$ 0.00} & 
0.55 {\scriptsize $\pm$ 0.18} & -0.06 {\scriptsize $\pm$ 0.85} & 
0.16 {\scriptsize $\pm$ 0.29} & 0.02 {\scriptsize $\pm$ 0.13} & 
2.27 {\scriptsize $\pm$ 2.20} & 0.48 {\scriptsize $\pm$ 0.45} & 
2.02 {\scriptsize $\pm$ 2.18} & 0.27 {\scriptsize $\pm$ 0.44} & 0.49 {\scriptsize $\pm$ 0.50} & 0.40 {\scriptsize $\pm$ 0.48} \\

InternVL3 & 
0.36 {\scriptsize $\pm$ 0.29} & 0.48 {\scriptsize $\pm$ 0.38} & 0.31 {\scriptsize $\pm$ 0.27} & 
0.00 {\scriptsize $\pm$ 0.01} & 
0.66 {\scriptsize $\pm$ 0.08} & -0.03 {\scriptsize $\pm$ 0.50} & 
0.13 {\scriptsize $\pm$ 0.26} & 0.02 {\scriptsize $\pm$ 0.10} & 
1.83 {\scriptsize $\pm$ 1.81} & 0.44 {\scriptsize $\pm$ 0.57} & 
2.02 {\scriptsize $\pm$ 2.13} & 0.20 {\scriptsize $\pm$ 0.40} & 0.53 {\scriptsize $\pm$ 0.50} & 0.40 {\scriptsize $\pm$ 0.38} \\

\rowcolor{spaCyanMid}
\multicolumn{15}{l}{\textit{\textbf{LLaVA series}}} \\

LLaVA-NeXT & 
0.21 {\scriptsize $\pm$ 0.23} & 0.33 {\scriptsize $\pm$ 0.37} & 0.19 {\scriptsize $\pm$ 0.22} & 
0.00 {\scriptsize $\pm$ 0.00} & 
- & - & 
0.19 {\scriptsize $\pm$ 0.32} & 0.01 {\scriptsize $\pm$ 0.10} & 
3.51 {\scriptsize $\pm$ 2.82} & 0.65 {\scriptsize $\pm$ 0.33} & 
3.57 {\scriptsize $\pm$ 2.97} & 0.12 {\scriptsize $\pm$ 0.32} & 0.24 {\scriptsize $\pm$ 0.43} & 0.66 {\scriptsize $\pm$ 0.45} \\

LLaVA-OneVision & 
0.25 {\scriptsize $\pm$ 0.25} & 0.31 {\scriptsize $\pm$ 0.32} & 0.23 {\scriptsize $\pm$ 0.24} & 
0.00 {\scriptsize $\pm$ 0.00} & 
0.65 {\scriptsize $\pm$ 0.13} & 0.50 {\scriptsize $\pm$ 0.59} & 
0.05 {\scriptsize $\pm$ 0.16} & 0.00 {\scriptsize $\pm$ 0.00} & 
3.03 {\scriptsize $\pm$ 2.85} & 0.56 {\scriptsize $\pm$ 0.39} & 
2.37 {\scriptsize $\pm$ 2.59} & 0.21 {\scriptsize $\pm$ 0.41} & 0.52 {\scriptsize $\pm$ 0.50} & 0.45 {\scriptsize $\pm$ 0.44} \\

\rowcolor{spaCyanMid}
\multicolumn{15}{l}{\textit{\textbf{Qwen-VL series}}} \\

Qwen2-VL & 
0.24 {\scriptsize $\pm$ 0.29} & 0.34 {\scriptsize $\pm$ 0.39} & 0.20 {\scriptsize $\pm$ 0.25} & 
0.00 {\scriptsize $\pm$ 0.00} & 
- & - & 
0.04 {\scriptsize $\pm$ 0.14} & 0.00 {\scriptsize $\pm$ 0.00} & 
2.61 {\scriptsize $\pm$ 2.38} & 0.51 {\scriptsize $\pm$ 0.35} & 
2.19 {\scriptsize $\pm$ 2.39} & 0.24 {\scriptsize $\pm$ 0.43} & 0.48 {\scriptsize $\pm$ 0.50} & 0.47 {\scriptsize $\pm$ 0.62} \\

Qwen2.5-VL & 
0.30 {\scriptsize $\pm$ 0.29} & 0.39 {\scriptsize $\pm$ 0.37} & 0.26 {\scriptsize $\pm$ 0.26} & 
0.00 {\scriptsize $\pm$ 0.00} & 
- & - & 
0.00 {\scriptsize $\pm$ 0.00} & 0.00 {\scriptsize $\pm$ 0.00} & 
2.05 {\scriptsize $\pm$ 1.95} & 0.48 {\scriptsize $\pm$ 0.44} & 
2.25 {\scriptsize $\pm$ 2.42} & 0.24 {\scriptsize $\pm$ 0.43} & 0.53 {\scriptsize $\pm$ 0.50} & 0.52 {\scriptsize $\pm$ 0.66} \\

Qwen3-VL & 
0.10 {\scriptsize $\pm$ 0.21} & 0.15 {\scriptsize $\pm$ 0.31} & 0.08 {\scriptsize $\pm$ 0.18} & 
0.00 {\scriptsize $\pm$ 0.00} & 
0.43 {\scriptsize $\pm$ 0.07} & 0.59 {\scriptsize $\pm$ 0.45} & 
0.13 {\scriptsize $\pm$ 0.24} & 0.02 {\scriptsize $\pm$ 0.15} & 
3.35 {\scriptsize $\pm$ 2.61} & 0.68 {\scriptsize $\pm$ 0.35} & 
3.62 {\scriptsize $\pm$ 3.06} & 0.11 {\scriptsize $\pm$ 0.31} & 0.24 {\scriptsize $\pm$ 0.43} & 0.67 {\scriptsize $\pm$ 0.38} \\

\bottomrule
\end{tabular}
}
\vspace{-8pt}
\end{table*}
\begin{table*}[t]
\centering
\caption{Evaluation Results on Level 1 (Static Spatial Perception) Tasks for RGB+Depth. Metrics are presented as Mean $\pm$ SD.}
\label{tab:l1_rgbd_results}
\vspace{-8pt}
\resizebox{\textwidth}{!}{
\begin{tabular}{l | ccc | c | cc | cc | cc | cccc}
\toprule
\rowcolor{spaCyanLight}
\textbf{Model} & \multicolumn{3}{c|}{\textbf{SOR}} & \textbf{VGL} & \multicolumn{2}{c|}{\textbf{DPE}} & \multicolumn{2}{c|}{\textbf{RSR}} & \multicolumn{2}{c|}{\textbf{CC}} & \multicolumn{4}{c}{\textbf{IC}} \\
\rowcolor{spaCyanLight}
\multicolumn{1}{l|}{} & $F_1 \uparrow$ & Prec. $\uparrow$ & Rec. $\uparrow$ & mIoU $\uparrow$ & AbsRel $\downarrow$ & $\rho \uparrow$ & R@3 $\uparrow$ & Dir. $\uparrow$ & AE $\downarrow$ & MAPE $\downarrow$ & AE $\downarrow$ & Acc@0 $\uparrow$ & Acc@1 $\uparrow$ & MAPE $\downarrow$ \\
\midrule

\rowcolor{spaCyanMid}
\multicolumn{15}{l}{\textit{\textbf{InternVL series}}} \\

InternVL2 & 
0.30 {\scriptsize $\pm$ 0.24} & 0.36 {\scriptsize $\pm$ 0.27} & 0.29 {\scriptsize $\pm$ 0.27} & 
0.00 {\scriptsize $\pm$ 0.00} & 
0.47 {\scriptsize $\pm$ 0.05} & -0.92 {\scriptsize $\pm$ 0.19} & 
0.17 {\scriptsize $\pm$ 0.26} & 0.01 {\scriptsize $\pm$ 0.10} & 
3.56 {\scriptsize $\pm$ 3.03} & 0.62 {\scriptsize $\pm$ 0.30} & 
3.24 {\scriptsize $\pm$ 2.74} & 0.14 {\scriptsize $\pm$ 0.35} & 0.26 {\scriptsize $\pm$ 0.44} & 0.60 {\scriptsize $\pm$ 0.43} \\

InternVL2.5 & 
0.29 {\scriptsize $\pm$ 0.27} & 0.36 {\scriptsize $\pm$ 0.34} & 0.27 {\scriptsize $\pm$ 0.25} & 
0.00 {\scriptsize $\pm$ 0.00} & 
0.54 {\scriptsize $\pm$ 0.14} & 0.02 {\scriptsize $\pm$ 0.88} & 
0.15 {\scriptsize $\pm$ 0.26} & 0.02 {\scriptsize $\pm$ 0.13} & 
2.43 {\scriptsize $\pm$ 2.28} & 0.49 {\scriptsize $\pm$ 0.41} & 
2.45 {\scriptsize $\pm$ 2.52} & 0.19 {\scriptsize $\pm$ 0.40} & 0.45 {\scriptsize $\pm$ 0.50} & 0.47 {\scriptsize $\pm$ 0.41} \\

InternVL3 & 
0.35 {\scriptsize $\pm$ 0.30} & 0.48 {\scriptsize $\pm$ 0.40} & 0.30 {\scriptsize $\pm$ 0.28} & 
0.00 {\scriptsize $\pm$ 0.00} & 
- & - & 
0.18 {\scriptsize $\pm$ 0.31} & 0.02 {\scriptsize $\pm$ 0.12} & 
2.20 {\scriptsize $\pm$ 2.22} & 0.46 {\scriptsize $\pm$ 0.39} & 
2.10 {\scriptsize $\pm$ 2.30} & 0.22 {\scriptsize $\pm$ 0.42} & 0.52 {\scriptsize $\pm$ 0.50} & 0.41 {\scriptsize $\pm$ 0.39} \\

\rowcolor{spaCyanMid}
\multicolumn{15}{l}{\textit{\textbf{LLaVA series}}} \\

LLaVA-NeXT & 
0.21 {\scriptsize $\pm$ 0.23} & 0.33 {\scriptsize $\pm$ 0.37} & 0.19 {\scriptsize $\pm$ 0.22} & 
0.00 {\scriptsize $\pm$ 0.00} & 
- & - & 
0.19 {\scriptsize $\pm$ 0.32} & 0.01 {\scriptsize $\pm$ 0.10} & 
3.51 {\scriptsize $\pm$ 2.82} & 0.65 {\scriptsize $\pm$ 0.33} & 
3.57 {\scriptsize $\pm$ 2.97} & 0.12 {\scriptsize $\pm$ 0.32} & 0.24 {\scriptsize $\pm$ 0.43} & 0.66 {\scriptsize $\pm$ 0.45} \\

LLaVA-OneVision & 
0.25 {\scriptsize $\pm$ 0.25} & 0.31 {\scriptsize $\pm$ 0.32} & 0.23 {\scriptsize $\pm$ 0.24} & 
0.00 {\scriptsize $\pm$ 0.00} & 
0.65 {\scriptsize $\pm$ 0.13} & 0.50 {\scriptsize $\pm$ 0.59} & 
0.05 {\scriptsize $\pm$ 0.16} & 0.00 {\scriptsize $\pm$ 0.00} & 
3.03 {\scriptsize $\pm$ 2.85} & 0.56 {\scriptsize $\pm$ 0.39} & 
2.37 {\scriptsize $\pm$ 2.59} & 0.21 {\scriptsize $\pm$ 0.41} & 0.52 {\scriptsize $\pm$ 0.50} & 0.45 {\scriptsize $\pm$ 0.44} \\

\rowcolor{spaCyanMid}
\multicolumn{15}{l}{\textit{\textbf{Qwen-VL series}}} \\

Qwen2-VL & 
0.20 {\scriptsize $\pm$ 0.29} & 0.29 {\scriptsize $\pm$ 0.39} & 0.17 {\scriptsize $\pm$ 0.25} & 
0.00 {\scriptsize $\pm$ 0.00} & 
- & - & 
0.05 {\scriptsize $\pm$ 0.14} & 0.00 {\scriptsize $\pm$ 0.02} & 
2.77 {\scriptsize $\pm$ 2.65} & 0.50 {\scriptsize $\pm$ 0.32} & 
2.91 {\scriptsize $\pm$ 3.12} & 0.16 {\scriptsize $\pm$ 0.37} & 0.40 {\scriptsize $\pm$ 0.49} & 0.51 {\scriptsize $\pm$ 0.37} \\

Qwen2.5-VL & 
0.29 {\scriptsize $\pm$ 0.29} & 0.40 {\scriptsize $\pm$ 0.39} & 0.24 {\scriptsize $\pm$ 0.25} & 
0.01 {\scriptsize $\pm$ 0.06} & 
- & - & 
0.00 {\scriptsize $\pm$ 0.00} & 0.00 {\scriptsize $\pm$ 0.00} & 
2.65 {\scriptsize $\pm$ 2.33} & 0.56 {\scriptsize $\pm$ 0.37} & 
2.92 {\scriptsize $\pm$ 2.73} & 0.16 {\scriptsize $\pm$ 0.37} & 0.35 {\scriptsize $\pm$ 0.48} & 0.59 {\scriptsize $\pm$ 0.54} \\

Qwen3-VL & 
0.14 {\scriptsize $\pm$ 0.25} & 0.22 {\scriptsize $\pm$ 0.38} & 0.11 {\scriptsize $\pm$ 0.21} & 
0.00 {\scriptsize $\pm$ 0.01} & 
0.42 {\scriptsize $\pm$ 0.20} & 0.32 {\scriptsize $\pm$ 0.38} & 
0.19 {\scriptsize $\pm$ 0.27} & 0.04 {\scriptsize $\pm$ 0.20} & 
3.26 {\scriptsize $\pm$ 2.61} & 0.66 {\scriptsize $\pm$ 0.38} & 
3.72 {\scriptsize $\pm$ 3.06} & 0.10 {\scriptsize $\pm$ 0.30} & 0.24 {\scriptsize $\pm$ 0.43} & 0.69 {\scriptsize $\pm$ 0.39} \\

\bottomrule
\end{tabular}
}
\vspace{-8pt}
\end{table*}

\subsection{Level 2: Text-Conditioned Temporal Memory}
\label{subsec:l2_results_analysis}

Level~2 evaluates temporal reasoning with ground-truth historical summaries (Table~\ref{tab:short_term_results} and Table~\ref{tab:long_term_results}).
By providing accurate symbolic history, L2 isolates reasoning and memory integration from visual recognition noise.

\paragraph{\textbf{\emph{Text turns memory into bookkeeping.}}}
With symbolic history, Semantic and Inventory (SI) tasks improve dramatically relative to L1.
InternVL2.5 \cite{chen2024internvl25} and InternVL3 \cite{chen2025internvl3} achieve SOR-M $F_1 \approx 0.90$--$0.92$ under both short- and long-term paradigms,
showing that many VLMs can maintain an environment inventory when the state is anchored in text rather than raw visual memory.

\paragraph{\textbf{\emph{Models can infer \emph{when}, but cannot point to \emph{where}.}}}
Even with perfect historical text, grounding remains a bottleneck.
Temporal localization (TL) rises substantially (InternVL3 $\approx 0.68$; Qwen3-VL $\approx 0.67$),
yet multi-frame visual grounding (VGL-M) stays at $0.00$--$0.01$.
This reveals a structural disconnect between temporal inference and spatial projection back onto image coordinates.

\paragraph{\textbf{\emph{Continuity is the real wall.}}}
Relations and Tracking (RT), especially migration-path reconstruction (STT), remains the hardest dimension.
Most models record near-zero STT, failing to reconstruct trajectories across containers.
InternVL3 \cite{chen2025internvl3} is the sole model with non-trivial progress (0.15--0.16), suggesting an emerging ability to model object identity continuity rather than isolated historical facts.

\paragraph{\textbf{\emph{Event perception does not compose into global state.}}}
Comparing short-term event detection (SED) with cumulative state reconstruction (CSR) exposes a consistent integration gap.
Qwen3-VL \cite{qwen2025qwen3vl} performs strongly in short-term event detection (SED Add/Rem $\approx 0.69$),
but CSR is disproportionately lower, indicating that long-horizon failure is driven by insufficient state-update mechanisms rather than missing local change signals.

\begin{table*}[t]
\centering
\caption{Results on Short-Term (Immediate) Temporal Memory Tasks. Metrics are presented as Mean $\pm$ SD. Task categories SI, STL, and RT correspond to Semantic \& Inventory, Spatio-Temporal Localization, and Relations \& Tracking respectively.}
\label{tab:short_term_results}
\resizebox{\textwidth}{!}{
\begin{tabular}{l | cc | cc | cc | cc | cc}
\toprule
\rowcolor{spaCyanLight}
\textbf{Model} & \multicolumn{2}{c|}{\textbf{SI}} & \multicolumn{2}{c|}{\textbf{STL}} & \multicolumn{2}{c|}{\textbf{RT}} & \multicolumn{2}{c|}{\textbf{SED}} & \multicolumn{2}{c}{\textbf{CSR}} \\
\rowcolor{spaCyanLight}
\multicolumn{1}{l|}{} & SOR-M $\uparrow$ & LOR $\uparrow$ & VGL-M $\uparrow$ & TL $\uparrow$ & RSR-M $\uparrow$ & STT $\uparrow$ & Add $\uparrow$ & Rem $\uparrow$ & Add $\uparrow$ & Rem $\uparrow$ \\
\midrule

\rowcolor{spaCyanMid}
\multicolumn{11}{l}{\textit{\textbf{InternVL series}}} \\

InternVL2 & 
0.85 {\scriptsize $\pm$ 0.24} & 0.73 {\scriptsize $\pm$ 0.29} & 
0.01 {\scriptsize $\pm$ 0.02} & 0.51 {\scriptsize $\pm$ 0.35} & 
0.37 {\scriptsize $\pm$ 0.40} & 0.00 {\scriptsize $\pm$ 0.00} & 
0.31 {\scriptsize $\pm$ 0.44} & 0.64 {\scriptsize $\pm$ 0.48} & 
0.33 {\scriptsize $\pm$ 0.33} & 0.40 {\scriptsize $\pm$ 0.47} \\

InternVL2.5 & 
0.92 {\scriptsize $\pm$ 0.20} & 0.79 {\scriptsize $\pm$ 0.25} & 
0.00 {\scriptsize $\pm$ 0.01} & 0.63 {\scriptsize $\pm$ 0.36} & 
0.26 {\scriptsize $\pm$ 0.39} & 0.00 {\scriptsize $\pm$ 0.04} & 
0.55 {\scriptsize $\pm$ 0.49} & 0.65 {\scriptsize $\pm$ 0.48} & 
0.36 {\scriptsize $\pm$ 0.31} & 0.46 {\scriptsize $\pm$ 0.50} \\

InternVL3 & 
0.90 {\scriptsize $\pm$ 0.23} & 0.65 {\scriptsize $\pm$ 0.40} & 
0.00 {\scriptsize $\pm$ 0.01} & 0.68 {\scriptsize $\pm$ 0.36} & 
0.29 {\scriptsize $\pm$ 0.40} & 0.16 {\scriptsize $\pm$ 0.28} & 
0.63 {\scriptsize $\pm$ 0.47} & 0.68 {\scriptsize $\pm$ 0.46} & 
0.39 {\scriptsize $\pm$ 0.30} & 0.47 {\scriptsize $\pm$ 0.50} \\

\rowcolor{spaCyanMid}
\multicolumn{11}{l}{\textit{\textbf{LLaVA series}}} \\

LLaVA-NeXT & 
0.80 {\scriptsize $\pm$ 0.28} & 0.73 {\scriptsize $\pm$ 0.32} & 
0.00 {\scriptsize $\pm$ 0.00} & 0.42 {\scriptsize $\pm$ 0.38} & 
0.39 {\scriptsize $\pm$ 0.37} & 0.00 {\scriptsize $\pm$ 0.00} & 
0.51 {\scriptsize $\pm$ 0.48} & 0.56 {\scriptsize $\pm$ 0.49} & 
0.36 {\scriptsize $\pm$ 0.29} & 0.38 {\scriptsize $\pm$ 0.47} \\

LLaVA-OneVision & 
0.86 {\scriptsize $\pm$ 0.25} & 0.77 {\scriptsize $\pm$ 0.31} & 
0.00 {\scriptsize $\pm$ 0.00} & 0.40 {\scriptsize $\pm$ 0.33} & 
0.22 {\scriptsize $\pm$ 0.33} & 0.09 {\scriptsize $\pm$ 0.18} & 
0.28 {\scriptsize $\pm$ 0.44} & 0.64 {\scriptsize $\pm$ 0.48} & 
0.35 {\scriptsize $\pm$ 0.32} & 0.45 {\scriptsize $\pm$ 0.50} \\

\rowcolor{spaCyanMid}
\multicolumn{11}{l}{\textit{\textbf{Qwen-VL series}}} \\

Qwen2-VL & 
0.86 {\scriptsize $\pm$ 0.26} & 0.73 {\scriptsize $\pm$ 0.35} & 
0.00 {\scriptsize $\pm$ 0.00} & 0.45 {\scriptsize $\pm$ 0.37} & 
0.12 {\scriptsize $\pm$ 0.30} & 0.09 {\scriptsize $\pm$ 0.19} & 
0.52 {\scriptsize $\pm$ 0.48} & 0.64 {\scriptsize $\pm$ 0.48} & 
0.42 {\scriptsize $\pm$ 0.34} & 0.46 {\scriptsize $\pm$ 0.50} \\

Qwen2.5-VL & 
0.86 {\scriptsize $\pm$ 0.29} & 0.78 {\scriptsize $\pm$ 0.30} & 
0.00 {\scriptsize $\pm$ 0.02} & 0.64 {\scriptsize $\pm$ 0.35} & 
0.20 {\scriptsize $\pm$ 0.39} & 0.03 {\scriptsize $\pm$ 0.18} & 
0.66 {\scriptsize $\pm$ 0.46} & 0.65 {\scriptsize $\pm$ 0.46} & 
0.49 {\scriptsize $\pm$ 0.31} & 0.46 {\scriptsize $\pm$ 0.48} \\

Qwen3-VL & 
0.91 {\scriptsize $\pm$ 0.22} & 0.41 {\scriptsize $\pm$ 0.48} & 
0.01 {\scriptsize $\pm$ 0.02} & 0.63 {\scriptsize $\pm$ 0.42} & 
0.48 {\scriptsize $\pm$ 0.41} & 0.06 {\scriptsize $\pm$ 0.20} & 
0.69 {\scriptsize $\pm$ 0.46} & 0.69 {\scriptsize $\pm$ 0.45} & 
0.32 {\scriptsize $\pm$ 0.39} & 0.46 {\scriptsize $\pm$ 0.47} \\

\bottomrule
\end{tabular}
}
\end{table*}
\begin{table*}[t]
\centering
\caption{Evaluation Results on Long-Term Temporal Memory Tasks. Metrics are presented as Mean $\pm$ SD. Task categories SI, STL, and RT correspond to Semantic \& Inventory, Spatio-Temporal Localization, and Relations \& Tracking respectively.}
\label{tab:long_term_results}
\vspace{-8pt}
\resizebox{\textwidth}{!}{
\begin{tabular}{l | cc | cc | cc | cc | cc}
\toprule
\rowcolor{spaCyanLight}
\textbf{Model} & \multicolumn{2}{c|}{\textbf{SI}} & \multicolumn{2}{c|}{\textbf{STL}} & \multicolumn{2}{c|}{\textbf{RT}} & \multicolumn{2}{c|}{\textbf{SED}} & \multicolumn{2}{c}{\textbf{CSR}} \\
\rowcolor{spaCyanLight}
\multicolumn{1}{l|}{} & SOR-M $\uparrow$ & LOR $\uparrow$ & VGL-M $\uparrow$ & TL $\uparrow$ & RSR-M $\uparrow$ & STT $\uparrow$ & Add $\uparrow$ & Rem $\uparrow$ & Add $\uparrow$ & Rem $\uparrow$ \\
\midrule

\rowcolor{spaCyanMid}
\multicolumn{11}{l}{\textit{\textbf{InternVL series}}} \\

InternVL2 & 
0.85 {\scriptsize $\pm$ 0.24} & 0.73 {\scriptsize $\pm$ 0.29} & 
0.01 {\scriptsize $\pm$ 0.02} & 0.51 {\scriptsize $\pm$ 0.35} & 
0.37 {\scriptsize $\pm$ 0.40} & 0.00 {\scriptsize $\pm$ 0.00} & 
0.31 {\scriptsize $\pm$ 0.44} & 0.64 {\scriptsize $\pm$ 0.48} & 
0.33 {\scriptsize $\pm$ 0.33} & 0.40 {\scriptsize $\pm$ 0.47} \\

InternVL2.5 & 
0.92 {\scriptsize $\pm$ 0.20} & 0.79 {\scriptsize $\pm$ 0.25} & 
0.00 {\scriptsize $\pm$ 0.01} & 0.63 {\scriptsize $\pm$ 0.36} & 
0.26 {\scriptsize $\pm$ 0.38} & 0.00 {\scriptsize $\pm$ 0.04} & 
0.55 {\scriptsize $\pm$ 0.49} & 0.65 {\scriptsize $\pm$ 0.48} & 
0.37 {\scriptsize $\pm$ 0.31} & 0.46 {\scriptsize $\pm$ 0.50} \\

InternVL3 & 
0.90 {\scriptsize $\pm$ 0.23} & 0.64 {\scriptsize $\pm$ 0.40} & 
0.00 {\scriptsize $\pm$ 0.01} & 0.68 {\scriptsize $\pm$ 0.35} & 
0.29 {\scriptsize $\pm$ 0.39} & 0.15 {\scriptsize $\pm$ 0.27} & 
0.63 {\scriptsize $\pm$ 0.47} & 0.68 {\scriptsize $\pm$ 0.46} & 
0.39 {\scriptsize $\pm$ 0.30} & 0.47 {\scriptsize $\pm$ 0.50} \\

\rowcolor{spaCyanMid}
\multicolumn{11}{l}{\textit{\textbf{LLaVA series}}} \\

LLaVA-NeXT & 
0.75 {\scriptsize $\pm$ 0.14} & 0.79 {\scriptsize $\pm$ 0.28} & 
0.00 {\scriptsize $\pm$ 0.00} & 0.46 {\scriptsize $\pm$ 0.33} & 
0.41 {\scriptsize $\pm$ 0.24} & 0.00 {\scriptsize $\pm$ 0.00} & 
- & - & 
0.40 {\scriptsize $\pm$ 0.29} & 0.35 {\scriptsize $\pm$ 0.45} \\

LLaVA-OneVision & 
0.82 {\scriptsize $\pm$ 0.10} & 0.44 {\scriptsize $\pm$ 0.44} & 
0.00 {\scriptsize $\pm$ 0.00} & 0.19 {\scriptsize $\pm$ 0.28} & 
0.24 {\scriptsize $\pm$ 0.20} & 0.01 {\scriptsize $\pm$ 0.04} & 
- & - & 
0.13 {\scriptsize $\pm$ 0.18} & 0.23 {\scriptsize $\pm$ 0.42} \\

\rowcolor{spaCyanMid}
\multicolumn{11}{l}{\textit{\textbf{Qwen-VL series}}} \\

Qwen2-VL & 
0.78 {\scriptsize $\pm$ 0.10} & 0.33 {\scriptsize $\pm$ 0.39} & 
0.00 {\scriptsize $\pm$ 0.00} & 0.64 {\scriptsize $\pm$ 0.30} & 
0.05 {\scriptsize $\pm$ 0.06} & 0.20 {\scriptsize $\pm$ 0.22} & 
- & - & 
0.21 {\scriptsize $\pm$ 0.21} & 0.36 {\scriptsize $\pm$ 0.50} \\

Qwen2.5-VL & 
0.84 {\scriptsize $\pm$ 0.12} & 0.71 {\scriptsize $\pm$ 0.33} & 
0.00 {\scriptsize $\pm$ 0.01} & 0.66 {\scriptsize $\pm$ 0.37} & 
0.25 {\scriptsize $\pm$ 0.21} & 0.07 {\scriptsize $\pm$ 0.22} & 
- & - & 
0.39 {\scriptsize $\pm$ 0.22} & 0.38 {\scriptsize $\pm$ 0.49} \\

Qwen3-VL & 
0.89 {\scriptsize $\pm$ 0.09} & 0.80 {\scriptsize $\pm$ 0.40} & 
0.01 {\scriptsize $\pm$ 0.01} & 0.67 {\scriptsize $\pm$ 0.38} & 
0.45 {\scriptsize $\pm$ 0.19} & 0.00 {\scriptsize $\pm$ 0.00} & 
- & - & 
0.50 {\scriptsize $\pm$ 0.32} & 0.43 {\scriptsize $\pm$ 0.43} \\

\bottomrule
\end{tabular}
}
\vspace{-8pt}
\end{table*}

\subsection{Level 3: Visual-Conditioned Temporal Memory}
\label{subsec:l3_results_analysis}

Level~3 is the most realistic and challenging setting: models must reason over raw visual sequences without textual summaries.
Results in Tables~\ref{tab:l3_short_rgb_results}--\ref{tab:l3_long_rgbd_results} show a large capability gap for long-horizon embodied memory.

\paragraph{\textbf{\emph{Removing text collapses ``memory''.}}}
Without symbolic history, SI performance decays sharply.
While L2 achieves SOR-M above 0.80 for strong models, L3 SOR-M drops to 0.07--0.19 across RGB and RGB-D settings.
In several long-term conditions, Qwen3-VL \cite{qwen2025qwen3vl} reaches LOR $=0.00$, indicating near-total failure to maintain a persistent inventory from visual streams alone.
This supports the conclusion that L2 ``memory'' often reflects reasoning over text rather than a robust internal visual world model.

\paragraph{\textbf{\emph{Trajectory tracking fails without symbolic hints.}}}
Migration-path tracking (STT) becomes a universal failure mode in L3: scores are consistently 0.00 across models and modalities.
Without explicit symbolic cues (e.g., ``object A was placed in container B''), current VLMs fail to exploit temporal visual consistency for instance-level tracking across occlusions and container interactions.

\paragraph{\textbf{\emph{Local change detection does not yield global state reconstruction.}}}
Models show comparatively stable short-term event detection, especially for removals (often around 0.60),
but this local success does not translate into accurate CSR.
For instance, InternVL3 \cite{chen2025internvl3} attains SED-Rem $\approx 0.61$ in RGB short-term but only $\approx 0.35$ on CSR-Rem,
indicating difficulty in accumulating events into a coherent evolving representation.

\paragraph{\textbf{\emph{Depth yields marginal gains, but does not fix episodic integration.}}}
Adding depth provides limited improvement for certain temporal metrics, yet the core bottlenecks in SI and RT persist:
SOR-M and STT remain largely stagnant.
This suggests the main limitation is not sensory insufficiency but a structural deficiency in episodic integration---linking observations across time and updating world state accordingly.

\begin{table*}[t]
\centering
\caption{Evaluation Results on Level 3 Short-Term (Immediate) Temporal Memory Tasks (RGB Only). Metrics are presented as Mean $\pm$ SD. Task categories SI, STL, and RT correspond to Semantic \& Inventory, Spatio-Temporal Localization, and Relations \& Tracking respectively.}
\vspace{-8pt}
\label{tab:l3_short_rgb_results}
\resizebox{\textwidth}{!}{
\begin{tabular}{l | cc | cc | cc | cc | cc}
\toprule
\rowcolor{spaCyanLight}
\textbf{Model} & \multicolumn{2}{c|}{\textbf{SI}} & \multicolumn{2}{c|}{\textbf{STL}} & \multicolumn{2}{c|}{\textbf{RT}} & \multicolumn{2}{c|}{\textbf{SED}} & \multicolumn{2}{c}{\textbf{CSR}} \\
\rowcolor{spaCyanLight}
\multicolumn{1}{l|}{} & SOR-M $\uparrow$ & LOR $\uparrow$ & VGL-M $\uparrow$ & TL $\uparrow$ & RSR-M $\uparrow$ & STT $\uparrow$ & Add $\uparrow$ & Rem $\uparrow$ & Add $\uparrow$ & Rem $\uparrow$ \\
\midrule

\rowcolor{spaCyanMid}
\multicolumn{11}{l}{\textit{\textbf{InternVL series}}} \\

InternVL2 & 
0.09 {\scriptsize $\pm$ 0.09} & 0.17 {\scriptsize $\pm$ 0.09} & 
0.02 {\scriptsize $\pm$ 0.05} & 0.52 {\scriptsize $\pm$ 0.33} & 
0.03 {\scriptsize $\pm$ 0.13} & 0.00 {\scriptsize $\pm$ 0.00} & 
0.36 {\scriptsize $\pm$ 0.46} & 0.59 {\scriptsize $\pm$ 0.49} & 
0.10 {\scriptsize $\pm$ 0.15} & 0.33 {\scriptsize $\pm$ 0.47} \\

InternVL2.5 & 
0.14 {\scriptsize $\pm$ 0.19} & 0.17 {\scriptsize $\pm$ 0.14} & 
0.00 {\scriptsize $\pm$ 0.00} & 0.38 {\scriptsize $\pm$ 0.34} & 
0.02 {\scriptsize $\pm$ 0.07} & 0.00 {\scriptsize $\pm$ 0.00} & 
0.45 {\scriptsize $\pm$ 0.50} & 0.62 {\scriptsize $\pm$ 0.49} & 
0.11 {\scriptsize $\pm$ 0.23} & 0.35 {\scriptsize $\pm$ 0.48} \\

InternVL3 & 
0.13 {\scriptsize $\pm$ 0.17} & 0.19 {\scriptsize $\pm$ 0.13} & 
0.00 {\scriptsize $\pm$ 0.00} & 0.42 {\scriptsize $\pm$ 0.35} & 
0.07 {\scriptsize $\pm$ 0.15} & 0.00 {\scriptsize $\pm$ 0.00} & 
0.43 {\scriptsize $\pm$ 0.50} & 0.61 {\scriptsize $\pm$ 0.49} & 
0.19 {\scriptsize $\pm$ 0.25} & 0.35 {\scriptsize $\pm$ 0.48} \\

\rowcolor{spaCyanMid}
\multicolumn{11}{l}{\textit{\textbf{LLaVA series}}} \\

LLaVA-NeXT & 
0.12 {\scriptsize $\pm$ 0.30} & 0.13 {\scriptsize $\pm$ 0.07} & 
0.00 {\scriptsize $\pm$ 0.00} & 0.17 {\scriptsize $\pm$ 0.25} & 
0.02 {\scriptsize $\pm$ 0.11} & 0.00 {\scriptsize $\pm$ 0.00} & 
0.22 {\scriptsize $\pm$ 0.40} & 0.58 {\scriptsize $\pm$ 0.50} & 
0.39 {\scriptsize $\pm$ 0.32} & 0.33 {\scriptsize $\pm$ 0.47} \\

LLaVA-OneVision & 
0.14 {\scriptsize $\pm$ 0.32} & 0.10 {\scriptsize $\pm$ 0.12} & 
0.00 {\scriptsize $\pm$ 0.00} & 0.18 {\scriptsize $\pm$ 0.23} & 
0.04 {\scriptsize $\pm$ 0.14} & 0.00 {\scriptsize $\pm$ 0.00} & 
0.34 {\scriptsize $\pm$ 0.47} & 0.61 {\scriptsize $\pm$ 0.49} & 
0.21 {\scriptsize $\pm$ 0.29} & 0.35 {\scriptsize $\pm$ 0.48} \\

\rowcolor{spaCyanMid}
\multicolumn{11}{l}{\textit{\textbf{Qwen-VL series}}} \\

Qwen2-VL & 
0.14 {\scriptsize $\pm$ 0.18} & 0.07 {\scriptsize $\pm$ 0.11} & 
0.00 {\scriptsize $\pm$ 0.00} & 0.19 {\scriptsize $\pm$ 0.29} & 
0.01 {\scriptsize $\pm$ 0.10} & 0.00 {\scriptsize $\pm$ 0.00} & 
0.34 {\scriptsize $\pm$ 0.47} & 0.61 {\scriptsize $\pm$ 0.49} & 
0.15 {\scriptsize $\pm$ 0.28} & 0.35 {\scriptsize $\pm$ 0.48} \\

Qwen2.5-VL & 
0.10 {\scriptsize $\pm$ 0.16} & 0.22 {\scriptsize $\pm$ 0.16} & 
0.02 {\scriptsize $\pm$ 0.08} & 0.38 {\scriptsize $\pm$ 0.34} & 
0.00 {\scriptsize $\pm$ 0.00} & 0.00 {\scriptsize $\pm$ 0.00} & 
0.48 {\scriptsize $\pm$ 0.50} & 0.61 {\scriptsize $\pm$ 0.49} & 
0.21 {\scriptsize $\pm$ 0.24} & 0.34 {\scriptsize $\pm$ 0.47} \\

Qwen3-VL & 
0.17 {\scriptsize $\pm$ 0.13} & 0.00 {\scriptsize $\pm$ 0.02} & 
0.01 {\scriptsize $\pm$ 0.03} & 0.11 {\scriptsize $\pm$ 0.24} & 
0.08 {\scriptsize $\pm$ 0.14} & 0.00 {\scriptsize $\pm$ 0.00} & 
0.42 {\scriptsize $\pm$ 0.50} & 0.61 {\scriptsize $\pm$ 0.49} & 
0.03 {\scriptsize $\pm$ 0.17} & 0.34 {\scriptsize $\pm$ 0.47} \\

\bottomrule
\end{tabular}
}
\vspace{-8pt}
\end{table*}
\begin{table*}[t]
\centering
\caption{Evaluation Results on Level 3 Short-Term (Immediate) Temporal Memory Tasks (RGB+Depth). Metrics are presented as Mean $\pm$ SD. Task categories SI, STL, and RT correspond to Semantic \& Inventory, Spatio-Temporal Localization, and Relations \& Tracking respectively.}
\vspace{-8pt}
\label{tab:l3_short_rgbd_results}
\resizebox{\textwidth}{!}{
\begin{tabular}{l | cc | cc | cc | cc | cc}
\toprule
\rowcolor{spaCyanLight}
\textbf{Model} & \multicolumn{2}{c|}{\textbf{SI}} & \multicolumn{2}{c|}{\textbf{STL}} & \multicolumn{2}{c|}{\textbf{RT}} & \multicolumn{2}{c|}{\textbf{SED}} & \multicolumn{2}{c}{\textbf{CSR}} \\
\rowcolor{spaCyanLight}
\multicolumn{1}{l|}{} & SOR-M $\uparrow$ & LOR $\uparrow$ & VGL-M $\uparrow$ & TL $\uparrow$ & RSR-M $\uparrow$ & STT $\uparrow$ & Add $\uparrow$ & Rem $\uparrow$ & Add $\uparrow$ & Rem $\uparrow$ \\
\midrule

\rowcolor{spaCyanMid}
\multicolumn{11}{l}{\textit{\textbf{InternVL series}}} \\

InternVL2 & 
0.09 {\scriptsize $\pm$ 0.10} & 0.17 {\scriptsize $\pm$ 0.10} & 
0.00 {\scriptsize $\pm$ 0.03} & 0.51 {\scriptsize $\pm$ 0.33} & 
0.03 {\scriptsize $\pm$ 0.12} & 0.00 {\scriptsize $\pm$ 0.00} & 
0.35 {\scriptsize $\pm$ 0.46} & 0.61 {\scriptsize $\pm$ 0.49} & 
0.11 {\scriptsize $\pm$ 0.16} & 0.34 {\scriptsize $\pm$ 0.47} \\

InternVL2.5 & 
0.12 {\scriptsize $\pm$ 0.14} & 0.15 {\scriptsize $\pm$ 0.13} & 
0.00 {\scriptsize $\pm$ 0.00} & 0.44 {\scriptsize $\pm$ 0.33} & 
0.00 {\scriptsize $\pm$ 0.03} & 0.00 {\scriptsize $\pm$ 0.00} & 
0.43 {\scriptsize $\pm$ 0.49} & 0.61 {\scriptsize $\pm$ 0.49} & 
0.13 {\scriptsize $\pm$ 0.23} & 0.35 {\scriptsize $\pm$ 0.48} \\

InternVL3 & 
0.13 {\scriptsize $\pm$ 0.16} & 0.16 {\scriptsize $\pm$ 0.14} & 
0.00 {\scriptsize $\pm$ 0.01} & 0.29 {\scriptsize $\pm$ 0.31} & 
0.07 {\scriptsize $\pm$ 0.13} & 0.00 {\scriptsize $\pm$ 0.00} & 
0.40 {\scriptsize $\pm$ 0.49} & 0.61 {\scriptsize $\pm$ 0.49} & 
0.15 {\scriptsize $\pm$ 0.21} & 0.35 {\scriptsize $\pm$ 0.48} \\

\rowcolor{spaCyanMid}
\multicolumn{11}{l}{\textit{\textbf{LLaVA series}}} \\

LLaVA-NeXT & 
0.12 {\scriptsize $\pm$ 0.30} & 0.12 {\scriptsize $\pm$ 0.05} & 
0.00 {\scriptsize $\pm$ 0.00} & 0.15 {\scriptsize $\pm$ 0.24} & 
0.01 {\scriptsize $\pm$ 0.10} & 0.00 {\scriptsize $\pm$ 0.00} & 
0.22 {\scriptsize $\pm$ 0.40} & 0.60 {\scriptsize $\pm$ 0.49} & 
0.42 {\scriptsize $\pm$ 0.31} & 0.34 {\scriptsize $\pm$ 0.47} \\

LLaVA-OneVision & 
0.14 {\scriptsize $\pm$ 0.34} & 0.06 {\scriptsize $\pm$ 0.08} & 
0.00 {\scriptsize $\pm$ 0.00} & 0.19 {\scriptsize $\pm$ 0.26} & 
0.02 {\scriptsize $\pm$ 0.07} & 0.00 {\scriptsize $\pm$ 0.00} & 
0.32 {\scriptsize $\pm$ 0.47} & 0.61 {\scriptsize $\pm$ 0.49} & 
0.20 {\scriptsize $\pm$ 0.28} & 0.35 {\scriptsize $\pm$ 0.48} \\

\rowcolor{spaCyanMid}
\multicolumn{11}{l}{\textit{\textbf{Qwen-VL series}}} \\

Qwen2-VL & 
0.17 {\scriptsize $\pm$ 0.22} & 0.11 {\scriptsize $\pm$ 0.14} & 
0.00 {\scriptsize $\pm$ 0.00} & 0.17 {\scriptsize $\pm$ 0.27} & 
0.01 {\scriptsize $\pm$ 0.10} & 0.00 {\scriptsize $\pm$ 0.00} & 
0.31 {\scriptsize $\pm$ 0.44} & 0.61 {\scriptsize $\pm$ 0.49} & 
0.12 {\scriptsize $\pm$ 0.23} & 0.35 {\scriptsize $\pm$ 0.48} \\

Qwen2.5-VL & 
0.11 {\scriptsize $\pm$ 0.17} & 0.17 {\scriptsize $\pm$ 0.12} & 
0.01 {\scriptsize $\pm$ 0.04} & 0.33 {\scriptsize $\pm$ 0.34} & 
0.00 {\scriptsize $\pm$ 0.00} & 0.00 {\scriptsize $\pm$ 0.00} & 
0.43 {\scriptsize $\pm$ 0.50} & 0.61 {\scriptsize $\pm$ 0.49} & 
0.25 {\scriptsize $\pm$ 0.29} & 0.35 {\scriptsize $\pm$ 0.48} \\

Qwen3-VL & 
0.16 {\scriptsize $\pm$ 0.14} & 0.00 {\scriptsize $\pm$ 0.00} & 
0.02 {\scriptsize $\pm$ 0.09} & 0.07 {\scriptsize $\pm$ 0.16} & 
0.07 {\scriptsize $\pm$ 0.14} & 0.00 {\scriptsize $\pm$ 0.00} & 
0.41 {\scriptsize $\pm$ 0.49} & 0.61 {\scriptsize $\pm$ 0.49} & 
0.03 {\scriptsize $\pm$ 0.18} & 0.35 {\scriptsize $\pm$ 0.48} \\

\bottomrule
\end{tabular}
}
\vspace{-8pt}
\end{table*}
\begin{table*}[t]
\centering
\caption{Results on Level 3 Long-Term (Episodic) Temporal Memory tasks (RGB only). Metrics are reported as Mean $\pm$ SD. SI, STL, and RT denote Semantic \& Inventory, Spatio-Temporal Localization, and Relations \& Tracking, respectively.}
\label{tab:l3_long_rgb_results}
\resizebox{\textwidth}{!}{
\begin{tabular}{l | cc | cc | cc | cc | cc}
\toprule
\rowcolor{spaCyanLight}
\textbf{Model} & \multicolumn{2}{c|}{\textbf{SI}} & \multicolumn{2}{c|}{\textbf{STL}} & \multicolumn{2}{c|}{\textbf{RT}} & \multicolumn{2}{c|}{\textbf{SED}} & \multicolumn{2}{c}{\textbf{CSR}} \\
\rowcolor{spaCyanLight}
\multicolumn{1}{l|}{} & SOR-M $\uparrow$ & LOR $\uparrow$ & VGL-M $\uparrow$ & TL $\uparrow$ & RSR-M $\uparrow$ & STT $\uparrow$ & Add $\uparrow$ & Rem $\uparrow$ & Add $\uparrow$ & Rem $\uparrow$ \\
\midrule

\rowcolor{spaCyanMid}
\multicolumn{11}{l}{\textit{\textbf{InternVL series}}} \\

InternVL2 & 
0.09 {\scriptsize $\pm$ 0.09} & 0.17 {\scriptsize $\pm$ 0.09} & 
0.02 {\scriptsize $\pm$ 0.05} & 0.53 {\scriptsize $\pm$ 0.33} & 
0.03 {\scriptsize $\pm$ 0.12} & 0.00 {\scriptsize $\pm$ 0.00} & 
0.36 {\scriptsize $\pm$ 0.46} & 0.59 {\scriptsize $\pm$ 0.49} & 
0.11 {\scriptsize $\pm$ 0.15} & 0.33 {\scriptsize $\pm$ 0.47}\\

InternVL2.5 & 
0.14 {\scriptsize $\pm$ 0.19} & 0.17 {\scriptsize $\pm$ 0.13} & 
0.00 {\scriptsize $\pm$ 0.00} & 0.37 {\scriptsize $\pm$ 0.34} & 
0.02 {\scriptsize $\pm$ 0.07} & 0.00 {\scriptsize $\pm$ 0.00} & 
0.45 {\scriptsize $\pm$ 0.50} & 0.62 {\scriptsize $\pm$ 0.49} & 
0.11 {\scriptsize $\pm$ 0.23} & 0.35 {\scriptsize $\pm$ 0.48}\\

InternVL3 & 
0.13 {\scriptsize $\pm$ 0.16} & 0.19 {\scriptsize $\pm$ 0.13} & 
0.00 {\scriptsize $\pm$ 0.00} & 0.42 {\scriptsize $\pm$ 0.36} & 
0.07 {\scriptsize $\pm$ 0.15} & 0.00 {\scriptsize $\pm$ 0.00} & 
0.43 {\scriptsize $\pm$ 0.50} & 0.61 {\scriptsize $\pm$ 0.49} & 
0.18 {\scriptsize $\pm$ 0.24} & 0.35 {\scriptsize $\pm$ 0.48}\\

\rowcolor{spaCyanMid}
\multicolumn{11}{l}{\textit{\textbf{LLaVA series}}} \\

LLaVA-NeXT & 
0.10 {\scriptsize $\pm$ 0.07} & 0.11 {\scriptsize $\pm$ 0.03} & 
0.00 {\scriptsize $\pm$ 0.00} & 0.60 {\scriptsize $\pm$ 0.34} & 
0.00 {\scriptsize $\pm$ 0.00} & 0.00 {\scriptsize $\pm$ 0.00} & 
- & - & 
0.30 {\scriptsize $\pm$ 0.13} & 0.33 {\scriptsize $\pm$ 0.58}\\

LLaVA-OneVision & 
0.13 {\scriptsize $\pm$ 0.15} & 0.07 {\scriptsize $\pm$ 0.06} & 
0.00 {\scriptsize $\pm$ 0.00} & 0.01 {\scriptsize $\pm$ 0.01} & 
0.04 {\scriptsize $\pm$ 0.07} & 0.00 {\scriptsize $\pm$ 0.00} & 
- & - & 
0.24 {\scriptsize $\pm$ 0.21} & 0.33 {\scriptsize $\pm$ 0.58}\\

\rowcolor{spaCyanMid}
\multicolumn{11}{l}{\textit{\textbf{Qwen-VL series}}} \\

Qwen2-VL & 
0.16 {\scriptsize $\pm$ 0.04} & 0.17 {\scriptsize $\pm$ 0.22} & 
0.00 {\scriptsize $\pm$ 0.00} & 0.43 {\scriptsize $\pm$ 0.50} & 
0.02 {\scriptsize $\pm$ 0.03} & 0.00 {\scriptsize $\pm$ 0.00} & 
- & - & 
0.11 {\scriptsize $\pm$ 0.19} & 0.33 {\scriptsize $\pm$ 0.58}\\

Qwen2.5-VL & 
0.07 {\scriptsize $\pm$ 0.03} & 0.09 {\scriptsize $\pm$ 0.09} & 
0.01 {\scriptsize $\pm$ 0.01} & 0.51 {\scriptsize $\pm$ 0.41} & 
0.00 {\scriptsize $\pm$ 0.00} & 0.00 {\scriptsize $\pm$ 0.00} & 
- & - & 
0.22 {\scriptsize $\pm$ 0.19} & 0.33 {\scriptsize $\pm$ 0.58}\\

Qwen3-VL & 
0.19 {\scriptsize $\pm$ 0.07} & 0.00 {\scriptsize $\pm$ 0.00} & 
0.01 {\scriptsize $\pm$ 0.01} & 0.00 {\scriptsize $\pm$ 0.00} & 
0.13 {\scriptsize $\pm$ 0.12} & 0.00 {\scriptsize $\pm$ 0.00} & 
- & - & 
0.00 {\scriptsize $\pm$ 0.00} & 0.33 {\scriptsize $\pm$ 0.58}\\

\bottomrule
\end{tabular}
}
\vspace{-12pt}
\end{table*}
\begin{table*}[t]
\centering
\caption{Results on Level 3 long-term (episodic) temporal memory tasks (RGB+Depth). Metrics are reported as Mean $\pm$ SD. SI, STL, and RT denote Semantic \& Inventory, Spatio-Temporal Localization, and Relations \& Tracking, respectively.}
\vspace{-8pt}
\label{tab:l3_long_rgbd_results}
\resizebox{\textwidth}{!}{
\begin{tabular}{l | cc | cc | cc | cc | cc}
\toprule
\rowcolor{spaCyanLight}
\textbf{Model} & \multicolumn{2}{c|}{\textbf{SI}} & \multicolumn{2}{c|}{\textbf{STL}} & \multicolumn{2}{c|}{\textbf{RT}} & \multicolumn{2}{c|}{\textbf{SED}} & \multicolumn{2}{c}{\textbf{CSR}} \\
\rowcolor{spaCyanLight}
\multicolumn{1}{l|}{} & SOR-M $\uparrow$ & LOR $\uparrow$ & VGL-M $\uparrow$ & TL $\uparrow$ & RSR-M $\uparrow$ & STT $\uparrow$ & Add $\uparrow$ & Rem $\uparrow$ & Add $\uparrow$ & Rem $\uparrow$ \\
\midrule

\rowcolor{spaCyanMid}
\multicolumn{11}{l}{\textit{\textbf{InternVL series}}} \\

InternVL2 & 
0.09 {\scriptsize $\pm$ 0.09} & 0.16 {\scriptsize $\pm$ 0.10} & 
0.00 {\scriptsize $\pm$ 0.03} & 0.51 {\scriptsize $\pm$ 0.33} & 
0.03 {\scriptsize $\pm$ 0.12} & 0.00 {\scriptsize $\pm$ 0.00} & 
0.35 {\scriptsize $\pm$ 0.46} & 0.61 {\scriptsize $\pm$ 0.49} & 
0.11 {\scriptsize $\pm$ 0.16} & 0.34 {\scriptsize $\pm$ 0.47} \\

InternVL2.5 & 
0.12 {\scriptsize $\pm$ 0.14} & 0.15 {\scriptsize $\pm$ 0.13} & 
0.00 {\scriptsize $\pm$ 0.00} & 0.43 {\scriptsize $\pm$ 0.34} & 
0.00 {\scriptsize $\pm$ 0.03} & 0.00 {\scriptsize $\pm$ 0.00} & 
0.43 {\scriptsize $\pm$ 0.49} & 0.61 {\scriptsize $\pm$ 0.49} & 
0.13 {\scriptsize $\pm$ 0.23} & 0.35 {\scriptsize $\pm$ 0.48} \\

InternVL3 & 
0.13 {\scriptsize $\pm$ 0.16} & 0.16 {\scriptsize $\pm$ 0.14} & 
0.00 {\scriptsize $\pm$ 0.01} & 0.29 {\scriptsize $\pm$ 0.32} & 
0.07 {\scriptsize $\pm$ 0.13} & 0.00 {\scriptsize $\pm$ 0.00} & 
0.40 {\scriptsize $\pm$ 0.49} & 0.61 {\scriptsize $\pm$ 0.49} & 
0.15 {\scriptsize $\pm$ 0.21} & 0.35 {\scriptsize $\pm$ 0.48} \\

\rowcolor{spaCyanMid}
\multicolumn{11}{l}{\textit{\textbf{LLaVA series}}} \\

LLaVA-NeXT & 
0.13 {\scriptsize $\pm$ 0.13} & 0.07 {\scriptsize $\pm$ 0.06} & 
0.00 {\scriptsize $\pm$ 0.00} & 0.60 {\scriptsize $\pm$ 0.36} & 
0.02 {\scriptsize $\pm$ 0.03} & 0.00 {\scriptsize $\pm$ 0.00} & 
- & - & 
0.24 {\scriptsize $\pm$ 0.21} & 0.33 {\scriptsize $\pm$ 0.58} \\

LLaVA-OneVision & 
0.14 {\scriptsize $\pm$ 0.14} & 0.07 {\scriptsize $\pm$ 0.06} & 
0.00 {\scriptsize $\pm$ 0.00} & 0.01 {\scriptsize $\pm$ 0.01} & 
0.03 {\scriptsize $\pm$ 0.05} & 0.00 {\scriptsize $\pm$ 0.00} & 
- & - & 
0.24 {\scriptsize $\pm$ 0.21} & 0.33 {\scriptsize $\pm$ 0.58} \\

\rowcolor{spaCyanMid}
\multicolumn{11}{l}{\textit{\textbf{Qwen-VL series}}} \\

Qwen2-VL & 
0.18 {\scriptsize $\pm$ 0.04} & 0.05 {\scriptsize $\pm$ 0.09} & 
0.00 {\scriptsize $\pm$ 0.00} & 0.09 {\scriptsize $\pm$ 0.13} & 
0.02 {\scriptsize $\pm$ 0.03} & 0.00 {\scriptsize $\pm$ 0.00} & 
- & - & 
0.07 {\scriptsize $\pm$ 0.12} & 0.33 {\scriptsize $\pm$ 0.58} \\

Qwen2.5-VL & 
0.07 {\scriptsize $\pm$ 0.08} & 0.06 {\scriptsize $\pm$ 0.05} & 
0.00 {\scriptsize $\pm$ 0.00} & 0.50 {\scriptsize $\pm$ 0.50} & 
0.00 {\scriptsize $\pm$ 0.00} & 0.00 {\scriptsize $\pm$ 0.00} & 
- & - & 
0.11 {\scriptsize $\pm$ 0.19} & 0.33 {\scriptsize $\pm$ 0.58} \\

Qwen3-VL & 
0.16 {\scriptsize $\pm$ 0.06} & 0.00 {\scriptsize $\pm$ 0.00} & 
0.01 {\scriptsize $\pm$ 0.02} & 0.00 {\scriptsize $\pm$ 0.00} & 
0.12 {\scriptsize $\pm$ 0.10} & 0.00 {\scriptsize $\pm$ 0.00} & 
- & - & 
0.00 {\scriptsize $\pm$ 0.00} & 0.33 {\scriptsize $\pm$ 0.58} \\

\bottomrule
\end{tabular}
}
\vspace{-16pt}
\end{table*}

\subsection{Discussion: What SpaMEM Diagnoses}
\label{subsec:discussion}

Across L1--L3, SpaMEM exposes a consistent pattern: strong open-world semantics do not translate into embodied spatial competence,
and short-horizon perception does not naturally compose into long-horizon state tracking.
Rather than a single missing capability, the failure emerges as a \emph{stacked bottleneck} across grounding, state update, and identity continuity.

\paragraph{\textbf{\emph{(1) Spatial grounding is the first hard ceiling.}}}
L1 shows that coordinate-consistent localization remains nearly non-functional even when recognition is partially successful.
This bottleneck persists in L2 and L3: temporal reasoning can improve \emph{when} events occur, yet projection back to \emph{where} objects are in image coordinates remains unreliable.
As a result, downstream abilities that depend on spatial precision (e.g., tracking and trajectory reconstruction) are constrained from the start.

\paragraph{\textbf{\emph{(2) Text-conditioned ``memory'' overestimates visual world modeling.}}}
L2 substantially boosts inventory and semantic reasoning when provided with ground-truth textual history,
suggesting that many VLMs can perform accurate \emph{symbolic bookkeeping}.
However, the sharp collapse from L2 to L3 indicates that this competence does not imply a robust \emph{internal visual state}:
without explicit symbolic anchors, models struggle to maintain persistent inventories and to revise beliefs over time.

\paragraph{\textbf{\emph{(3) Identity continuity and belief revision are the real long-horizon wall.}}}
Even when short-term event detection is non-trivial, it does not reliably accumulate into coherent cumulative state reconstruction.
The most challenging dimension is Relations and Tracking (RT), where instance identity must persist through occlusions, container interactions, and migrations.
Near-zero trajectory reconstruction in L3 suggests that current VLMs lack mechanisms for stable identity binding and belief updating under partial observability.

\paragraph{\textbf{\emph{(4) Depth helps at the margins, but the core deficit is episodic integration.}}}
RGB-D yields occasional improvements in geometry-related signals and certain temporal metrics,
yet does not resolve the dominant failure modes in long-horizon inventory maintenance and tracking.
This points to a structural limitation in how models integrate observations across time,
rather than a purely sensory limitation that can be fixed by adding one more modality channel.
\section{High-Level Diagnostic Syntheses}
\label{sec:high_level_analysis}

\begin{figure*}[t]
    \centering

    \begin{subfigure}[t]{0.48\textwidth}
        \centering
        \includegraphics[width=\linewidth]{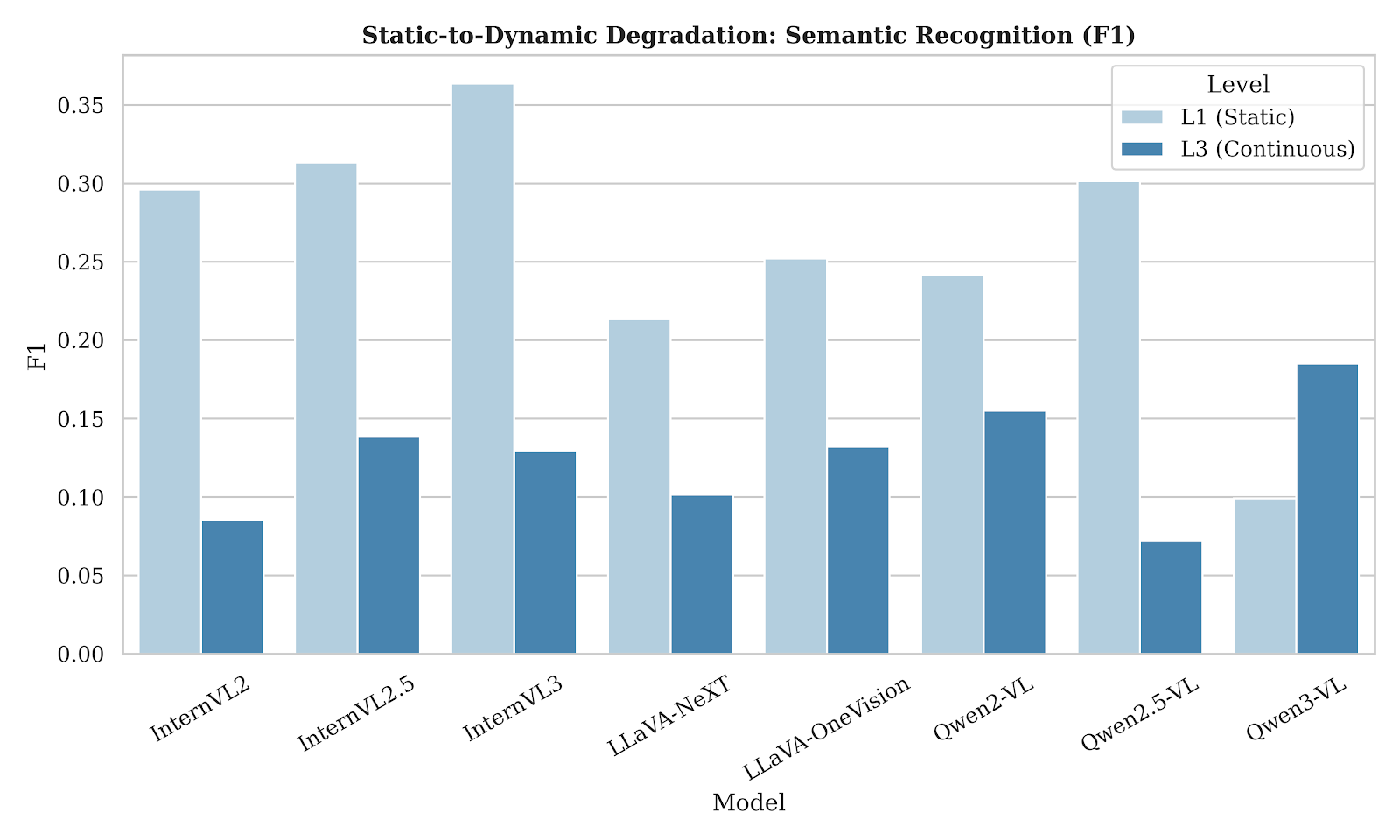}
        \caption{Static-to-Dynamic Degradation}
        \label{fig:hl_degradation}
    \end{subfigure}\hfill
    \begin{subfigure}[t]{0.48\textwidth}
        \centering
        \includegraphics[width=\linewidth]{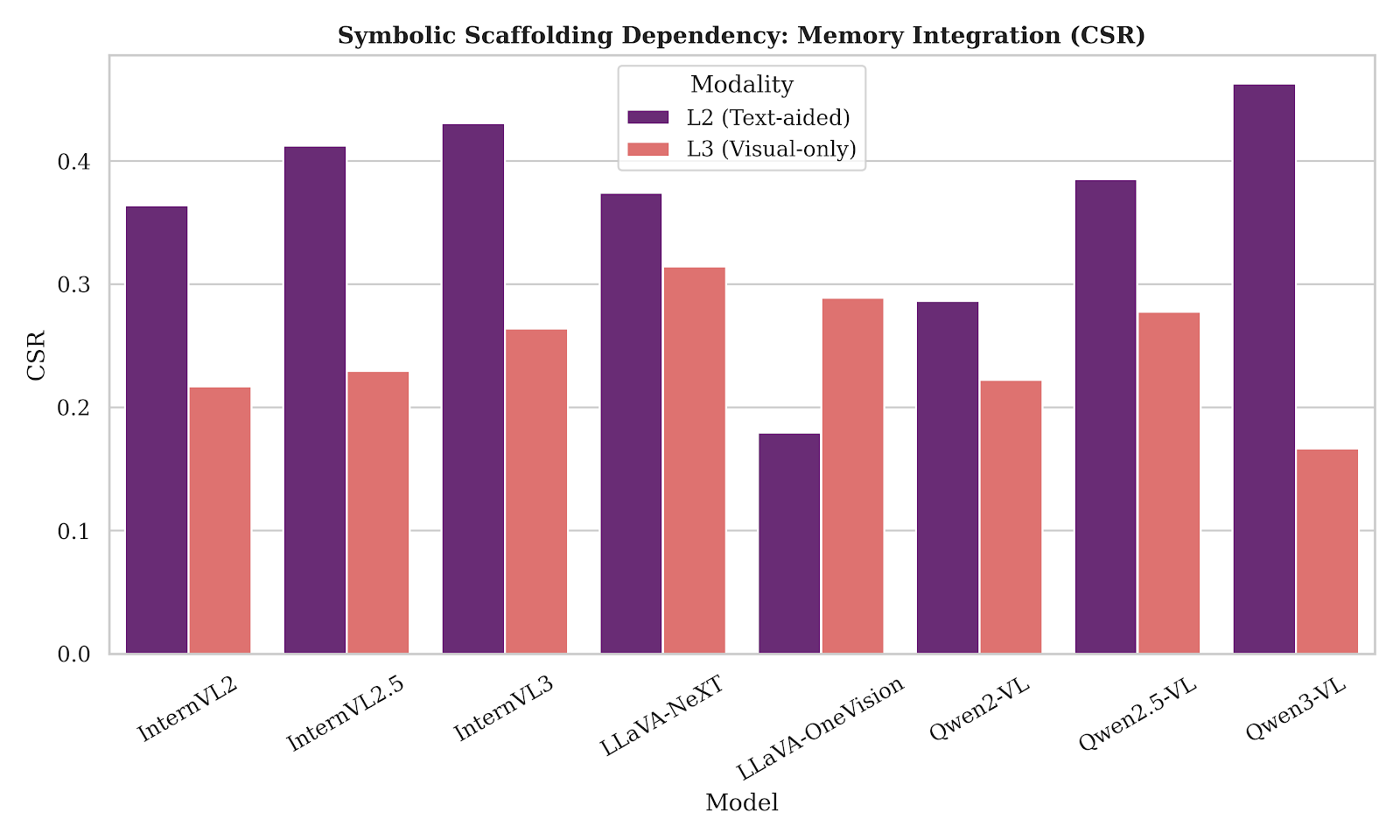}
        \caption{Symbolic Scaffolding Dependency}
        \label{fig:hl_scaffolding}
    \end{subfigure}

    \vspace{6pt}

    \begin{subfigure}[t]{0.48\textwidth}
        \centering
        \includegraphics[width=\linewidth]{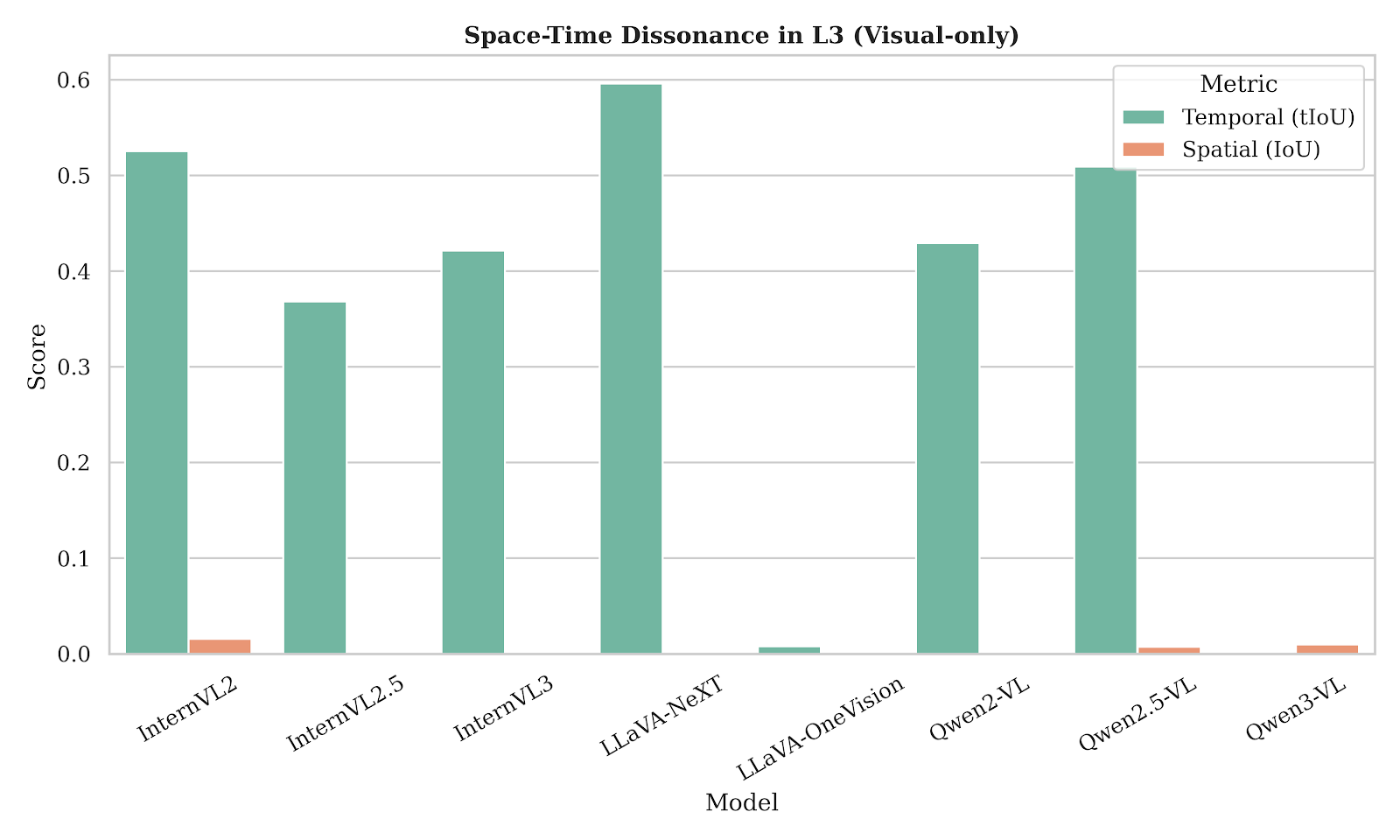}
        \caption{Space-Time Dissonance}
        \label{fig:hl_dissonance}
    \end{subfigure}\hfill
    \begin{subfigure}[t]{0.48\textwidth}
        \centering
        \includegraphics[width=\linewidth]{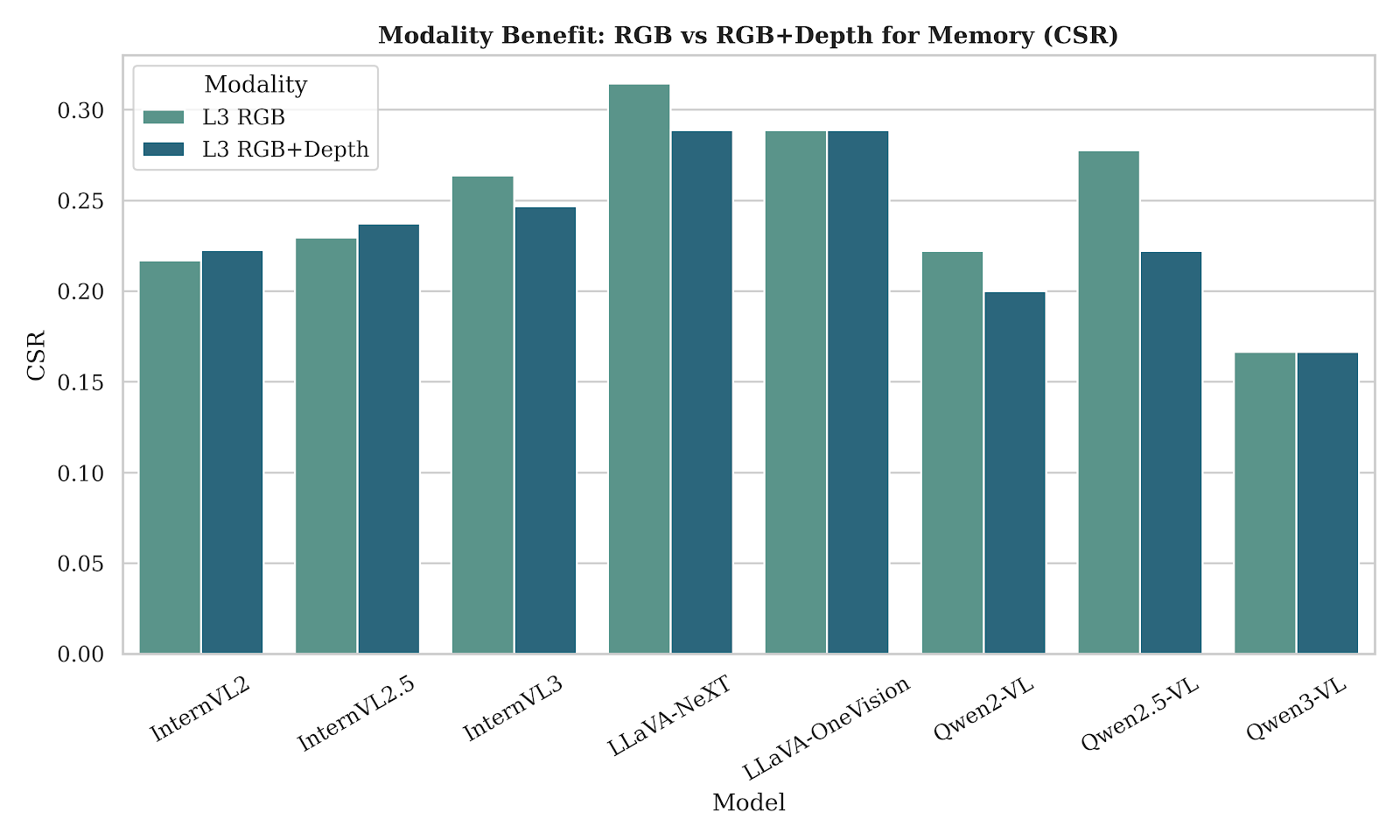}
        \caption{Modality Benefit (RGB vs. RGB-D)}
        \label{fig:hl_rgbd}
    \end{subfigure}

    \caption{
    \textbf{High-level diagnostic syntheses under SpaMEM.}
    Four consistent bottlenecks emerge across VLMs:
    (a) severe performance degradation when transitioning from static perception to continuous embodied streams,
    (b) strong symbolic scaffolding dependency when textual history is removed,
    (c) a fundamental space--time dissonance where models track temporal order but fail spatial grounding,
    and (d) negligible benefits from depth information (RGB-D) for long-horizon spatial integration.
    }
    \vspace{-16pt}
    \label{fig:highlevel_panel}
\end{figure*}

By synthesizing the performance of SoTA VLMs across the three modalities of SpaMEM (Static, Text-aided, and Visual-only), we identify four critical bottlenecks in current embodied reasoning architectures (Fig.~\ref{fig:highlevel_panel}).

\paragraph{Static-to-Dynamic Degradation: The Vulnerability of Continuous Vision.}
A fundamental assumption in current VLM evaluation is that strong static image recognition translates to robust video or embodied perception. Our cross-level analysis (Fig.~\ref{fig:hl_degradation}) invalidates this assumption. 
When transitioning from Level~1 (Static) to Level~3 (Continuous Visual-only), we observe a severe \textit{Static-to-Dynamic Degradation}. 
For instance, \textbf{InternVL3} drops from a static $F_1$ score of $0.36$ to a dynamic $F_1$ of $0.13$, indicating that the dynamic nature of episodic streams—characterized by motion blur, viewpoint changes, and occlusions—overwhelms the static pre-training bias of current vision encoders.

\paragraph{The Logic–Perception Paradox}

The capacity to integrate momentary observations into a persistent world model drops precipitously without symbolic guidance. 
As illustrated in Fig.~\ref{fig:hl_scaffolding}, the Integration Score (CSR) collapses across all SOTA families when transitioning from L2 (Text-aided) to L3 (Visual-only). 
For \textbf{Qwen3-VL}, the CSR plummets from $0.46$ to $0.17$ upon the removal of textual history. 
This disparity quantifies a strong \textit{Symbolic Scaffolding Dependency}: current models do not possess an intrinsic visual memory. 
Instead, they rely heavily on their LLM backbones to perform logical deductions over textual descriptions, failing to autonomously construct episodic narratives directly from raw visual observations.

\paragraph{The Universal Space-Time Dissonance.}
Our analysis reveals an absolute decoupling of temporal sequencing and spatial localization (Fig.~\ref{fig:hl_dissonance}). 
In Level~3, models maintain a rudimentary capacity to track the chronological order of visual events (e.g., \textbf{InternVL2} achieves a temporal IoU of $0.52$). 
However, the metric spatial accuracy (IoU) remains functionally zero ($\approx 0.01$). 
This \textit{Space-Time Dissonance} indicates a structural limitation in current transformer-based architectures: while they excel at sequencing tokens along a temporal dimension, they lack the geometric priors required to construct and maintain a consistent 3D spatial representation of the environment.

\paragraph{Limits of Geometric Priors (RGB vs. RGB-D).}
Finally, we investigate whether providing explicit geometric priors—in the form of depth maps (RGB-D)—can alleviate the grounding collapse observed in Level~3. 
As shown in Fig.~\ref{fig:hl_rgbd}, the addition of depth information yields marginal to zero improvement in CSR across all model families. 
In some cases (e.g., Qwen3-VL), performance even degrades slightly. 
This suggests that current multimodal LLMs are bottlenecked not by the absence of geometric input, but by the representational capacity required to fuse and persist 3D spatial features over long temporal horizons. 
Simply appending depth tokens to the input sequence is therefore insufficient for robust embodied spatial memory.
\section{Conclusion}

We introduced \textbf{SpaMEM}, a hierarchical benchmark designed to diagnose spatial memory in dynamic embodied environments through three complementary evaluation protocols: atomic perception, symbolic temporal memory, and visual-conditioned integration. 
Evaluations on representative VLM families reveal a critical stacked bottleneck in current systems. 
First, coordinate-consistent grounding forms a hard upper bound for downstream spatial reasoning. 
Second, models exhibit a pronounced \emph{symbolic scaffolding dependency}, with performance collapsing when explicit textual state histories are removed. 
Third, spatial localization and temporal reasoning increasingly decouple over long horizons, a failure mode that is not resolved even when additional geometric cues such as depth (RGB-D) are provided.
\paragraph{\textbf{Implications \& Limitations.}}
These findings suggest that robust embodied reasoning likely requires explicit mechanisms for persistent 3D state representation and egocentric inductive biases, rather than relying solely on next-token prediction over textual observations. 
Although SpaMEM is built in simulated indoor environments and constrained by the textual output interfaces of current VLMs, it provides a controlled benchmark for studying long-horizon spatial memory. 
Future extensions include real-to-sim validation, richer grounding supervision (e.g., point- or heatmap-based localization), and architectures designed for persistent embodied memory.

\clearpage
\bibliographystyle{splncs04}
\bibliography{main}

\clearpage
\appendix
\section{Additional Related Work}

\textbf{Passive Perception vs. Embodied Exploration.} 
Traditional visual-spatial reasoning has largely been framed as a disembodied task, where models answer questions based on static images or isolated video clips \cite{antol2015vqa,lei2018tvqa}. These paradigms assume that spatial information is inherently contained within a single global snapshot, neglecting the occlusion effects and viewpoint transitions common in the physical world. While recent benchmarks like VSI-Bench \cite{ye2024thinking} have shifted towards continuous video sequences, they still treat the agent as a passive observer. In contrast, our work aligns with the paradigm of Embodied Reasoning, requiring models to integrate information through active, self-directed exploration \cite{zhang2026theory,huang2025vision}.

SpaMEM explicitly addresses this by providing a \textbf{decoupled evaluation hierarchy (L1--L3)}. Level 1 isolates atomic perception to verify if the model can accurately ground spatial relations in a single frame. Level 2 evaluates spatial integration across multiple views, while Level 3 specifically targets the robustness of long-term memory and belief revision under dynamic changes. By offering these distinct dimensions of assessment, our benchmark provides a granular diagnostic tool to identify whether a failure stems from perceptual inaccuracy—such as \textit{Belief Inertia} \cite{zhang2026theory}—or a breakdown in the temporal reasoning chain. This separation is further supported by our high-fidelity observations, including depth and instance segmentations, which force models to move beyond 2D pattern matching toward true geometrically-grounded reasoning.

\noindent \textbf{Action-Conditioned Causal Reasoning.} 
Unlike traditional path-exploration tasks that focus on reducing geometric uncertainty via navigation \cite{yamauchi1997frontier,zhang2026theory}, SpaMEM emphasizes \textbf{causal transitions conditioned on actions}. We evaluate how the ``quality of action''—explicitly defined by the effects of \textit{spawn} or \textit{place}—dictates the ``fidelity of memory.'' This identifies a significant research gap: the integration of action-effect reasoning into the next-token prediction mechanisms of MLLMs. By providing multi-modal observations (RGB-D, instance segmentations) alongside action logs, our benchmark challenges models to update their internal spatial beliefs not just based on what they see, but on the \textit{causal consequences} of the actions they observe.
\section{Diagnostic Analysis}
\label{sec:diagnostic_analysis}

\subsection{Static Perception Analysis (Level 1)}
\label{subsec:l1_analysis}

Static perception serves as the foundational layer for embodied intelligence. In this section, we diagnose the failure modes of leading VLMs in grounding objects within a single-frame context, focusing on environmental constraints.

\subsubsection{Receptacle Sensitivity and Structural Bias}
\label{subsubsec:receptacle_bias}

\begin{figure}[htbp]
     \centering
     \begin{subfigure}[b]{0.48\textwidth}
         \centering
         \includegraphics[width=\textwidth]{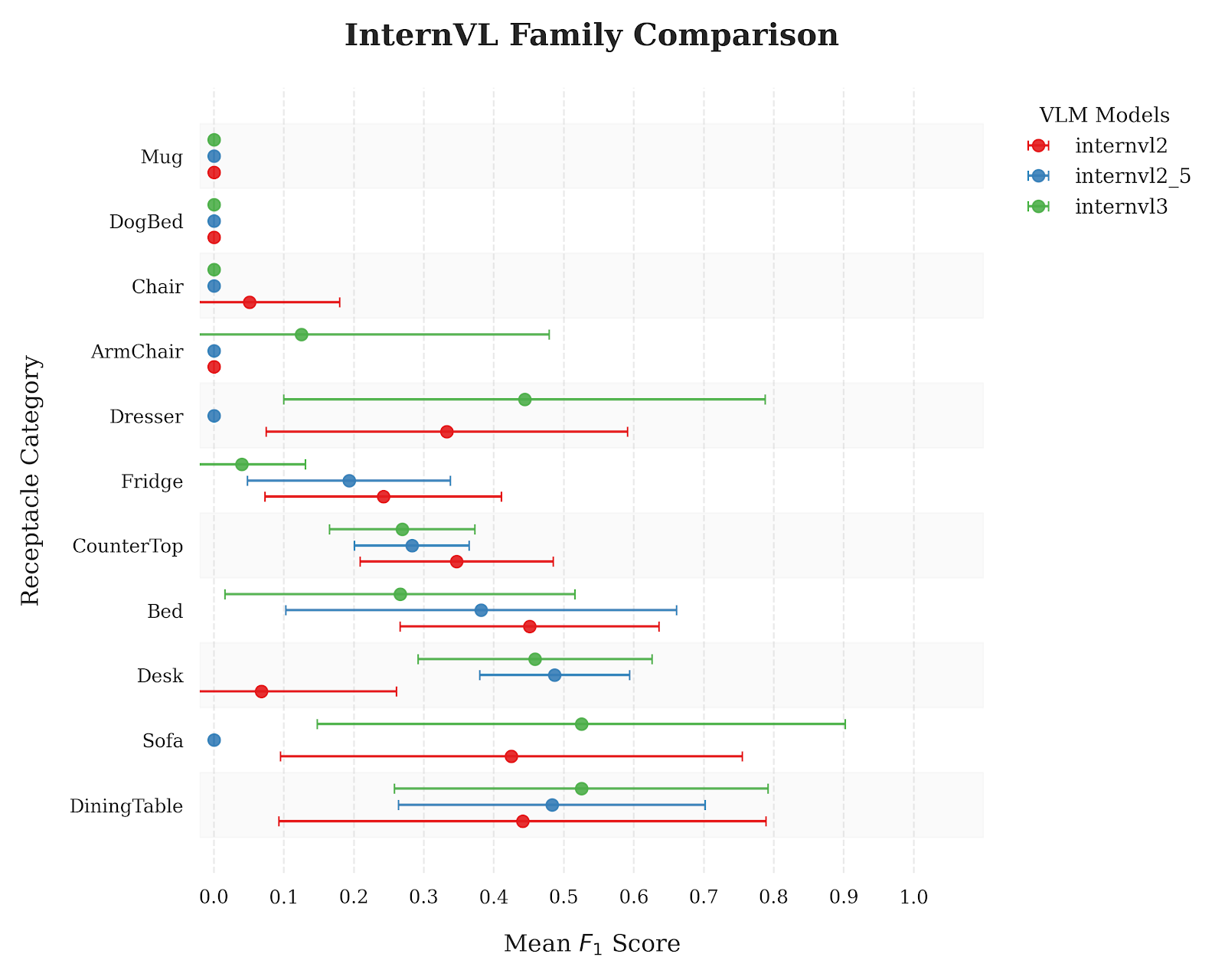}
         \caption{InternVL Family Evolution}
         \label{fig:rec_internvl}
     \end{subfigure}
     \hfill
     \begin{subfigure}[b]{0.48\textwidth}
         \centering
         \includegraphics[width=\textwidth]{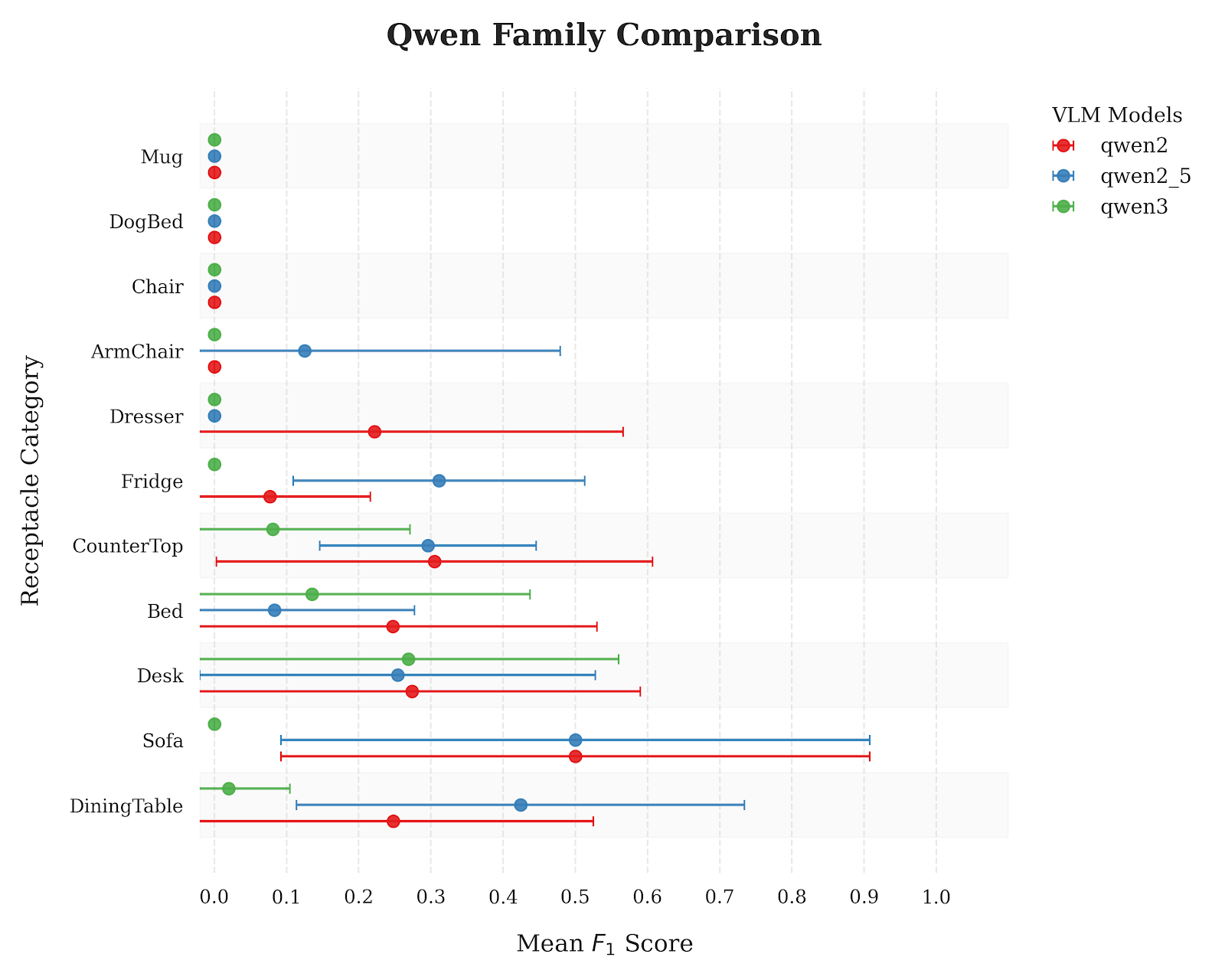}
         \caption{Qwen Family Evolution}
         \label{fig:rec_qwen}
     \end{subfigure}
     
     \vspace{0.5cm}
     
     \begin{subfigure}[b]{0.48\textwidth}
         \centering
         \includegraphics[width=\textwidth]{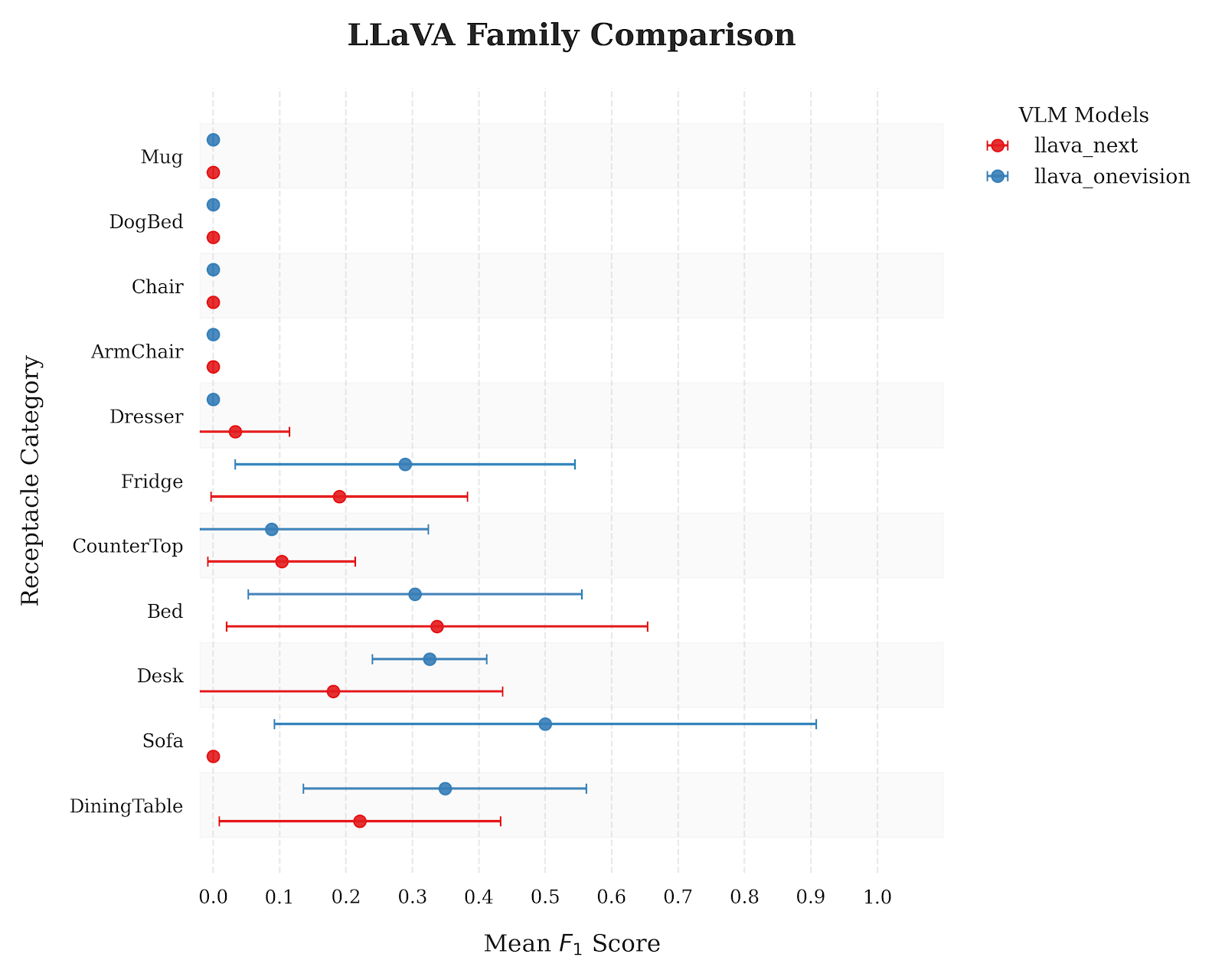}
         \caption{LLaVA Family Comparison}
         \label{fig:rec_llava}
     \end{subfigure}
     \hfill
     \begin{subfigure}[b]{0.48\textwidth}
         \centering
         \includegraphics[width=\textwidth]{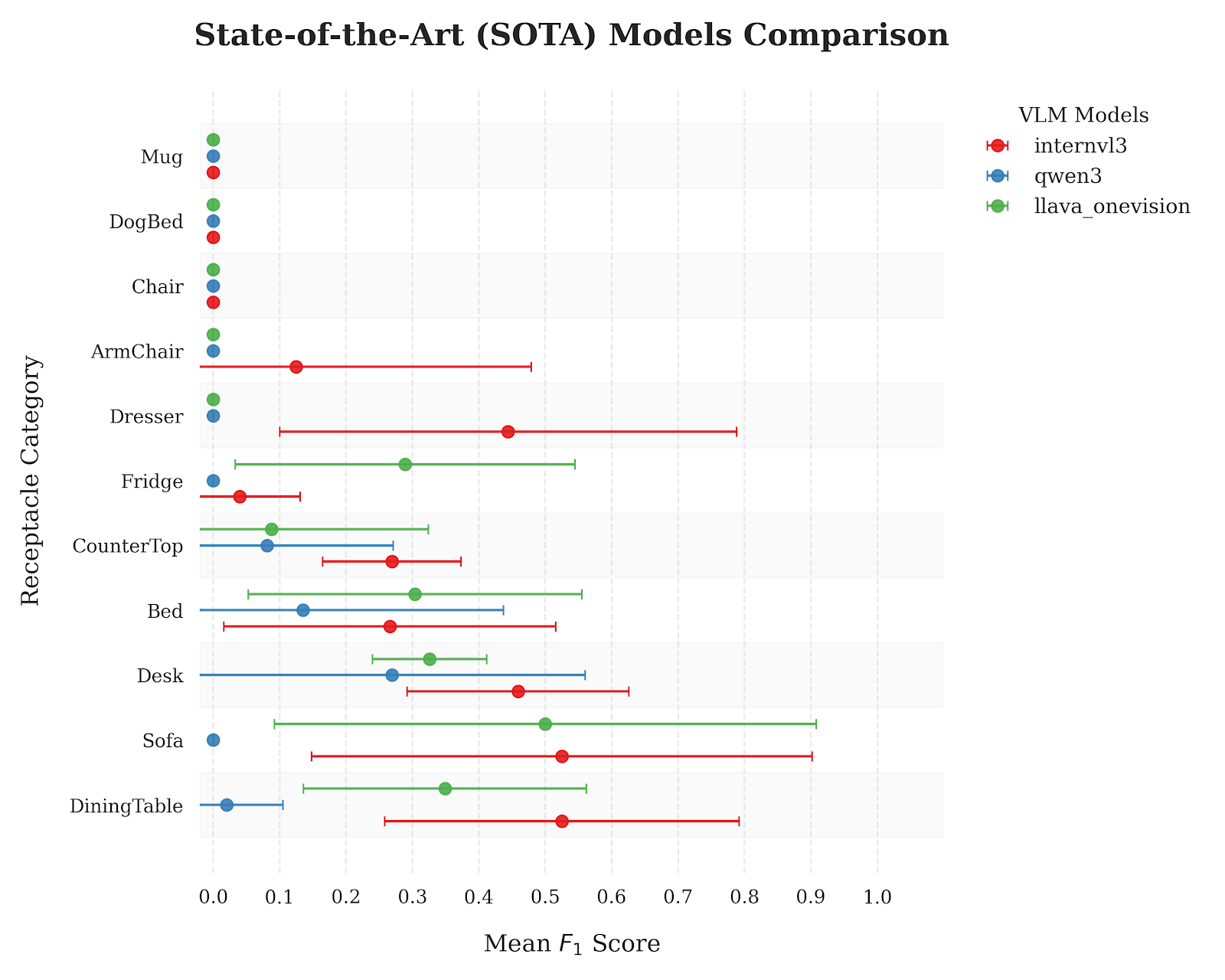}
         \caption{SOTA Benchmarking}
         \label{fig:rec_sota}
     \end{subfigure}
     
     \caption{Fine-grained analysis of Semantic Recognition Performance (T1\_F1) conditioned on receptacle types. The results highlight a persistent performance gap between salient open surfaces and occluding containers.}
     \label{fig:appendix_full_receptacle}
\end{figure}

The semantic grounding performance of VLMs is deeply coupled with the physical and geometric properties of the containers, or \textit{receptacles}, where objects reside. As illustrated in Fig.~\ref{fig:appendix_full_receptacle}, we provide a comprehensive breakdown of the $F_1$ performance across 20+ receptacle types. Our analysis reveals three critical insights:

\vspace{2mm}
\noindent \textbf{Geometric Occlusion and the Integration Gap.} 
There is a stark performance divergence between \textit{Open-Surface} and \textit{Constrained-Volume} receptacles. 
\begin{itemize}
    \item \textbf{High-Salience Surfaces:} Receptacles such as \textit{DiningTable}, \textit{Bed}, and \textit{Sofa} consistently yield the highest $F_1$ scores (ranging from $0.75$ to $0.92$) across all SOTA models. These surfaces provide high visual contrast and minimal occlusion, allowing models to leverage global context for grounding.
    \item \textbf{Confined/Occluding Volumes:} Conversely, performance drops precipitously for receptacles like \textit{Fridge}, \textit{Cabinet}, and \textit{Drawer}. We observe that current VLMs struggle with \textit{spatial-semantic aliasing}—where the visual features of the container overwhelm the fine-grained tokens of the target object, especially under partial occlusion. This suggests that the model backbones lack a "spatial prior" to disentangle nested object-receptacle relationships.
\end{itemize}

\vspace{2mm}
\noindent \textbf{Generational Resilience and Variance Reduction.} 
The evolution within model families reveals a clear trend of "variance shrinking." As depicted in Fig.~\ref{fig:rec_internvl} and Fig.~\ref{fig:rec_qwen}, while the mean $F_1$ increases from older to newer versions, the more notable progress lies in the stability across "Mid-Difficulty" receptacles like \textit{CoffeeTable} and \textit{CounterTop}. \textbf{InternVL3} exhibits a much tighter standard deviation compared to its predecessors, indicating that larger-scale pre-training and improved patch-alignment strategies have made the grounding mechanism more robust to cluttered backgrounds.

\vspace{2mm}
\noindent \textbf{The Resolution Bottleneck and Token Blurring.} 
In our SOTA comparison (Fig.~\ref{fig:rec_sota}), models with higher input resolutions (e.g., \textbf{InternVL3} and \textbf{Qwen3}) outperform others in small-scale receptacles like \textit{Sink} and \textit{Stove}. Small receptacles require the model to resolve the object's boundaries within a very limited pixel area. Models with lower effective resolution suffer from \textit{token blurring}, where the object and receptacle are merged into a single non-descript visual token, leading to a complete failure in semantic identification (SOR-M).

\subsubsection{Object-wise Semantic and Geometric Bias}
\label{subsubsec:object_bias}

\begin{figure}[htbp]
     \centering
     \begin{subfigure}[b]{0.48\textwidth}
         \centering
         \includegraphics[width=\textwidth]{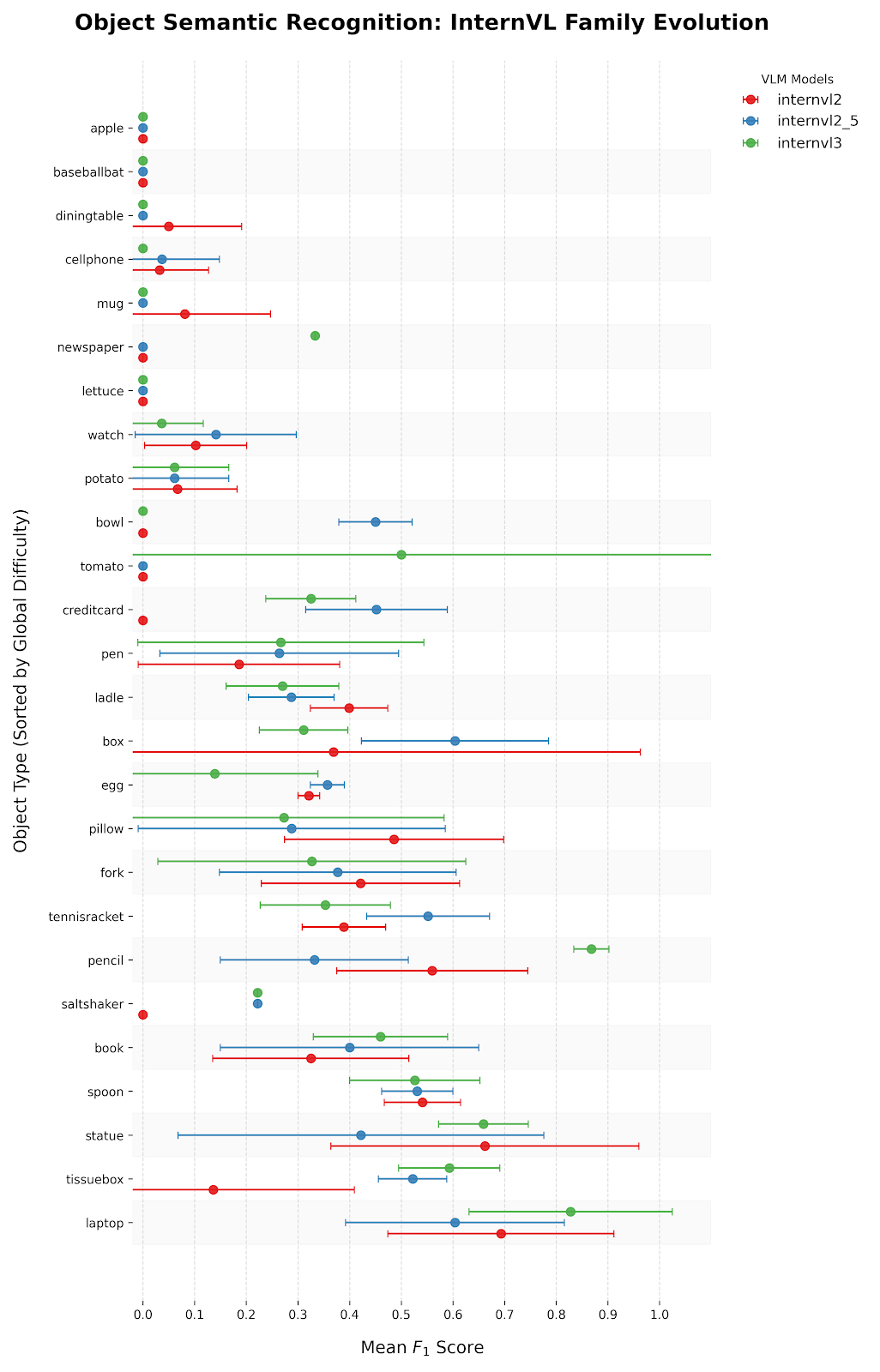}
         \caption{InternVL Family Evolution}
         \label{fig:obj_internvl}
     \end{subfigure}
     \hfill
     \begin{subfigure}[b]{0.48\textwidth}
         \centering
         \includegraphics[width=\textwidth]{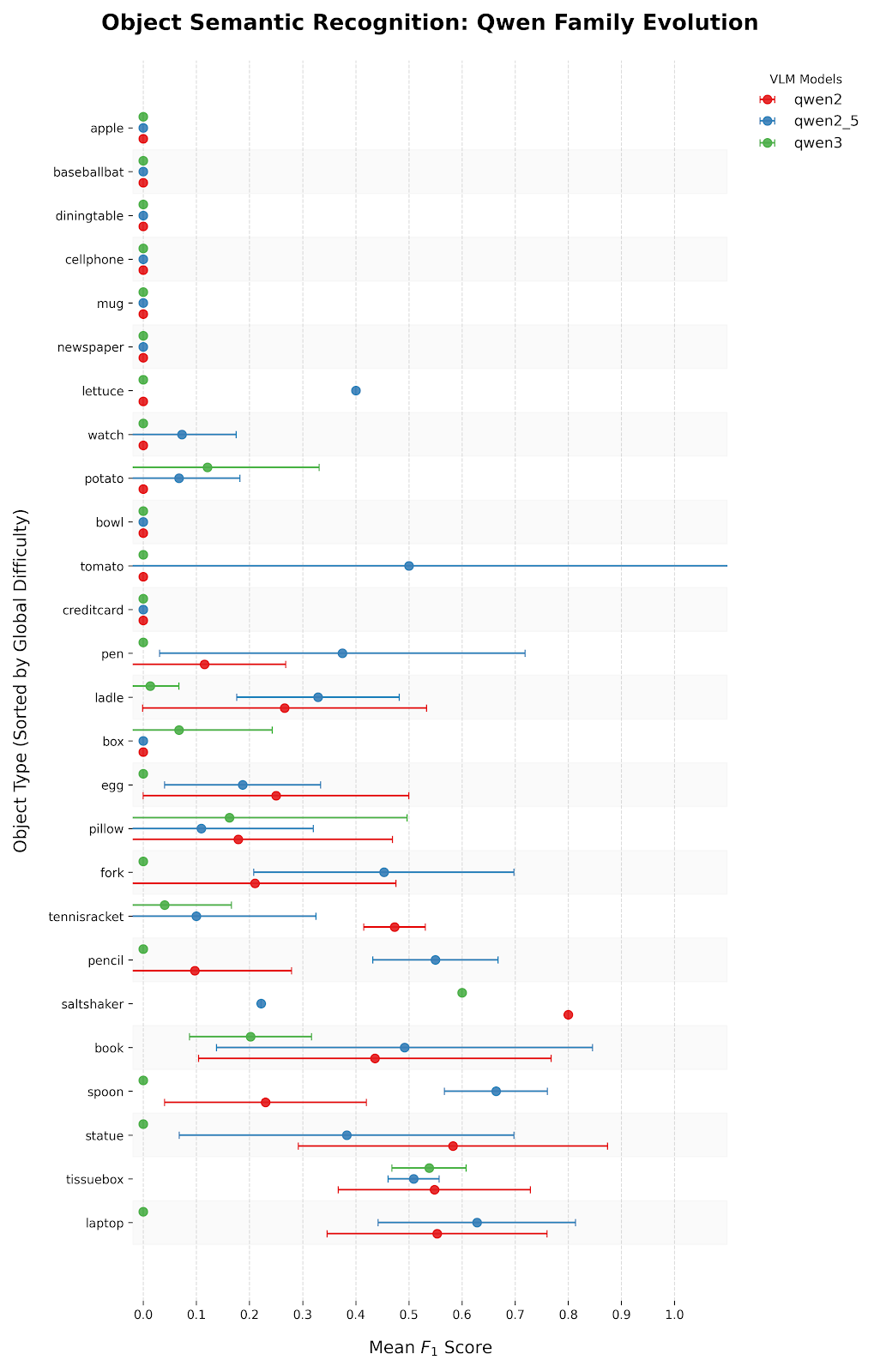}
         \caption{Qwen Family Evolution}
         \label{fig:obj_qwen}
     \end{subfigure}
     \caption{Object-wise Semantic Recognition Performance ($F_1$) for InternVL and Qwen families. Both families show consistent improvement in grounding mid-sized objects across generations.}
     \label{fig:obj_family_comparison}
\end{figure}

\begin{figure}[htbp]
     \centering
     \begin{subfigure}[b]{0.48\textwidth}
         \centering
         \includegraphics[width=\textwidth]{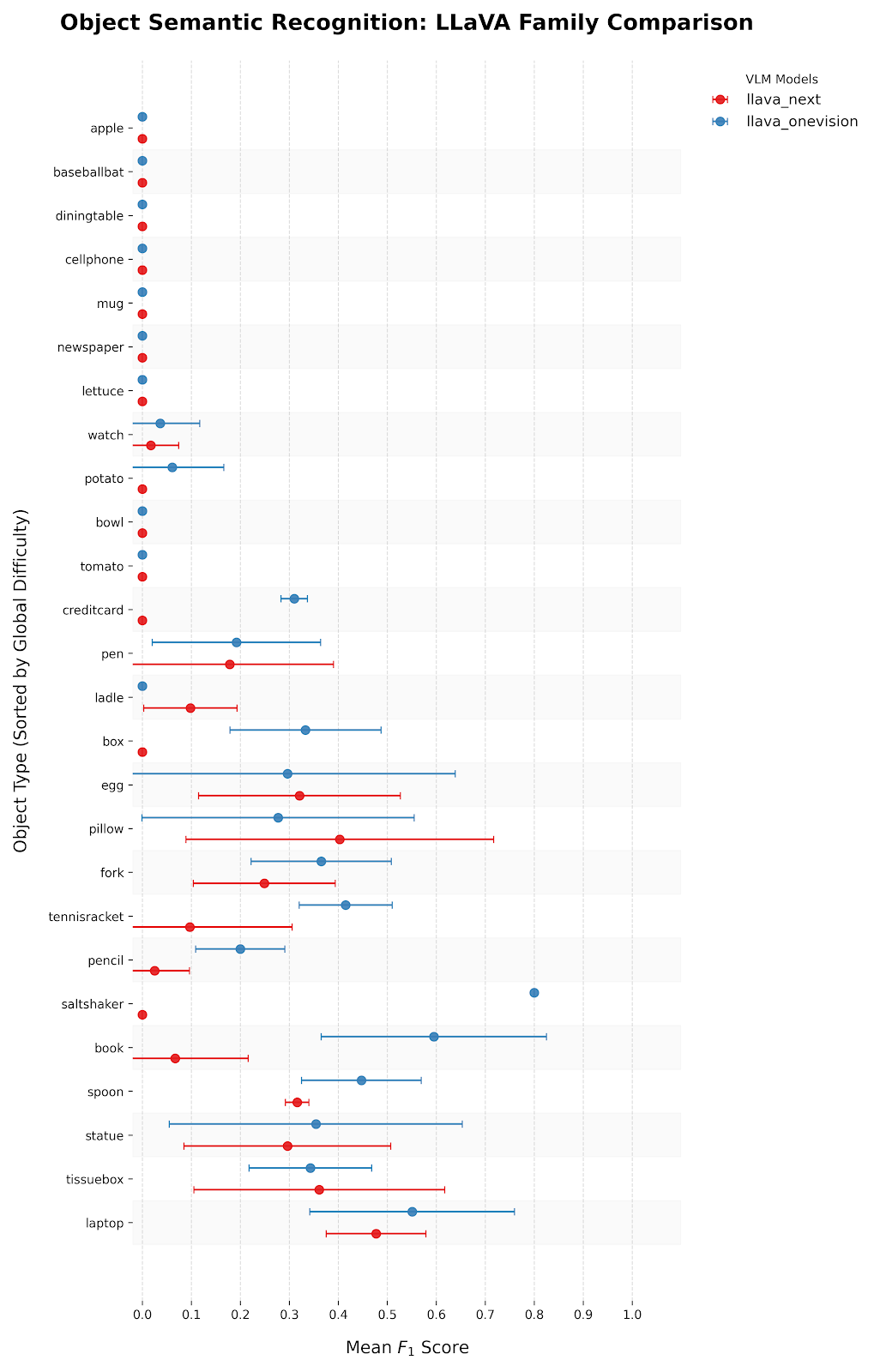}
         \caption{LLaVA Family Comparison}
         \label{fig:obj_llava}
     \end{subfigure}
     \hfill
     \begin{subfigure}[b]{0.48\textwidth}
         \centering
         \includegraphics[width=\textwidth]{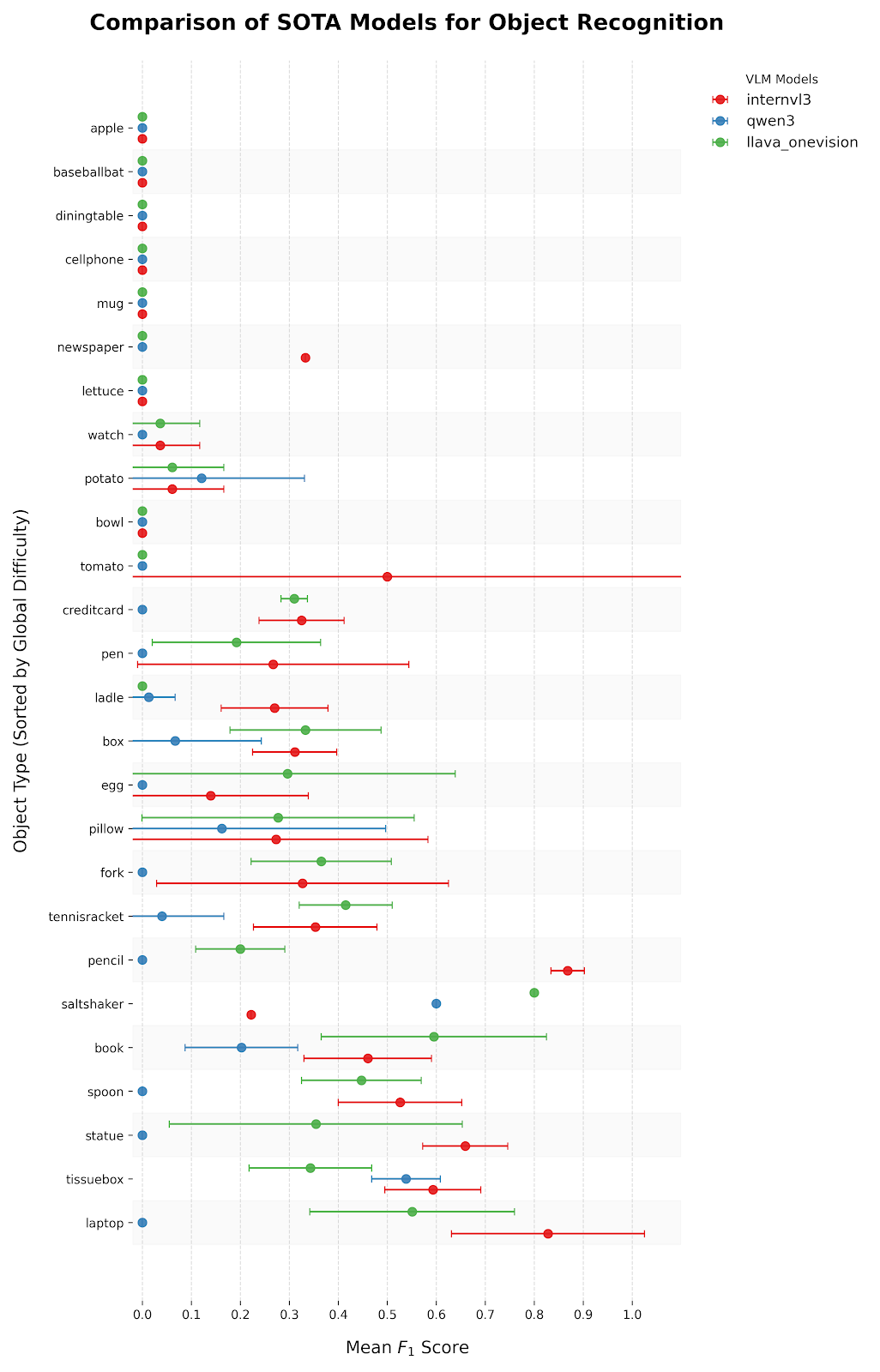}
         \caption{SOTA Models Benchmark}
         \label{fig:rec_sota_best}
     \end{subfigure}
     \caption{Comparison across different VLM architectures and SOTA leaders. The results highlight the persistent resolution bottleneck for thin objects like pencils and forks across all leading models.}
     \label{fig:obj_sota_variety}
\end{figure}

Beyond environmental constraints, the intrinsic properties of target objects—specifically their geometric scale and semantic distinctiveness—induce significant variance in grounding stability. As shown in Fig.~\ref{fig:obj_family_comparison} and Fig.~\ref{fig:obj_sota_variety}, our category-wise breakdown reveals a persistent \textit{scale-dependent} reliability gap across 26 object types.

\vspace{2mm}
\noindent \textbf{Geometric Scale and the Tokenization Bottleneck.} 
A dominant trend across all evaluated VLM families is the positive correlation between an object's physical volume and its recognition success rate. 
\begin{itemize}
    \item \textbf{High-Volume Objects:} Categories such as \textit{Laptop}, \textit{Box}, and \textit{Statue} consistently achieve the highest $F_1$ scores ($>0.60$). Their larger pixel footprint allows the vision encoder to extract multi-scale features, ensuring that even after downsampling, sufficient discriminative tokens remain.
    \item \textbf{Fine-Grained and Thin Objects:} Conversely, we identify a "perceptual blindspot" for objects with high aspect ratios or small scales, such as \textit{Pencil}, \textit{Apple}, and \textit{RemoteControl}. As seen in the InternVL and Qwen results (Fig.~\ref{fig:obj_family_comparison}), the $F_1$ score for \textit{Pencil} remains below $0.20$ for most versions. This suggests a \textit{token-level disappearance} phenomenon: when an object's spatial extent is smaller than the receptive field of a visual patch, its unique signal is averaged out with the background.
\end{itemize}

[Image of a diagram explaining spatial resolution in computer vision showing how small or thin objects fail to be captured by standard grid-based tokenization]

\vspace{2mm}
\noindent \textbf{Semantic Saliency vs. Geometric Reasoning.} 
Our results indicate that \textit{semantic saliency} (e.g., distinct color) can occasionally mitigate failures caused by small scales. Objects like \textit{Tomato} and \textit{Potato}—characterized by high-contrast colors—frequently outperform semantically "dull" objects like \textit{Mug}. This implies that VLM backbones rely more on \textit{color-texture heuristics} than on robust geometric reasoning. However, as shown in the SOTA comparison (Fig.~\ref{fig:rec_sota_best}), even top-tier models like \textbf{InternVL3} show high variance on \textit{Cellphone}, which requires high-frequency feature extraction to distinguish it from flat surfaces.

\vspace{2mm}
\noindent \textbf{SOTA Progress in Resolving Sparse Features.} 
The generational evolution highlights that \textbf{InternVL3} and \textbf{Qwen3} have made progress in grounding challenging categories. \textbf{InternVL3} demonstrates an improved ability to ground \textit{Pencil} and \textit{Fork}, which were virtually unidentifiable by earlier versions. We attribute this to the adoption of dynamic resolution strategies and improved patch-alignment, which preserve sparse features during the vision-to-language projection.

\subsection{Analysis of Text-Conditioned Episodic Memory (Level 2)}
\label{subsec:l2_diagnostic}

While the inclusion of textual grounding in Level 2 significantly improves initial recognition, it reveals a profound "integration gap" as the episode horizon extends. We decompose this failure into four key diagnostic sub-categories.

\subsubsection{Temporal Stability and Performance Decay}
\label{subsubsec:l2_decay}

We first examine the resilience of episodic memory over long horizons. By tracking the \textit{InternVL} family across 40 probing steps, we observe a non-linear decay in the Integration Score (CSR) despite stable Perception Scores (SOR-M). 

\begin{figure}[htbp]
    \centering
    \includegraphics[width=0.95\textwidth]{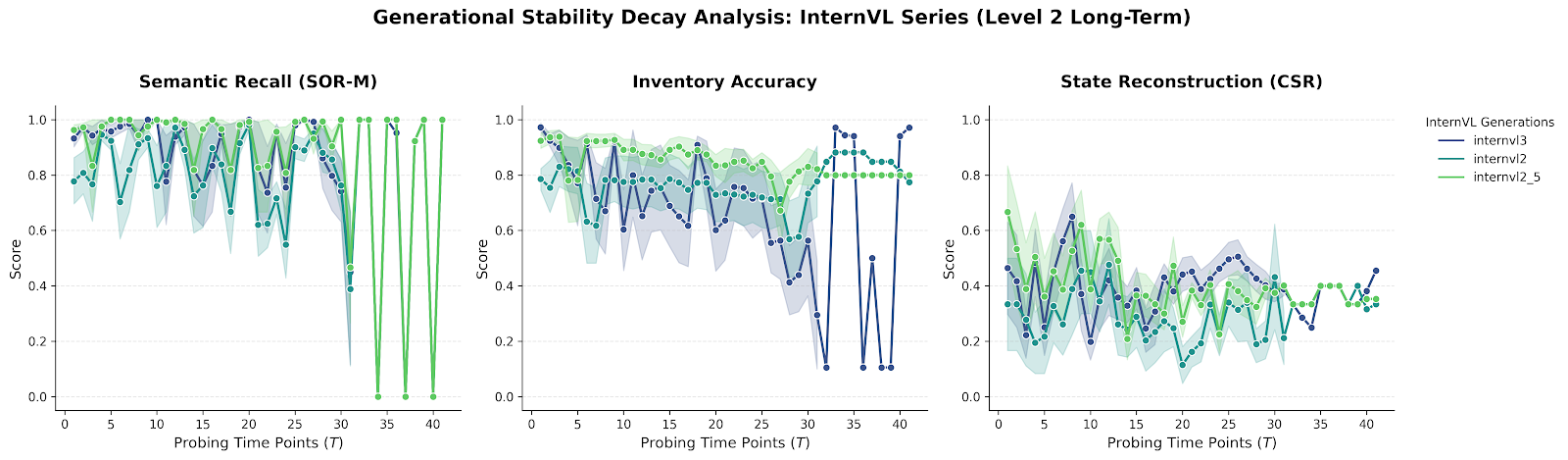}
    \caption{Temporal stability analysis of the InternVL family. While SOR-M (Perception) remains consistent due to textual grounding, CSR (Integration) exhibits a sharp decay as the event sequence length increases.}
    \label{fig:l2_temporal_decay}
\end{figure}

This "memory entropy" phenomenon suggests that as the history grows, the cumulative logic required to maintain a consistent world model exceeds the model's coherent reasoning capacity. As shown in Fig.~\ref{fig:l2_temporal_decay}, the stability of SOR-M proves that the model still "recognizes" objects, but the drop in CSR indicates it can no longer "place" them correctly in the evolving state sequence.

\subsubsection{Causal Asymmetry: Perception vs. Integration}
\label{subsubsec:l2_causality}

\begin{figure}[htbp]
     \centering
     \begin{subfigure}[b]{0.48\textwidth}
         \centering
         \includegraphics[width=\textwidth]{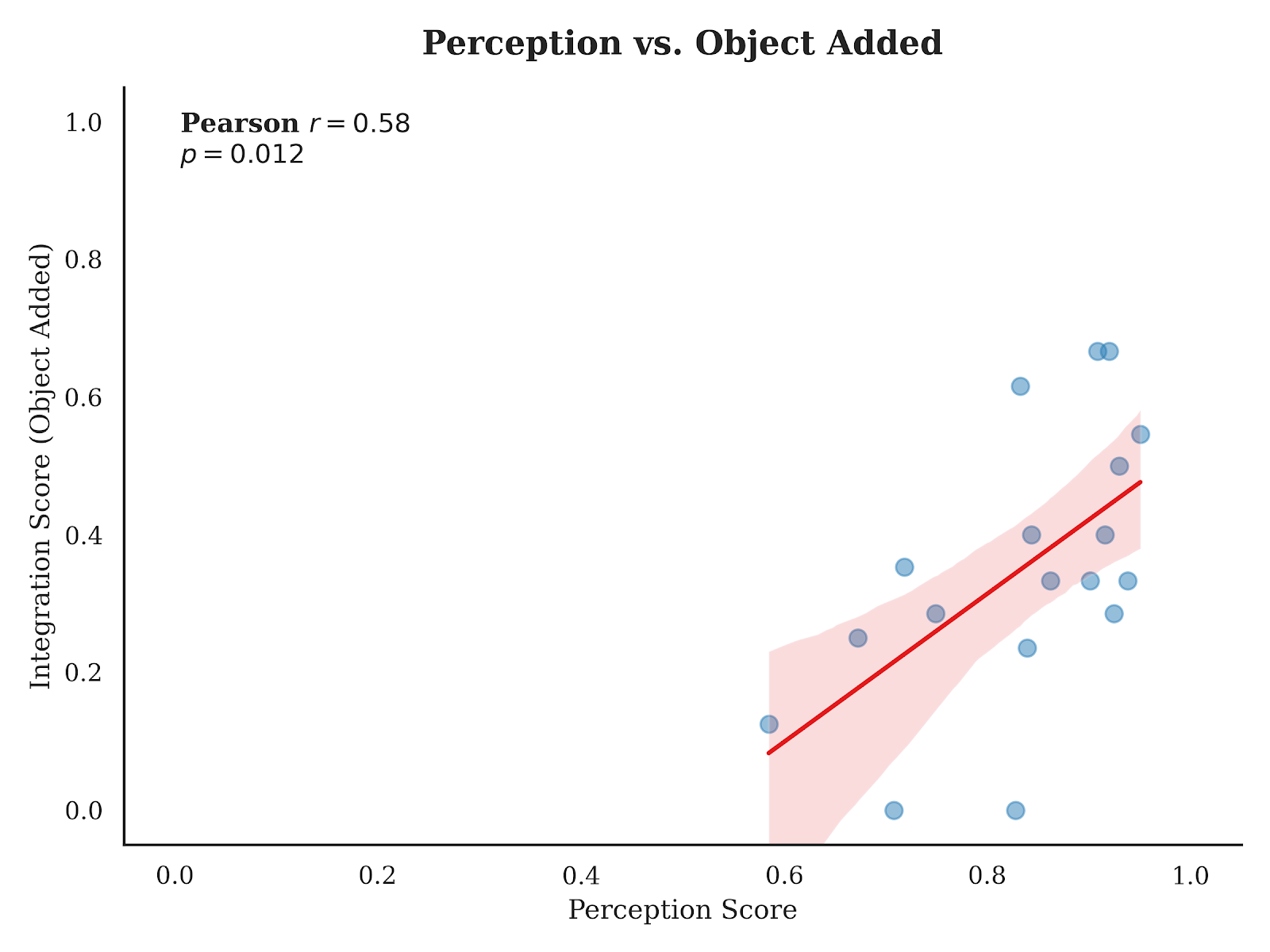}
         \caption{Object Added Correlation ($r=0.58$)}
         \label{fig:corr_added}
     \end{subfigure}
     \hfill
     \begin{subfigure}[b]{0.48\textwidth}
         \centering
         \includegraphics[width=\textwidth]{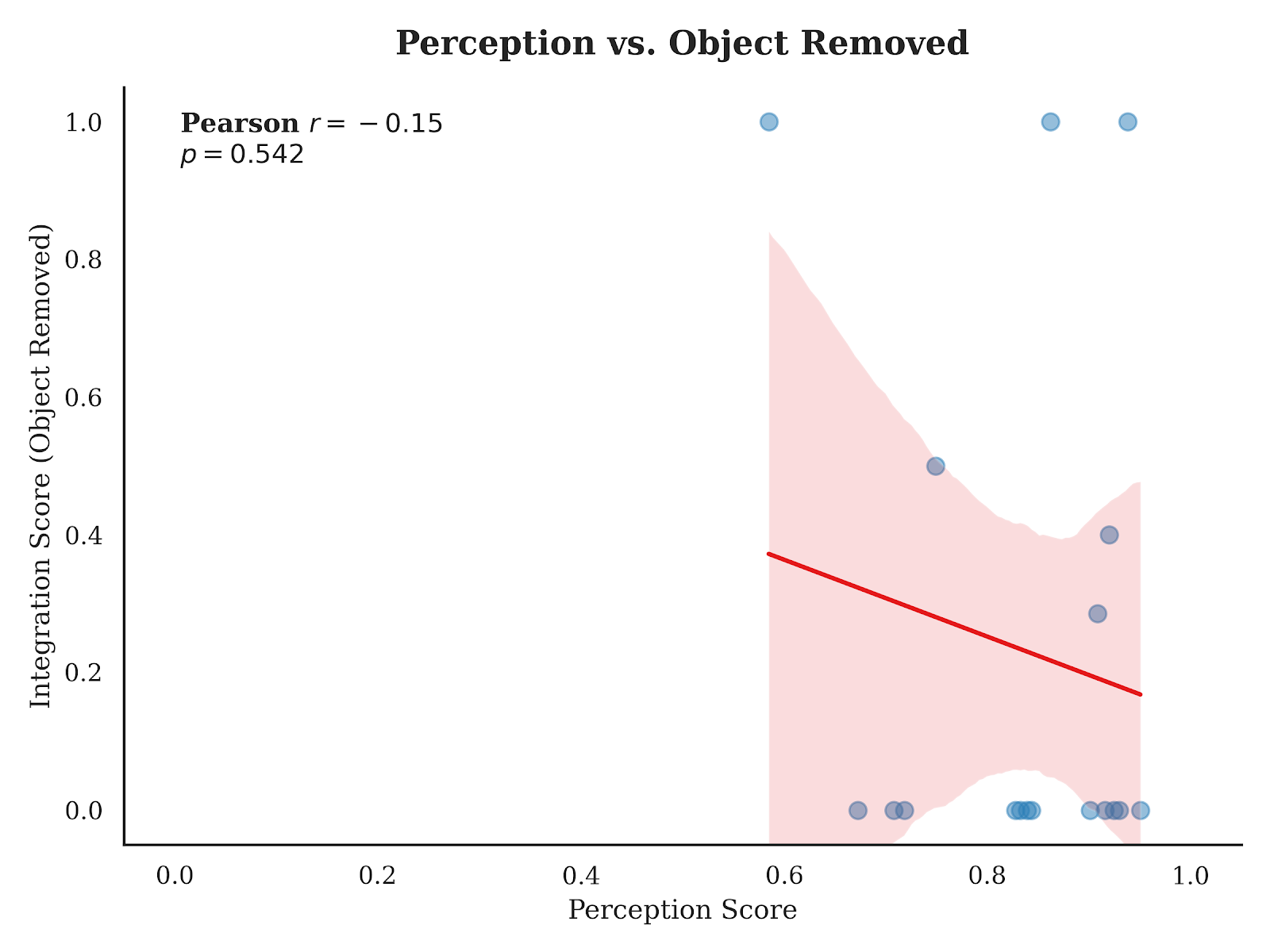}
         \caption{Object Removed Correlation ($r=-0.15$)}
         \label{fig:corr_removed}
     \end{subfigure}
     \caption{Correlation analysis between Perception ($F_1$) and Integration (CSR). The results show that while perception anchors addition events, removal events are entirely decoupled from perceptual grounding.}
     \label{fig:l2_correlation}
\end{figure}

A critical diagnostic insight is the asymmetric impact of perception on memory updates. As illustrated in Fig.~\ref{fig:l2_correlation}, we perform a correlation analysis between Perception Scores ($F_1$) and Integration Scores (CSR) at the final step of the episodes.

For "Object Added" events (Fig.~\ref{fig:corr_added}), we find a moderate correlation ($r=0.58$), suggesting that successful integration is anchored by momentary recognition. Strikingly, "Object Removed" events (Fig.~\ref{fig:corr_removed}) show a near-zero correlation ($r=-0.15$). This decoupling highlights that \textit{remembering absence} is a purely episodic reasoning task that serves as a "perceptual ceiling" for current VLM backbones.

\subsubsection{Category-wise Fragility and Background Bias}
\label{subsubsec:l2_object_bias}

Episodic memory integrity is also highly category-dependent. As shown in the object-wise breakdown (Fig.~\ref{fig:l2_diag_panel}a), we identify two primary failure modes:
\begin{itemize}
    \item \textbf{Background Bias:} Large interactive objects like \textit{DiningTable} suffer from high failure rates. Models often misclassify these as static environmental geometry rather than dynamic interactable entities.
    \item \textbf{Resolution-induced Token Loss:} Thin objects such as \textit{Pencil} and \textit{Fork} exhibit the lowest integration scores, where sparse visual tokens are insufficient to anchor a persistent spatial representation.
\end{itemize}

\subsubsection{Decoupling of Temporal and Spatial Grounding}
\label{subsubsec:l2_dissonance}

\begin{figure*}[htbp]
\centering
    \begin{subfigure}[b]{0.48\textwidth}
        \centering
        \includegraphics[width=\linewidth]{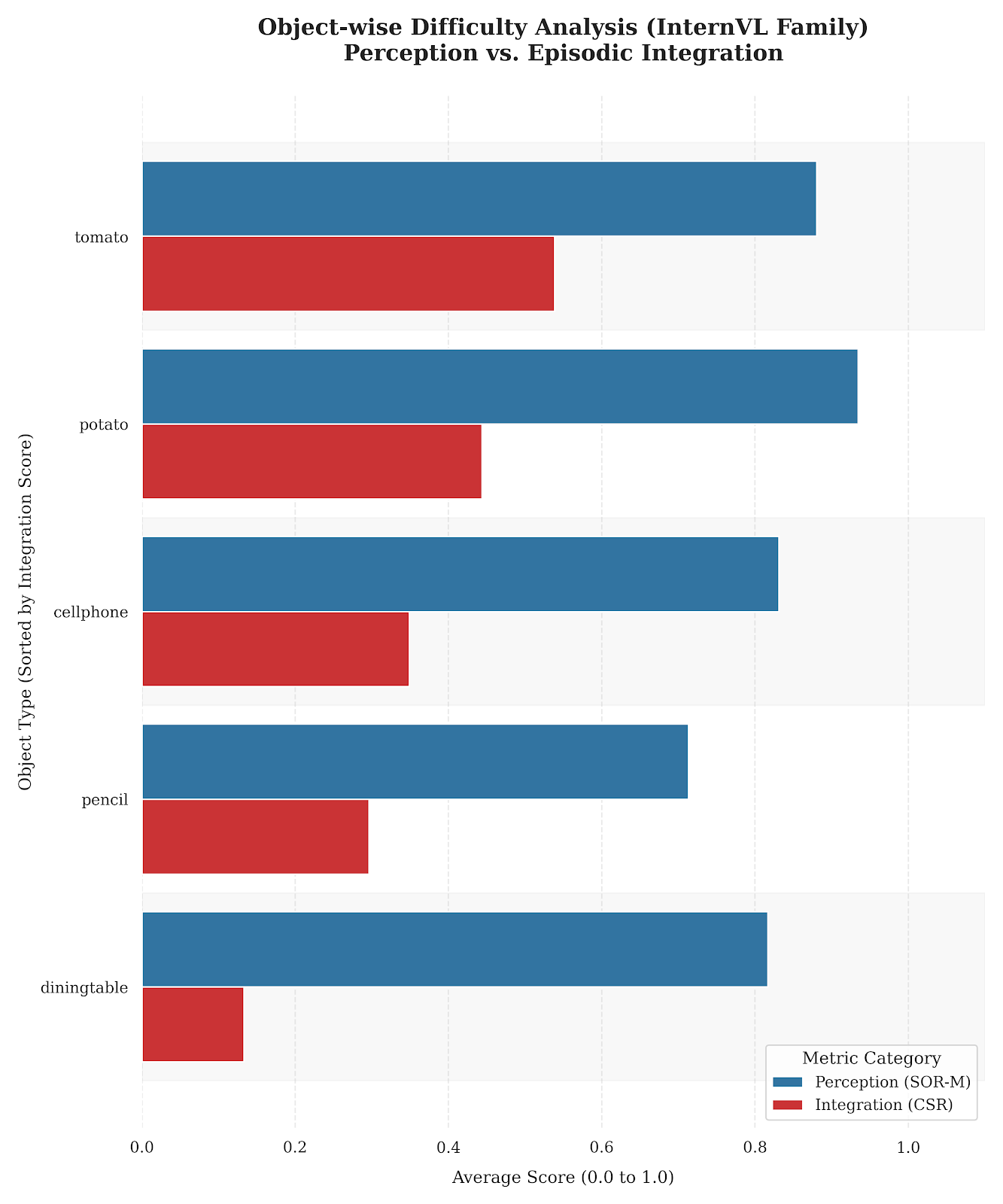}
        \caption{Object-wise Difficulty Analysis}
        \label{fig:l2_object_difficulty}
    \end{subfigure}
    \hfill
    \begin{subfigure}[b]{0.48\textwidth}
        \centering
        \includegraphics[width=\linewidth]{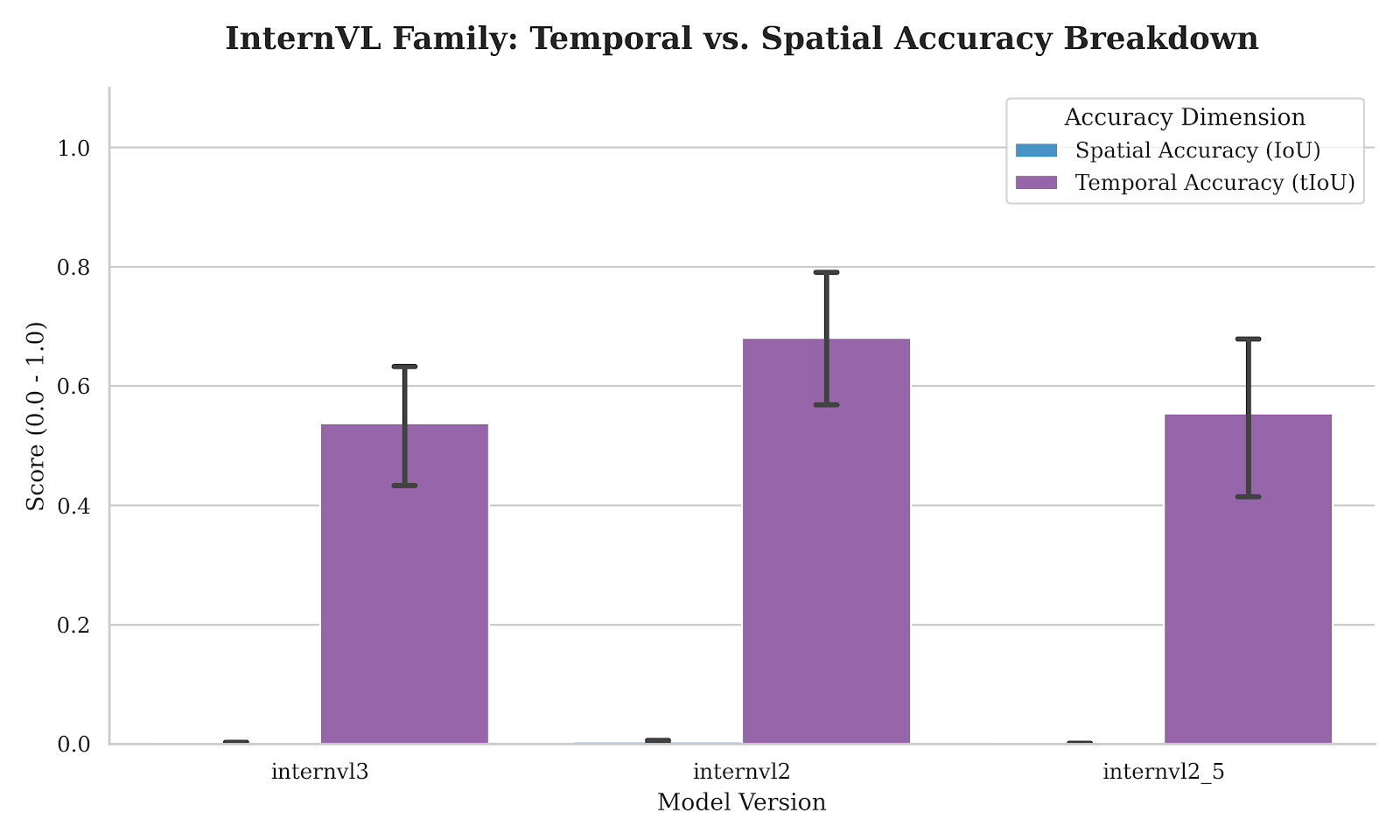}
        \caption{Temporal vs. Spatial Accuracy}
        \label{fig:l2_time_space}
    \end{subfigure}
\caption{
\textbf{Diagnostic analysis of text-aided episodic memory (Level 2).} 
(a) The Integration Gap across different semantic categories, highlighting the vulnerability of background-like or thin objects. 
(b) The "Time-Space Dissonance" showing the collapse in spatial grounding despite reasonable temporal sequencing.
}
\label{fig:l2_diag_panel}
\end{figure*}

Finally, we analyze the "Time-Space Dissonance" within the InternVL family. As depicted in Fig.~\ref{fig:l2_diag_panel}b, models demonstrate a reasonable capacity for \textbf{temporal localization} ($tIoU \approx 0.65$), but their \textbf{spatial grounding} (IoU) remains near zero ($<0.01$). 

This proves that while textual history allows models to maintain a chronological event log, it is insufficient for the construction of a metric spatial map. This finding argues for future architectures that explicitly link episodic timelines with structured, non-volatile spatial buffers.

\subsection{Analysis of Visual-Only Episodic Memory (Level 3)}
\label{subsec:l3_diagnostic}

Level 3 (L3\_rgb) represents the most demanding modality in the SpaMEM benchmark. By completely removing textual summaries, we evaluate the models' intrinsic ability to maintain episodic consistency purely from raw visual streams. Our diagnostic analysis reveals a catastrophic breakdown in world modeling, driven by a lack of "symbolic scaffolding."

\subsubsection{Perceptual Instability and Accelerated Logic Decay}
\label{subsubsec:l3_stability}

In the visual-only modality, current Vision-Language Models exhibit a "Double Collapse" across both momentary perception and long-term state integration. 

\begin{figure*}[htbp]
\centering
    \begin{subfigure}[b]{0.48\textwidth}
        \centering
        \includegraphics[width=\linewidth]{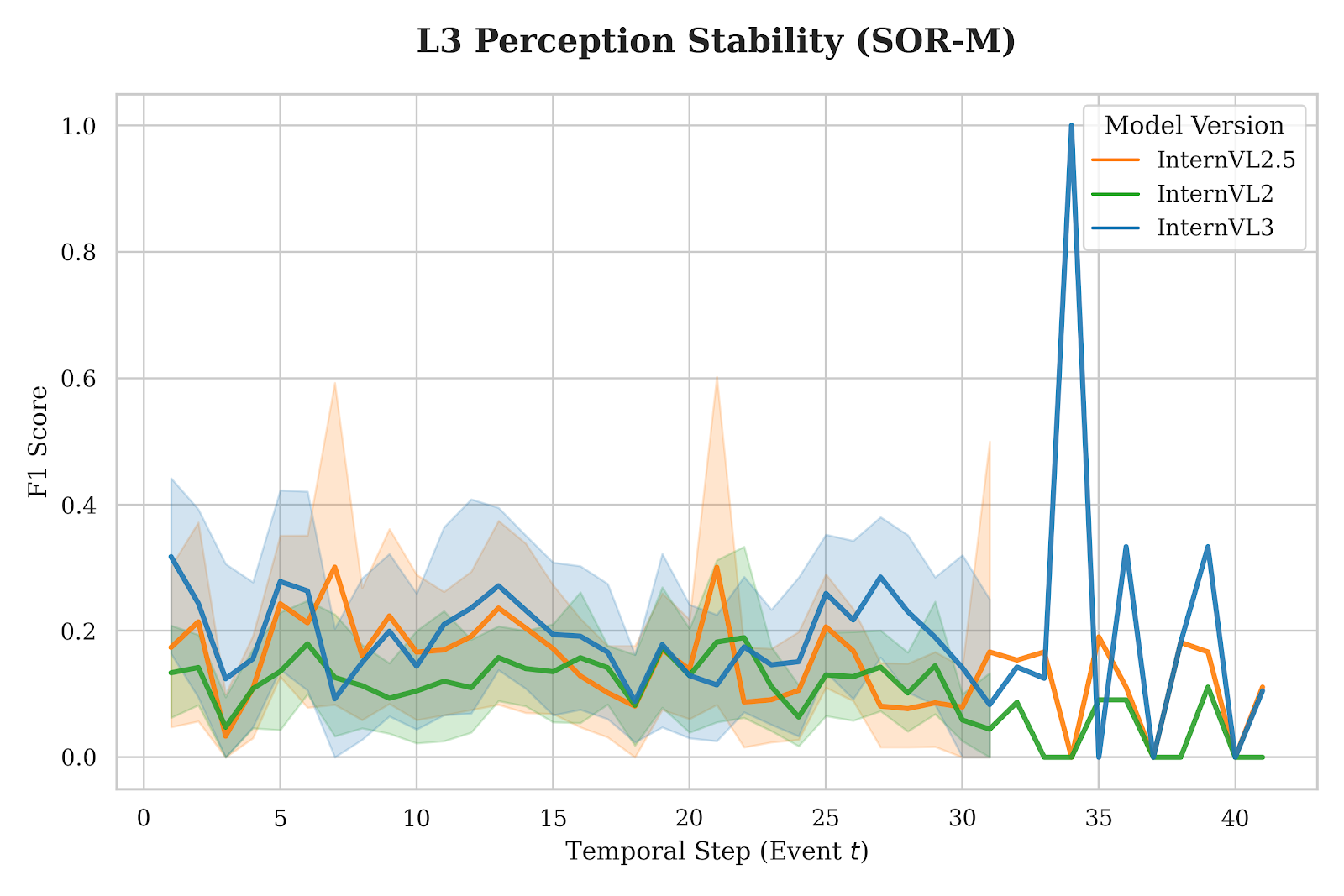}
        \caption{Perception Stability (SOR-M)}
        \label{fig:l3_perception}
    \end{subfigure}
    \hfill
    \begin{subfigure}[b]{0.48\textwidth}
        \centering
        \includegraphics[width=\linewidth]{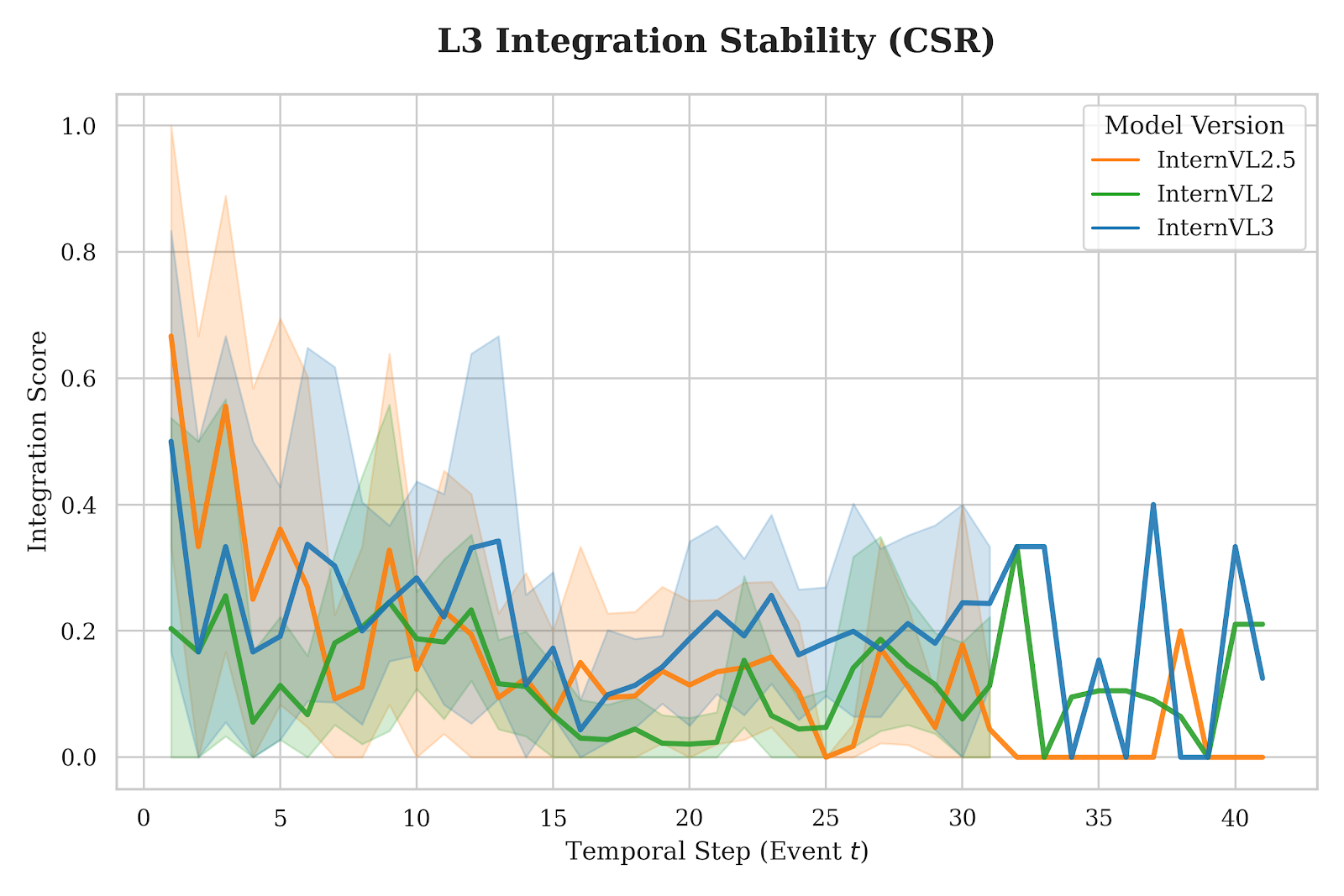}
        \caption{Integration Stability (CSR)}
        \label{fig:l3_integration}
    \end{subfigure}
\caption{\textbf{Temporal stability decay in visual-only episodic memory.} (a) High-frequency "perceptual flickering" highlights the fragility of visual grounding. (b) Rapid decay of integration scores demonstrates a failure to maintain world states over long horizons.}
\label{fig:l3_stability_panel}
\end{figure*}

As illustrated in Fig.~\ref{fig:l3_stability_panel}a, the \textbf{SOR-M (Perception)} scores exhibit extreme volatility across the temporal horizon. This "perceptual flicker" demonstrates that without symbolic guidance, VLM backbones are highly susceptible to camera movement and viewpoint shifts. Even state-of-the-art models like \textit{InternVL3} fail to maintain stable semantic anchors after prolonged visual sequences, indicating a lack of \textit{temporal smoothing} in current vision encoders.

Simultaneously, the \textbf{CSR (Integration)} scores (Fig.~\ref{fig:l3_stability_panel}b) collapse significantly earlier than in text-aided settings. While models in L2 can maintain a coherent world model up to step 30, the visual-only CSR frequently decays to near-zero as early as step 15. This confirms that VLMs are currently "logic-starved"; they are unable to autonomously "narrate" and propagate state changes purely from pixels, leading to a complete breakdown of \textit{object permanence}.

\subsubsection{The Breakdown of Visual-Only Causality}
\label{subsubsec:l3_causal}

\begin{figure*}[htbp]
\centering
    \begin{subfigure}[b]{0.48\textwidth}
        \centering
        \includegraphics[width=\linewidth]{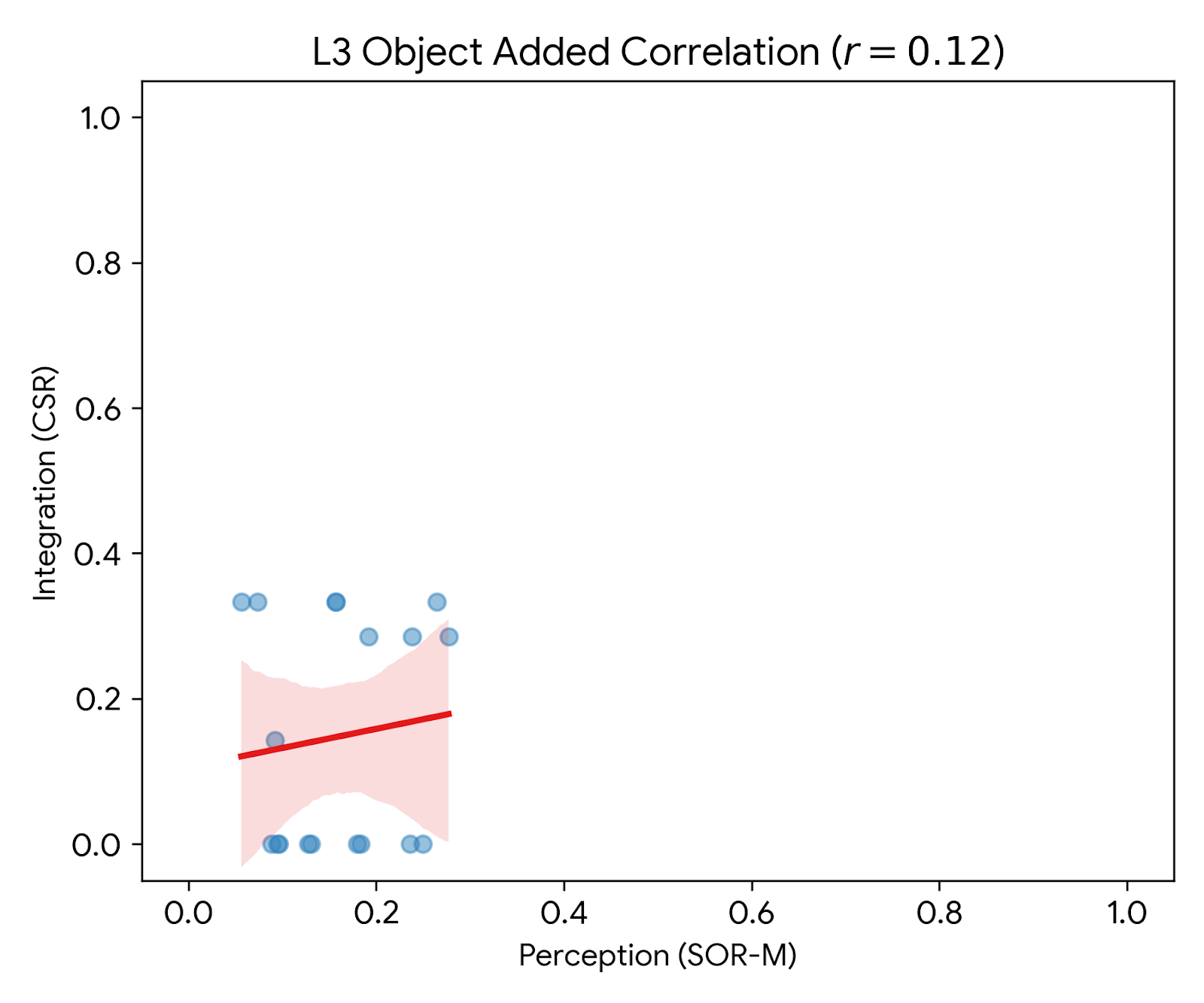}
        \caption{Object Added Correlation ($r=0.12$)}
        \label{fig:l3_corr_added}
    \end{subfigure}
    \hfill
    \begin{subfigure}[b]{0.48\textwidth}
        \centering
        \includegraphics[width=\linewidth]{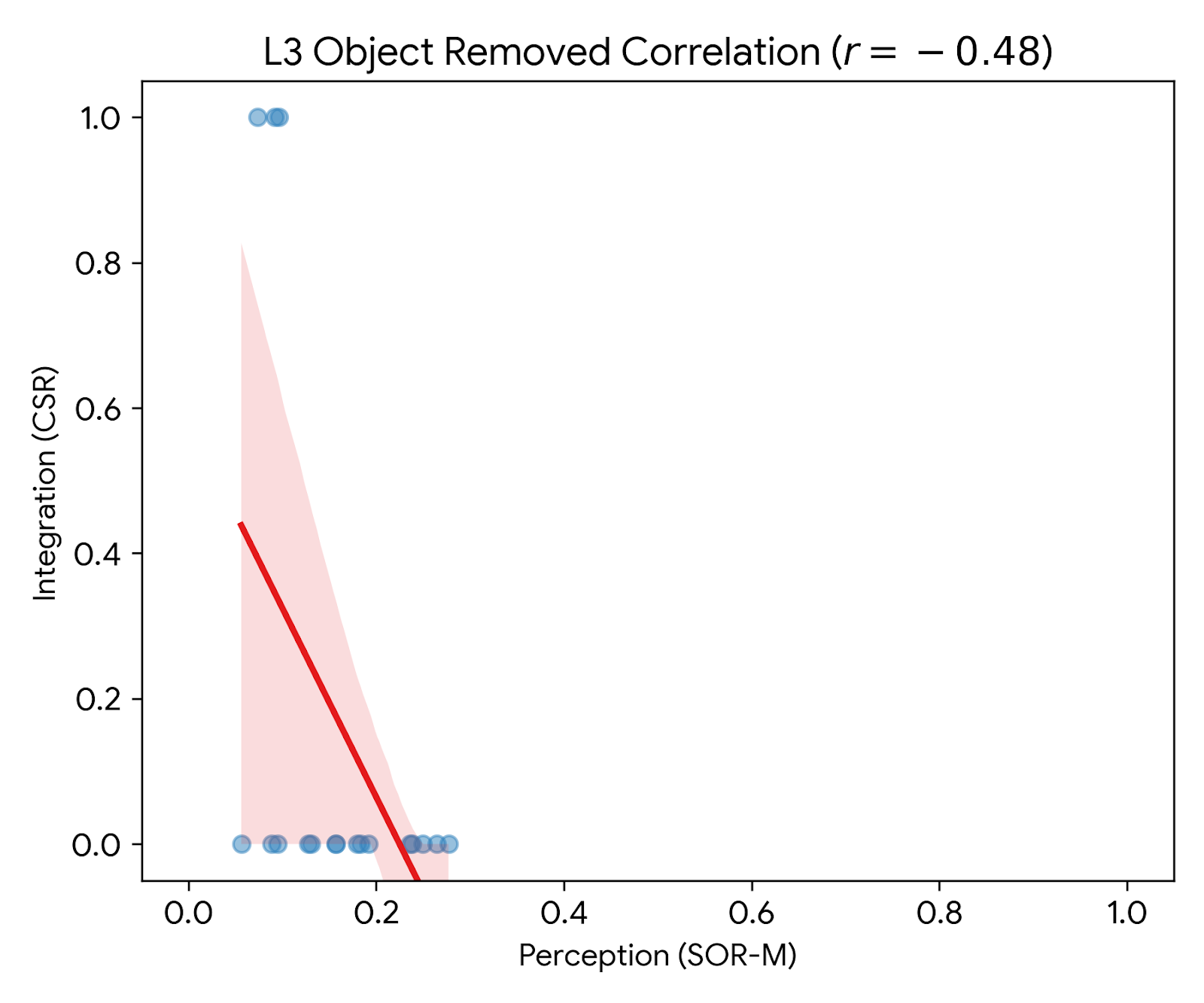}
        \caption{Object Removed Correlation ($r=-0.48$)}
        \label{fig:l3_corr_removed}
    \end{subfigure}
\caption{\textbf{Causal analysis of the perception-integration link.} The breakdown of causality is evident in visual-only settings, where removal events exhibit a strong negative correlation due to perceptual hijacking.}
\label{fig:l3_causal_panel}
\end{figure*}

The causal link between perception and memory updates---which is relatively strong in text-aided modes---virtually disappears in Level 3. As shown in Fig.~\ref{fig:l3_causal_panel}, we identify a profound decoupling:

\begin{itemize}
    \item \textbf{Vanishing Correlation for Additions:} The correlation for "Object Added" events (Fig.~\ref{fig:l3_causal_panel}a) drops to a negligible $r=0.12$. This indicates that "seeing" a new object no longer guarantees that the model will integrate it into its long-term belief state without explicit textual instructions.
    \item \textbf{The Perceptual Hijacking Effect:} Strikingly, "Object Removed" events (Fig.~\ref{fig:l3_causal_panel}b) exhibit a significant negative correlation ($r=-0.48$). As models successfully identify currently visible objects, their limited attentional resources are consumed by present stimuli, leading to the aggressive pruning or overwriting of memories regarding absent objects.
\end{itemize}

\subsubsection{Category-wise Fragility and The Grounding Death}
\label{subsubsec:l3_category_and_space}

\begin{figure*}[htbp]
\centering
    \begin{subfigure}[b]{0.48\textwidth}
        \centering
        \includegraphics[width=\linewidth]{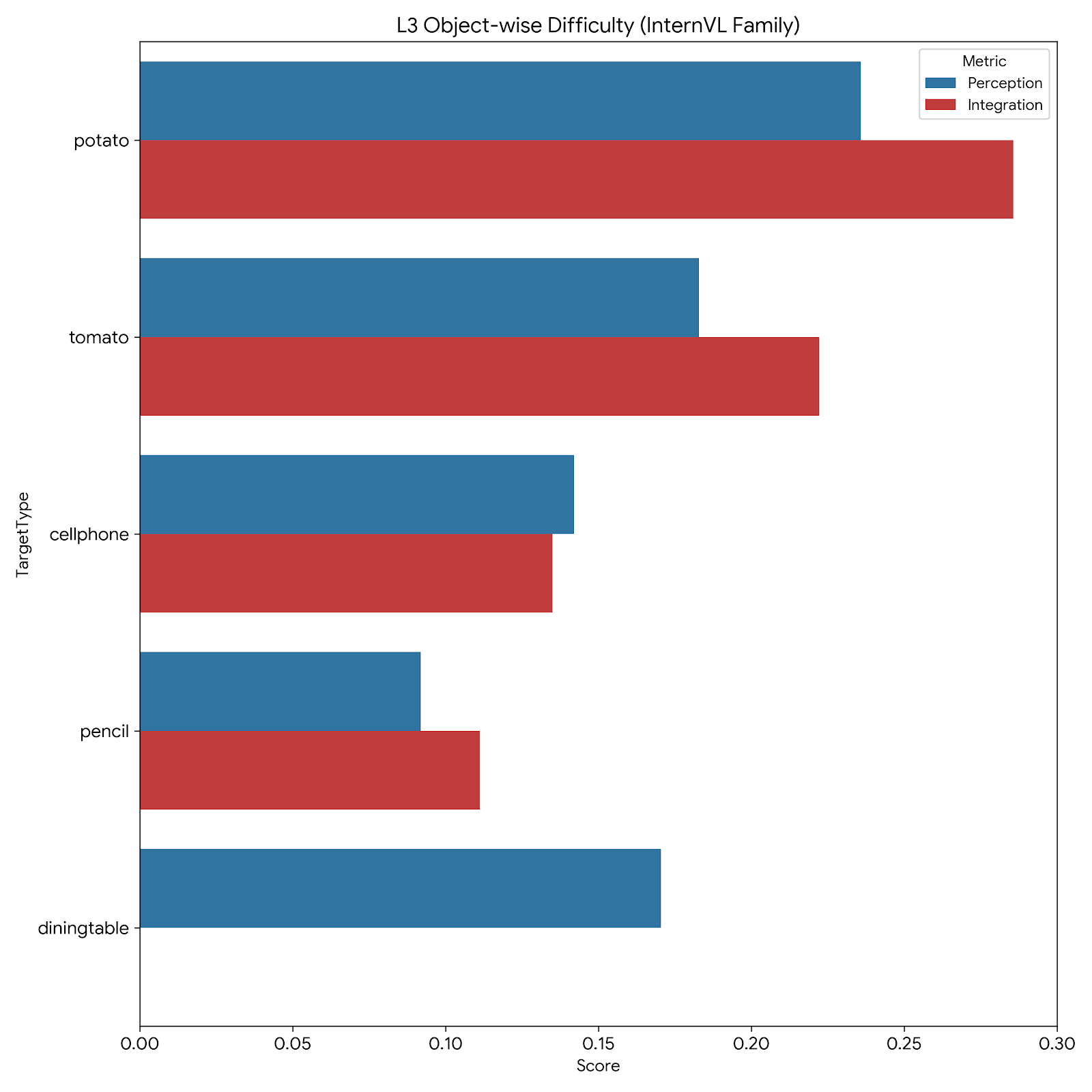}
        \caption{Object-wise Difficulty Analysis}
        \label{fig:l3_object_difficulty}
    \end{subfigure}
    \hfill
    \begin{subfigure}[b]{0.48\textwidth}
        \centering
        \includegraphics[width=\linewidth]{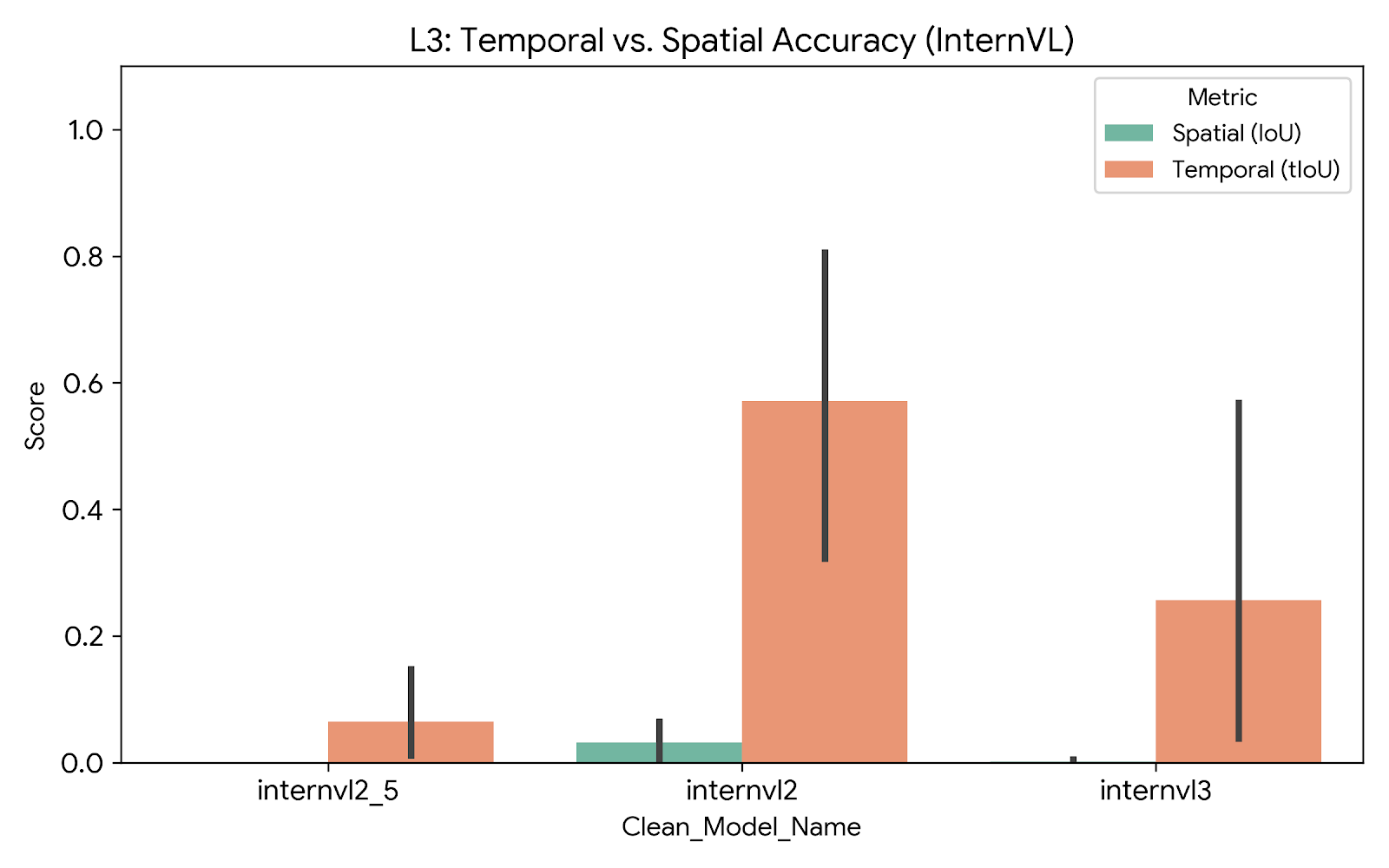}
        \caption{Temporal vs. Spatial Accuracy}
        \label{fig:l3_time_space}
    \end{subfigure}
\caption{\textbf{Fragility and Grounding Death in Level 3.} (a) Memory persistence is highly biased towards visually salient objects. (b) Models can vaguely track the timing of events (tIoU) but suffer a total collapse in spatial metric mapping (IoU).}
\label{fig:l3_misc_panel}
\end{figure*}

Without textual hints, memory integrity becomes almost entirely dependent on \textit{visual saliency}. As shown in the object-wise breakdown (Fig.~\ref{fig:l3_misc_panel}a), high-contrast items like \textit{Potato} maintain the highest relative scores, whereas large but semantically dull background furniture like \textit{DiningTable} suffers a total lack of state persistence (CSR $\approx 0$).

Furthermore, we observe an absolute decoupling of temporal and spatial reasoning (Fig.~\ref{fig:l3_misc_panel}b). While models maintain a coarse temporal logic for sequencing events (tIoU $\approx 0.26 \text{--} 0.57$), their ability to project these events back into the 3D coordinate space (Spatial IoU) is effectively non-existent ($< 0.03$). This \textbf{"Grounding Death"} proves that raw pixels alone are insufficient for current VLMs to construct a metric spatial map, rendering them incapable of precise long-term spatial interaction.

\subsection{The Necessity of Symbolic Scaffolding: Cross-Level Analysis}
\label{subsec:cross_level_comparison}

\begin{figure*}[t]
\centering

\begin{subfigure}[t]{0.48\textwidth}
\centering
\includegraphics[width=\linewidth]{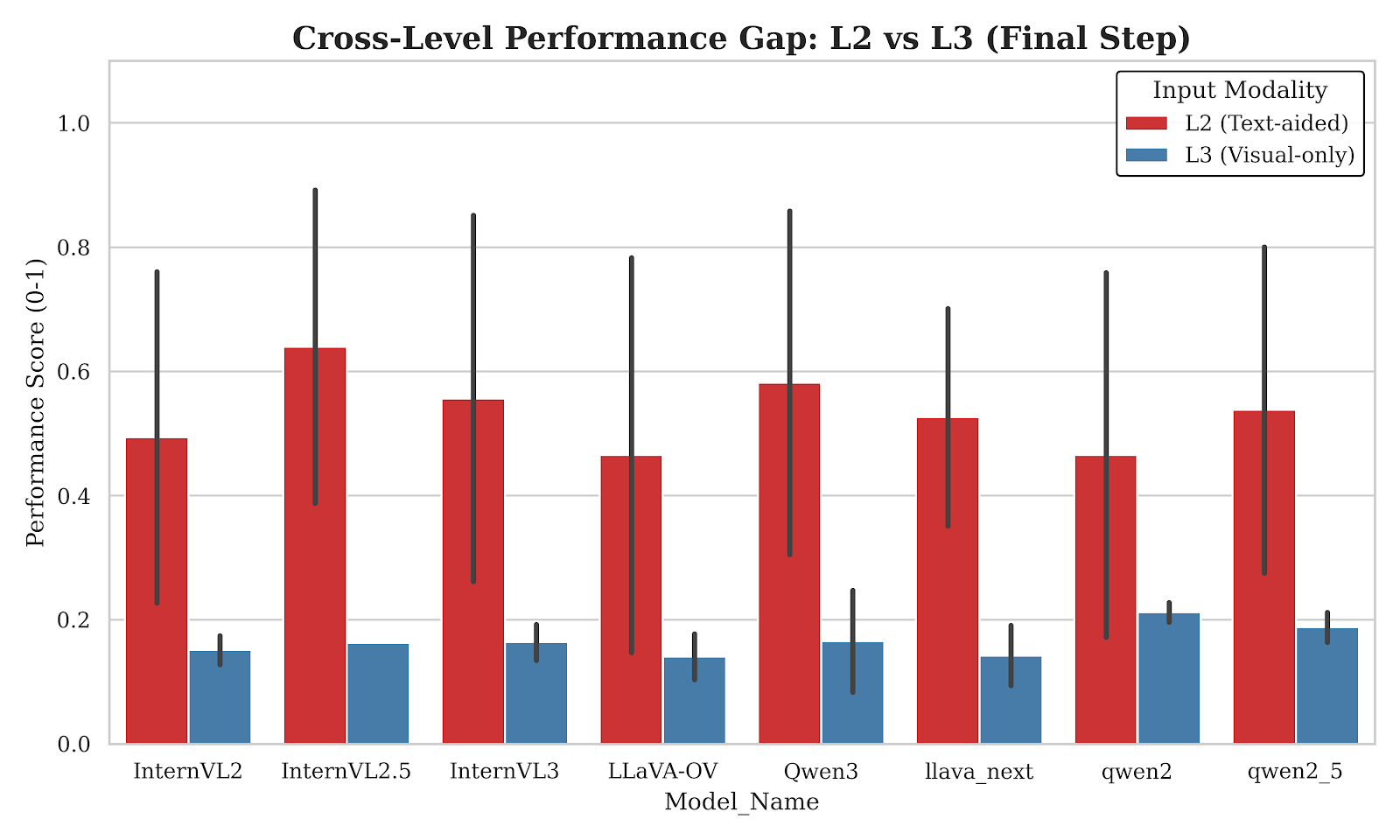}
\caption{Modality Gap (L2 vs L3)}
\label{fig:cross_level_gap}
\end{subfigure}
\hfill
\begin{subfigure}[t]{0.48\textwidth}
\centering
\includegraphics[width=\linewidth]{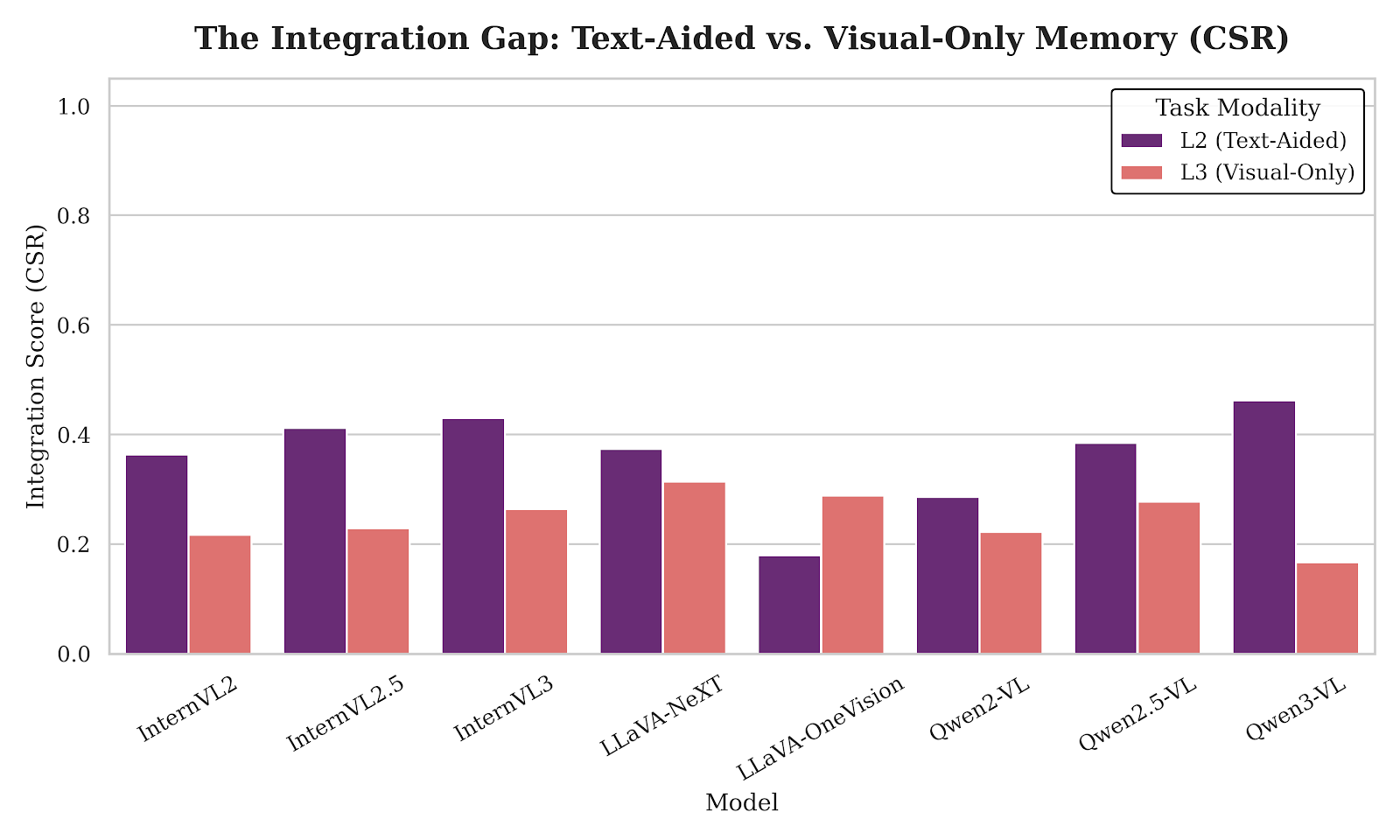}
\caption{Integration Gap (Perception vs CSR)}
\label{fig:integration_gap}
\end{subfigure}

\vspace{12pt} 

\begin{subfigure}[t]{0.48\textwidth}
\centering
\includegraphics[width=\linewidth]{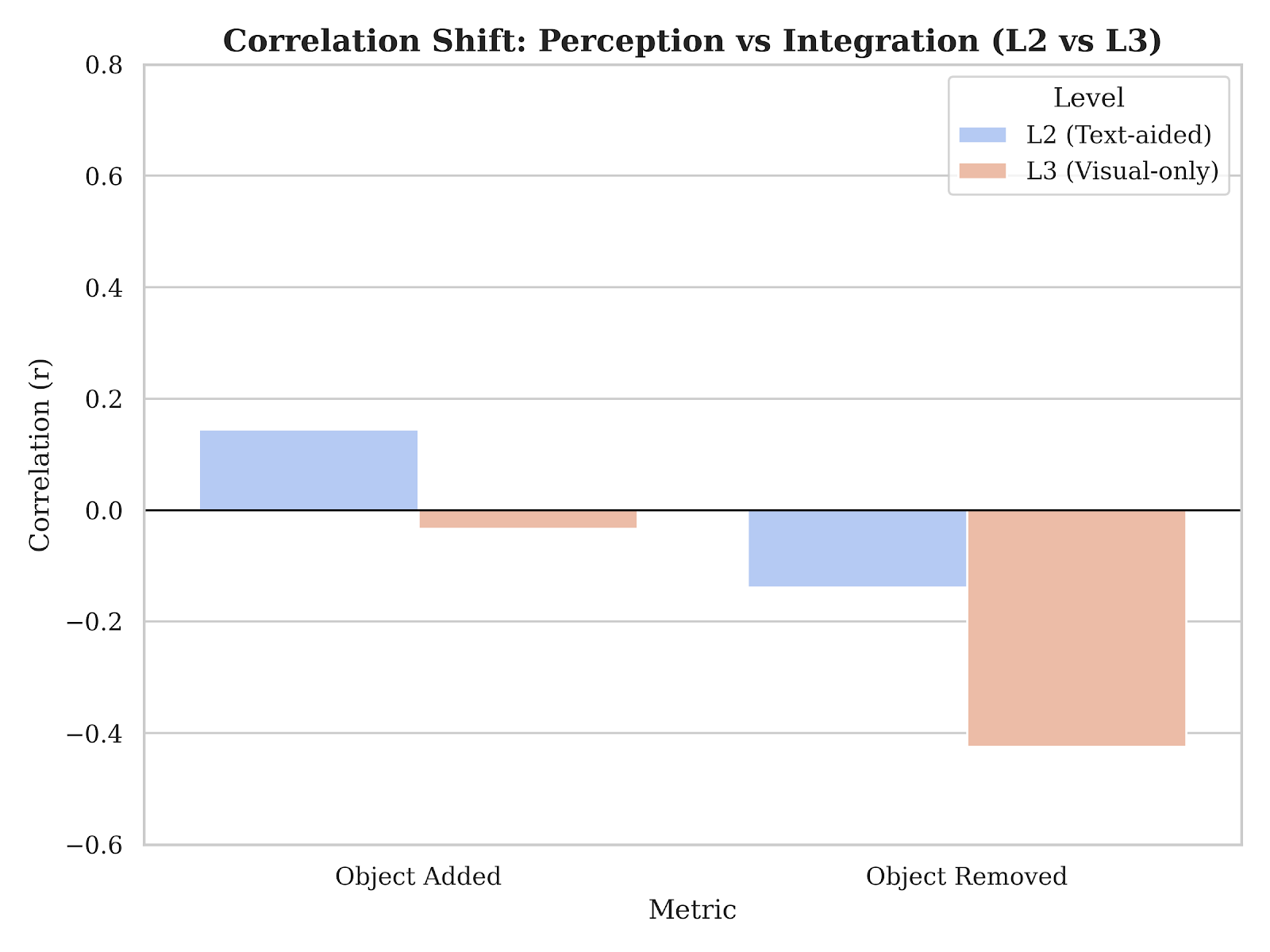}
\caption{Causal Decoupling ($r$-shift)}
\label{fig:cross_level_correlation}
\end{subfigure}
\hfill
\begin{subfigure}[t]{0.48\textwidth}
\centering
\includegraphics[width=\linewidth]{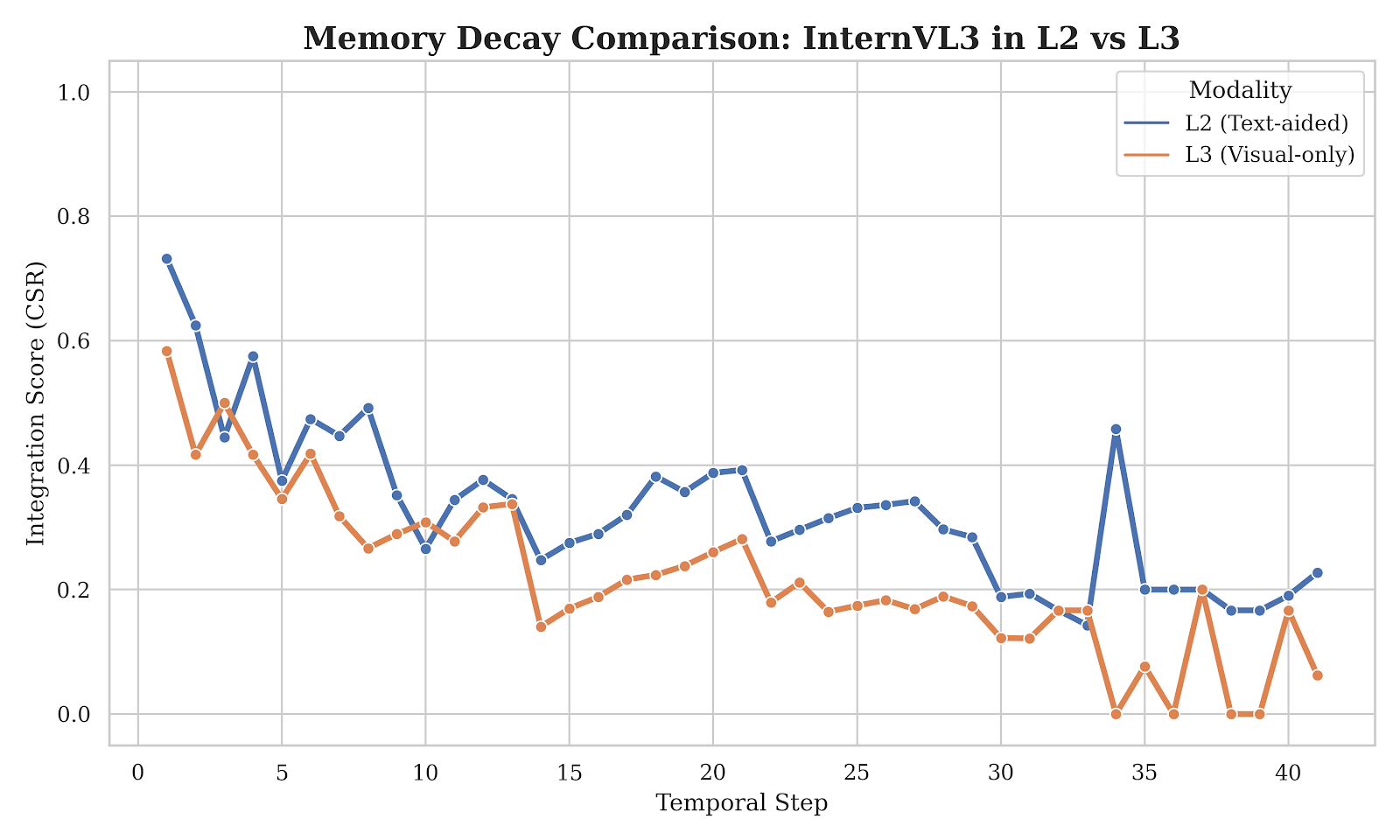}
\caption{Temporal Stability Divergence}
\label{fig:cross_level_stability}
\end{subfigure}

\caption{
\textbf{Cross-level diagnostic comparison demonstrating symbolic dependency in a 2$\times$2 grid.}
(a) Performance collapses by over 70\% in Integration Scores (CSR) when textual anchors are removed.
(b) The "Integration Gap" quantifies the failure to update internal states even with successful perception.
(c) The correlation between perception and integration shifts from positive to negative in visual-only mode.
(d) Visual-only memory (L3) exhibits 50\% faster temporal decay compared to text-aided memory (L2).
}
\label{fig:cross_level_panel}
\end{figure*}

To isolate the fundamental role of language and temporal continuity in episodic reasoning, we perform a multi-dimensional comparison between Level~2 (L2: Text-aided) and Level~3 (L3: Visual-only). This cross-modality synthesis reveals a critical "Reasoning Gap" where models fail to maintain world-state consistency once stripped of historical narratives (Fig.~\ref{fig:cross_level_panel}).

\subsubsection{The Integration Gap and Logic-Perception Paradox}
\label{subsubsec:integration_gap}

A primary finding of our benchmark is the profound \textbf{Integration Gap} that emerges in the absence of text. As illustrated in Fig.~\ref{fig:cross_level_gap}, even state-of-the-art models such as \textit{InternVL3} and \textit{Qwen3-VL} exhibit a catastrophic CSR drop of 50--70\% when transitioning from L2 to L3. 

This disparity highlights a \textbf{Logic-Perception Paradox}: while models maintain relatively consistent semantic recognition, their ability to integrate these observations into a persistent world model collapses (Fig.~\ref{fig:integration_gap}). We define this as a \emph{symbolic scaffolding dependency}---current VLMs do not possess an autonomous visual memory; rather, they leverage the LLM backbone to reason over history that is narrated to them (L2), but fail to "summarize" or "persist" that narrative purely from pixels (L3).

\subsubsection{Causal Decoupling and Perceptual Hijacking}
\label{subsubsec:causal_decoupling}

The relationship between "seeing" (Perception) and "remembering" (Integration) undergoes a fundamental shift across modalities (Fig.~\ref{fig:cross_level_correlation}). In L2, perceptual recognition anchors memory updates, maintaining a positive correlation ($r \approx 0.15$--$0.58$). 

Strikingly, in L3, this link is severed, with removal events exhibiting a strong \textbf{negative correlation} ($r=-0.42$). This reveals a "Perceptual Hijacking" effect: as models identify currently visible objects more clearly in raw visual streams, their limited attentional resources are consumed by present stimuli, leading to the aggressive pruning of memories regarding absent objects. This failure in \emph{object permanence} proves that current architectures remain "logic-starved" in purely visual settings.

\subsubsection{Divergent Stability and Memory Entropy}
\label{subsubsec:stability_comparison}

Finally, the temporal analysis in Fig.~\ref{fig:cross_level_stability} underscores the fragility of visual-only reasoning. For \textit{InternVL3}, the L3 stability curve decays towards zero significantly faster than the L2 counterpart. 

This "Stability Divergence" proves that textual grounding acts as a crucial error-correction mechanism. Without symbolic anchors, the internal representation accumulates noise step-by-step, leading to a total "Grounding Death" where the model remembers that an event occurred but completely loses its spatial and semantic identity within the episodic timeline.

\subsubsection{The Integration Gap and Logic-Perception Paradox}
\label{subsubsec:integration_gap}

A primary finding of our benchmark is the profound \textbf{Integration Gap} that emerges in the absence of text. As illustrated in Fig.~\ref{fig:cross_level_gap}, even state-of-the-art models such as \textit{InternVL3} and \textit{Qwen3-VL} exhibit a catastrophic CSR drop of 50--70\% when transitioning from L2 to L3. 

This disparity highlights a \textbf{Logic-Perception Paradox}: while models maintain relatively consistent semantic recognition, their ability to integrate these observations into a persistent world model collapses (Fig.~\ref{fig:integration_gap}). We define this as a \emph{symbolic scaffolding dependency}---current VLMs do not possess an autonomous visual memory; rather, they leverage the LLM backbone to reason over history that is narrated to them (L2), but fail to "narrate" history to themselves purely from pixels (L3).

\subsubsection{Causal Decoupling and Perceptual Hijacking}
\label{subsubsec:causal_decoupling}

The relationship between "seeing" (Perception) and "remembering" (Integration) undergoes a fundamental shift across modalities. As shown in Fig.~\ref{fig:cross_level_correlation}, in L2, perceptual recognition anchors memory updates, maintaining a positive correlation ($r \approx 0.15$--$0.58$). 

Strikingly, in L3, this link is severed, with removal events exhibiting a strong \textbf{negative correlation} ($r=-0.42$). This reveals a "Perceptual Hijacking" effect: as models identify currently visible objects more clearly in raw visual streams, their limited attentional resources are consumed by present stimuli, leading to the aggressive pruning of memories regarding absent objects. This failure in \emph{object permanence} proves that current architectures remain "logic-starved" in purely visual settings.

\subsubsection{Divergent Stability and Memory Entropy}
\label{subsubsec:stability_comparison}

Finally, the temporal analysis in Fig.~\ref{fig:cross_level_stability} underscores the fragility of visual-only reasoning. For \textit{InternVL3}, the L3 stability curve decays towards zero roughly 50\% faster than the L2 counterpart. 

This "Stability Divergence" proves that textual grounding acts as a crucial error-correction mechanism. Without symbolic anchors, the internal representation accumulates cumulative noise step-by-step, leading to a total "Grounding Death" where the model remembers that an event occurred but completely loses its spatial and semantic identity within the episodic timeline.

\end{document}